\pdfoutput=1

\documentclass[runningheads]{llncs}
\usepackage{graphicx}

\usepackage{tikz}
\usepackage{comment}
\usepackage{amsmath,amssymb} 
\usepackage{color}
\usepackage{subcaption}
\usepackage[font=small,skip=0pt]{caption}
\usepackage{multirow}
\usepackage{booktabs}
\usepackage{xcolor,colortbl}
\definecolor{Gray}{gray}{0.85}
\usepackage{hyperref} 

\usepackage[accsupp]{axessibility}  

\begin{document}
\pagestyle{headings}
\mainmatter
\def\ECCVSubNumber{5127}  

\title{SeedFormer: Patch Seeds based Point Cloud Completion with Upsample Transformer} 


\titlerunning{SeedFormer: Patch Seeds based Point Cloud Completion}
%

\author{Haoran Zhou\inst{1} \and
Yun Cao\inst{2} \and 
Wenqing Chu\inst{2} \and 
Junwei Zhu\inst{2} \and \\
Tong Lu\inst{1}\thanks{Corresponding author.} \and
Ying Tai\inst{2} \and 
Chengjie Wang\inst{2}
}
\authorrunning{H. Zhou et al.}
%
\institute{$^{1}$State Key Laboratory for Novel Software Technology, Nanjing University \\  
$^{2}$Youtu Lab, Tencent}
\maketitle

\begin{abstract}
Point cloud completion has become increasingly popular among generation tasks of 3D point clouds, as it is a challenging yet indispensable problem to recover the complete shape of a 3D object from its partial observation. In this paper, we propose a novel SeedFormer to improve the ability of detail preservation and recovery in point cloud completion.
Unlike previous methods based on a global feature vector, we introduce a new shape representation, namely Patch Seeds, which not only captures general structures from partial inputs but also preserves regional information of local patterns. Then, by integrating seed features into the generation process, we can recover faithful details for complete point clouds in a coarse-to-fine manner. 
Moreover, we devise an Upsample Transformer by extending the transformer structure into basic operations of point generators, which effectively incorporates spatial and semantic relationships between neighboring points.
Qualitative and quantitative evaluations demonstrate that our method outperforms state-of-the-art completion networks on several benchmark datasets. Our code is available at \url{https://github.com/hrzhou2/seedformer}.

\end{abstract}


\section{Introduction}
As a commonly-used and easily-acquired data format for describing 3D objects, point clouds have boosted wider research in computer vision for understanding 3D scenes and objects.
However, raw point clouds, routinely captured by LiDAR scanners or RGB-D cameras, are inevitably sparse and incomplete due to the limited sensor resolution and self-occlusion. It is an indispensable step to recover complete point clouds from partial/incomplete data in real-world scenes for various downstream applications \cite{rusu2008towards,liang2018deep,qi2018frustum}.


Recent years have witnessed an increasing number of approaches applying deep neural networks on point cloud completion. The dominant architecture~\cite{yuan2018pcn,achlioptas2018learning,xiang2021snowflakenet} employs a general encoder-decoder structure where a global feature (or called shape code) is extracted from partial inputs and is used to generate a complete point cloud in the decoding phase. However, this global feature structure possesses two intrinsic drawbacks in its representation ability: (i) fine-grained details are easily lost in the pooling operations in the encoding phase and can hardly be recovered from a diluted global feature in the generation, and (ii) such a global feature is captured from a partial point cloud, thus representing only the ``incomplete'' information of the seen part, and is contrary to the objective of generating the complete shape.

\begin{figure*}[t]
	\begin{center}
		\includegraphics[width=0.99\linewidth]{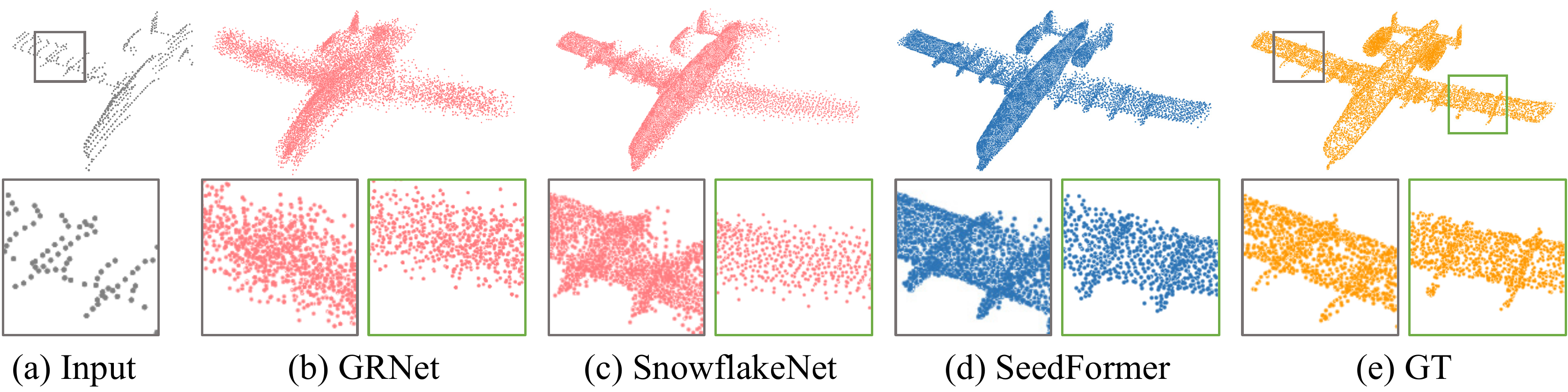}
	\end{center}
	\caption{Visual comparison of point cloud completion results. 
	Compared with GRNet~\cite{xie2020grnet} and SnowflakeNet~\cite{xiang2021snowflakenet}, SeedFormer is better at preserving existing structures (grey bounding box) and recovering missing details (green bounding box).}
	\label{fig:teaser}
\end{figure*}

Besides the overall network architecture, another problem is derived from the designs of point cloud generators. 
Unlike image inpainting which predicts RGB colors of the missing pixels, the 3D point generation is designed differently to predict $(x,y,z)$ coordinates which are unstructured yet continuously distributed in the 3D space. 
This suggests that the desired points, which can describe the missing parts of the target object, have close spatial and semantic relationships with surrounding points in their local neighborhoods.
However, existing methods, either using folding-based generators~\cite{yang2018foldingnet,yuan2018pcn,wen2020point,xie2021style,yu2021pointr} or following a hierarchical structure with MLP/deconvolution implementations~\cite{tchapmi2019topnet,huang2020pf,xiang2021snowflakenet}, attempt to process each point independently by splitting one target point into more. 
These designs neglect the distribution of existing points which results in poor recovery quality of geometric details (see in Fig.~\ref{fig:teaser}(b) and Fig.~\ref{fig:teaser}(c)).

To resolve the aforementioned problems, we propose a novel point cloud completion network, namely \emph{SeedFormer}, with better detail preservation and recovery ability as shown in Fig.~\ref{fig:teaser}. Based on the designed \emph{Patch Seeds} and \emph{Upsample Transformer}, the decoding phase of our method consists of two main steps of: (i) first generating the complete shape from incomplete features in the seed generator, and (ii) then recovering fine-grained details in a coarse-to-fine manner. In the first stage, unlike previous methods using global features, we introduce Patch Seeds as a latent representation in the point cloud completion architecture.
The Patch Seeds characterize the complete shape structure with learned features stored in local seeds. 
This helps generate more faithful details as it preserves regional information which is highly sparse in a global feature.
Secondly, a new point generator is designed for both seed generator and the subsequent layers. Following the idea discussed before, we propose to integrate useful local information into the generation operations by aggregating neighboring points in the proposed Upsample Transformer. In particular, we formulate the generation of new points as a transformer-style self-attention weighted average of point features in the local field. This leads to a better understanding of local geometric features captured by semantic relationships. Qualitative and quantitative evaluations demonstrate the clear superiority of SeedFormer over the state-of-the-art methods on several widely-used public datasets. Our main contributions can be summarized as follows:


\begin{itemize}
\item We propose a novel SeedFormer for point cloud completion, greatly improving the performance of generating complete point clouds in terms of both semantic understanding and detail preservation.

\item We introduce Patch Seeds as a new representation in the completion architecture to preserve regional information for recovering fine details. 

\item We design a new point generator, \textit{i.e.} Upsample Transformer, by extending the transformer structure into basic operations of generating points.



\end{itemize}

\label{sec:intro}

\section{Related Work}
\noindent\textbf{Voxelization-based shape completion.} The early attempts on 3D shape completion~\cite{dai2017shape,han2017high,stutz2018learning} rely on intermediate representations of voxel grids to describe 3D objects. It is a simple and direct way to apply powerful CNN structures on various 3D applications~\cite{maturana2015voxnet,le2018pointgrid,wang2017cnn}. However, this kind of methods inevitably suffers from information loss, and the computational cost increases heavily with regard to voxel resolution. Alternatively, Xie \textit{et al.}~\cite{xie2020grnet} propose to use gridding operations and 3D CNNs for coarse completion, followed by refinement steps to generate detailed structures.

\noindent\textbf{Point cloud completion.} Recently, state-of-the-art deep networks are designed to directly manipulate raw point cloud data, instead of introducing an intermediate representation. The pioneering work PointNet~\cite{qi2017pointnet} proposes to apply MLPs independently on each point and subsequently aggregate features through pooling operations to achieve permutation invariance. Following this architecture, PCN~\cite{yuan2018pcn} is the first learning-based method for point cloud completion, which adopts a similar encoder-decoder design with a global feature representing the input shape. It generates a complete point cloud from the global feature using MLPs and folding operations~\cite{yang2018foldingnet}. Focusing on the decoding phase of generating point clouds with more faithful details, several works~\cite{tchapmi2019topnet,wen2020point,huang2020pf} extend the generation process into multiple steps in a hierarchical structure. This coarse-to-fine strategy helps produce dense point clouds with details recovered gradually on the missing parts. Furthermore, Wang \textit{et al.}~\cite{wang2020cascaded} propose a cascaded refinement module to refine predicted points by computing a displacement offset. Similar ideas are also explored in~\cite{zhang2020detail,xia2021asfm,xie2021style,wen2021pmp}. In order to utilize supervision not only from 3D points but also in the 2D image domain, some other works~\cite{zhang2021view,xie2021style} use rendered single-view images to guide the completion task.
Among the aforementioned methods, the global feature design is widely used due to its efficiency and simplicity, while it still represents intrinsic drawbacks as we discussed before. PoinTr~\cite{yu2021pointr} proposes a set-to-set translation strategy which shares similar ideas with SeedFormer. However, PoinTr designs local proxies which are used for feature translation in transformer blocks. SeedFormer introduces Patch Seeds that can be propagated to the following upsample layers, focusing on the decoding process of recovering fine details from partial inputs. 

\noindent\textbf{Point cloud generators.} Generating 3D points is a fundamental step for point cloud processing and can be generalized to wider research areas. Following PointNet, early generative models~\cite{achlioptas2018learning} use fully-connected layers to produce point coordinates directly from a latent representation in auto-encoders~\cite{rumelhart1986learning,kingma2013auto} or GANs~\cite{goodfellow2014generative}. Then, FoldingNet~\cite{yang2018foldingnet} presents a new type of generators by adding variations from a canonical 2D grid in the point deformation. It requires lower computational cost while assuming that 3D object lies on 2D-manifold. Following this design, \cite{xie2021style} proposes a style-based folding operator by injecting shape information into point generation using a StyleGAN~\cite{karras2019style} design. Moreover, SnowflakeNet~\cite{xiang2021snowflakenet} proposes to generate new points by splitting parent point features through deconvolution. More recently, with the success of transformers in natural language processing~\cite{vaswani2017attention,wu2019pay,devlin2018bert,dai2019transformer}, an increasing number of research have developed such architecture for encoding 3D point clouds. In this work, we further extend the application of transformer-based structure into the basic operations of point cloud generation in the decoding phase, which also represents a new pattern of point generators with local aggregation.
\label{sec:related}

\section{Method}

\subsection{Architecture Overview}
\label{sec:method:overview}
\begin{figure*}[t]
	\begin{center}
		\includegraphics[width=0.98\linewidth]{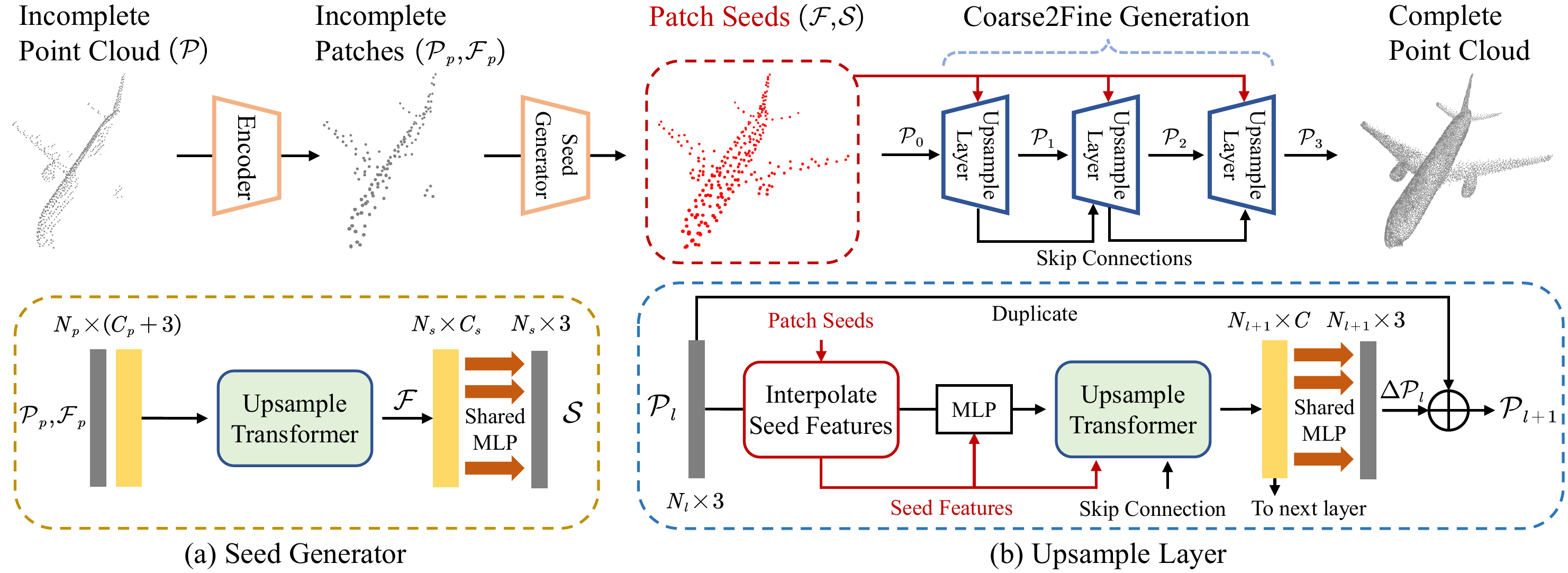} 	
	\end{center}
	\caption{The overall architecture of SeedFormer is shown in the upper part. (a) The seed generator is applied to obtain Patch Seeds which are subsequently propagated into each of the following layers by interpolating seed features. (b) Several upsample layers are used in the coarse-to-fine generation where an Upsample Transformer is applied for producing new points with skip-connection and seed feature encoding.}
	\label{fig:architecture}
\end{figure*}

The overall architecture of SeedFormer is shown in Fig.~\ref{fig:architecture}. We will introduce our method in detail as follows.

\noindent\textbf{Encoder.} Denote the input point cloud as $\mathcal{P} = \{\mathbf{p}_i | i=1,2,...,N\} \in \mathbb{R}^{N \times 3}$ where $N$ is the total number of points and $\mathbf{p}_i$ possesses the $(x,y,z)$ coordinates. The encoder applies point transformer \cite{zhao2021point} and set abstraction layers \cite{qi2017pointnet++} to extract features from the incomplete shape. The number of points is reduced progressively in each layer and then we obtain patch features $\mathcal{F}_p \in \mathbb{R}^{N_p \times C_p}$ and the corresponding patch center coordinates $\mathcal{P}_p \in \mathbb{R}^{N_p \times 3}$, which represent the local structures of the partial point cloud.

\noindent\textbf{Seed generator.} Seed generator aims to predict the overall structure of the complete shape.
It is designed to produce a coarse yet complete point cloud (seed points) as well as seed feature of each point which can capture regional information of local patterns. 
Given the extracted patch features $\mathcal{F}_p$ and center coordinates $\mathcal{P}_p$, we use Upsample Transformer (Sec.~\ref{sec:method:uptrans}) to generate a new set of seed features:
\begin{equation}
    \mathcal{F} = \text{UpTrans}(\mathcal{F}_p, \mathcal{P}_p).
\end{equation}
Then, we obtain the corresponding seed points $\mathcal{S} = \{x_i\}_{i=1}^{N_s} \in \mathbb{R}^{N_s \times 3}$ by applying MLPs on the generated seed features $\mathcal{F} = \{f_i\}_{i=1}^{N_s} \in \mathbb{R}^{N_s \times C_s}$. Grouping seed points and features together yields our Patch Seeds representation (more details are discussed in Sec.~\ref{sec:method:seed}). 


\noindent\textbf{Coarse-to-fine generation.} Afterwards, we follow the coarse-to-fine generation \cite{yang2018foldingnet,huang2020pf,xiang2021snowflakenet} to progressively recover faithful details in a hierarchical structure. This process consists of several upsample layers, each of which takes the previous point cloud and produces a dense output $\mathcal{P}_l$ ($l=1,2,...$) with an upsampling rate of $r_l$ where our Patch Seeds are closely involved. Specifically, each point in the input point cloud is upsampled into $r_l$ points using the Upsample Transformer. 
The coarse point cloud $\mathcal{P}_0$ which is fed to the first layer is produced by fusing seeds $\mathcal{S}$ and the input point cloud $\mathcal{P}$ using Farthest Point Sampling (FPS) \cite{qi2017pointnet++} to preserve partial structure of the original input \cite{wang2020cascaded}.

\subsection{Point Cloud Completion with Patch Seeds}
\label{sec:method:seed}
\textbf{Patch Seeds.} Patch Seeds serve as a new shape representation in point cloud completion. It consists of both seed coordinates $\mathcal{S}$ and features $\mathcal{F}$ where each seed covers a small region around this point with seed feature containing semantic clues about this area. With rich information provided in seed features, SeedFormer can recover ambiguous missing details as shown in Fig.~\ref{fig:teaser}(d). Thus, compared with a global feature, the Patch Seeds representation possesses two advantages: (i) it can preserve \textit{regional information of local patterns}, thus tiny details can be recovered in the complete point cloud; (ii) it represents the \textit{complete shape structure} which is recovered from the partial input with explicit supervision (Sec.~\ref{sec:method:loss}).

\noindent\textbf{Usage.} Throughout the following steps, Patch Seeds are incorporated into each of the upsample layers to provide regional information. Given an input point cloud $\mathcal{P}_l$, we propagate seed features to each point $p_i \in \mathcal{P}_l$ by interpolating feature values in its neighborhood $\mathcal{N}_s(i)$ ($k$ nearest seeds of $p_i$). We use weighted average based on inverse distances \cite{qi2017pointnet++} to compute the interpolated features:
\begin{equation}
    s_i^l = \frac{\sum_{j \in \mathcal{N}_s(i)} \hat{d}_{ij} f_j}{\sum_{j \in \mathcal{N}_s(i)} \hat{d}_{ij}} \quad \text{where} \quad \hat{d}_{ij} = \frac{1}{d_{ij}}.
\end{equation}
Here, $d_{ij}$ denotes the distance between $p_i$ and seed point $x_j$, and $f_j$ is the corresponding seed feature. The interpolated features in this layer are $\{s_i^l\}_{i=1}^{N_l}$.


\subsection{Upsample Transformer}
\label{sec:method:uptrans}
\begin{figure*}[t]
	\begin{center}
		\includegraphics[width=0.7\linewidth]{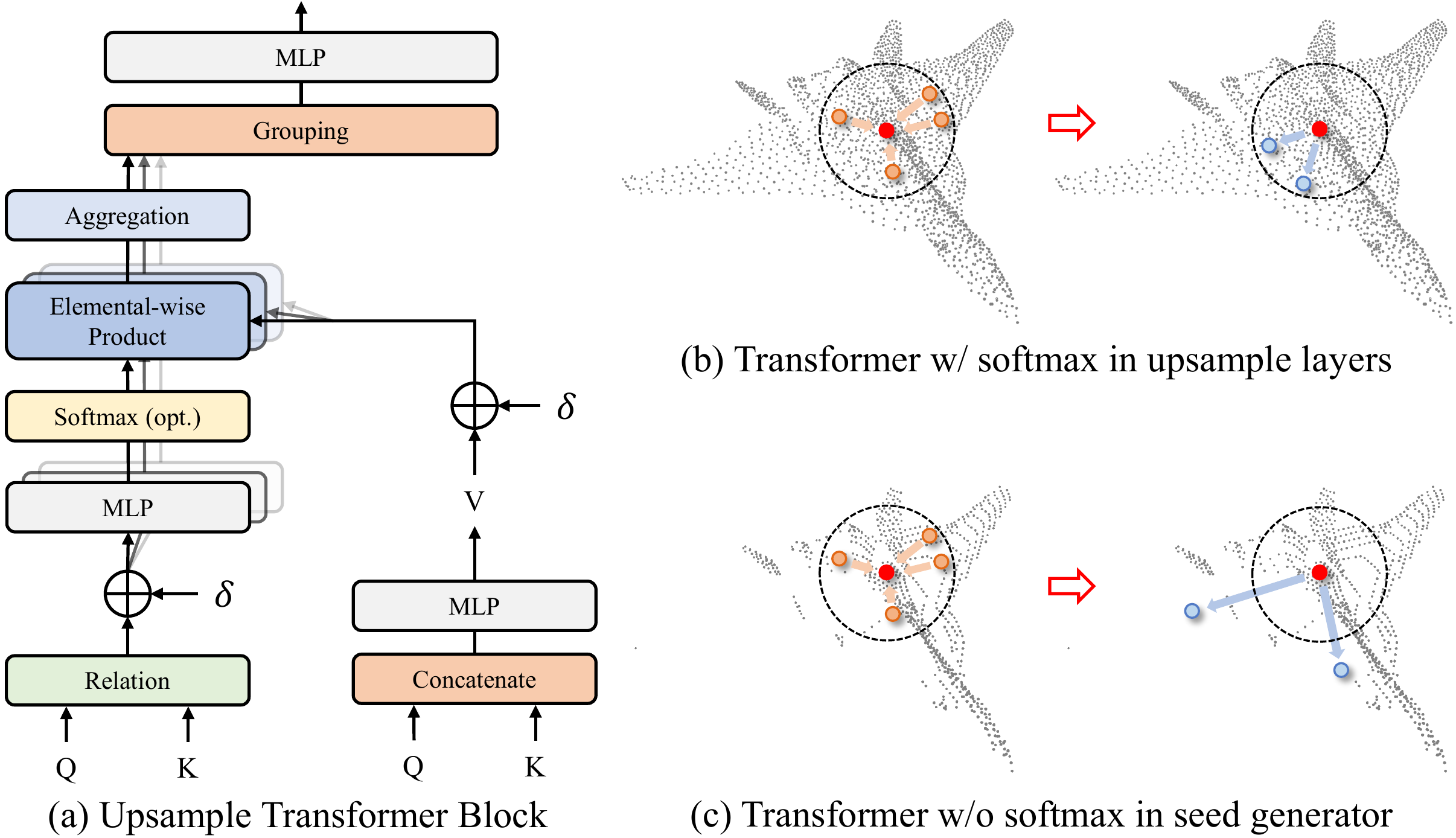} 	
	\end{center}
	\caption{(a) shows the design of our Upsample Transformer. The generation processes of new points around a target point (red) are shown: (b) an upsample layer can produce dense points within its local neighborhood; (c) the seed generator tends to generate seed points covering unseen parts of the object where the softmax function can be optionally disabled.}
	\label{fig:uptrans}
\end{figure*}

\noindent\textbf{Point generator via local aggregation.} 
The generation process in point cloud completion aims to produce a set of new points, either maintaining data fidelity by preserving geometric details or well inferring missing parts based on the existing shape structure. 
The commonly-used folding operations~\cite{yang2018foldingnet,yuan2018pcn,wen2020point,xie2021style} are designed to upsample each point independently with fixed 2D variants which leads to poor detail recovery. 
Differently, our Upsample Transformer is designed to incorporate closely-related local information into the generation by aggregating features from neighboring points. 
To this end, based on the point transformer~\cite{zhao2021point} structure, we propose to formulate the generation process of new points as a self-attention weighted average of point features in the local field (Fig.~\ref{fig:uptrans}(a)).
Moreover, following the idea of local aggregation, we can extend other successful encoder designs, \textit{e.g.}, PointNet++~\cite{qi2017pointnet++}, Transformers~\cite{vaswani2017attention,zhao2021point} or point cloud convolutions~\cite{wang2019dynamic,thomas2019kpconv,zhou2021adaptive}, into point generators in point cloud completion, to produce a new set of features with upsampled points.
This is further evaluated in Sec.~\ref{sec:eval:ablation}.

\noindent\textbf{Upsample layer.} 
We first explain how to apply the designed Upsample Transformer in an upsample layer (Fig.~\ref{fig:architecture}(b)).
Given the input points $\mathcal{P}_l \in \mathbb{R}^{N_l \times 3}$ and the corresponding interpolated seed features $\{s_i^l\}_{i=1}^{N_l}$, we concatenate them and apply a shared MLP to form the point-wise queries $\{q_i^l\}_{i=1}^{N_l}$. Following \cite{xiang2021snowflakenet}, we use the output features from the previous layer as keys $\{k_i^l\}_{i=1}^{N_l}$ in the transformer. It is used to preserve learned features of the existing points in the input.
Then, the values $\{v_i^l\}_{i=1}^{N_l}$ are obtained by applying a MLP on the concatenated keys and queries. 
Upsample Transformer applies a channel-wise attention using the subtraction relation in a local neighborhood $\mathcal{N}(i)$ ($k$ nearest neighbors) of each point:
\begin{equation}
    \hat{a}_{ijm} = \alpha_m(\beta(q_i^l) - \gamma(k_j^l) + \delta), j \in \mathcal{N}(i). \label{equ:att}
\end{equation}
Here, $\alpha_m$, $\beta$ and $\gamma$ are feature mapping functions (\textit{i.e.}, MLPs) to produce attention vectors. $\delta$ is a positional encoding vector to learn spatial relations. For each upsampled point relating to the centered point $p_i$, a specific kernel $\alpha_m$ is defined where $m=1,2,...,r_l$ indicates one of the $r_l$ generation processes in this layer ($r_l N_l = N_{l+1}$). Each kernel learns a certain geometric pattern of local characteristics which outputs a separate group of self-attention weights to form a new point. In addition, to normalize the computed weights into a balanced scale, $\hat{a}_{ijm}$ is applied by a softmax function:
\begin{equation}
    a_{ijm} = \frac{\exp(\hat{a}_{ijm}) }{\sum_{j \in \mathcal{N}(i)} \exp(\hat{a}_{ijm}) }.
\end{equation}
Then, we compute the generated point features by combining weights with the duplicated values:
\begin{equation}
    h_{im} = \sum_{j \in \mathcal{N}(i)} a_{ijm} * (\psi(v_j^l) + \delta),
\end{equation}
where $\psi$ is also a feature mapping function and $*$ denotes the elemental-wise product. Grouping all $h_{im}$ from each kernel yields our produced upsampled point features $\mathcal{H}_l = \{h_{im} | i=1,2,...N_l; m=1,2,...,r_l\} \in \mathbb{R}^{r_l N_l \times C}$. $\mathcal{H}_l$ also serves as the keys $\{k_i^{l+1}\}_{i=1}^{N_{l+1}}$ of next layer in the skip connection. Finally, we apply shared MLPs on point features to obtain a set of point displacement offsets $\Delta \mathcal{P}_l$. The output point cloud is defined as $\mathcal{P}_{l+1} = \hat{\mathcal{P}}_l + \Delta \mathcal{P}_l$, where $\hat{\mathcal{P}}_l$ is the duplicated point cloud, for both refining existing points and generating new points.

\noindent\textbf{Positional encoding with seed features.} The basic positional encoding in the self-attention computations is designed to capture spatial relations between 3D points~\cite{zhao2021point}. Besides, since the introduced seeds contain regional features with regard to the seed positions, we encode the interpolated features into \emph{regional} encoding as follows:
\begin{equation}
    \delta = \rho(p_i - p_j) + \theta(s_i - s_j).
\end{equation}
Here, $\delta$ encodes both positional relation and seed feature relation between $p_i$ and $p_j$. $\rho$ and $\theta$ are encoding functions using a two-layer MLP.

\noindent\textbf{Transformer without softmax.} Similarly, Upsample Transformer is also used in the seed generator but is implemented without skip connection and seed features (Fig.~\ref{fig:architecture}(a)).
Generating seeds involves a different objective of producing seed points which can predict missing regions and cover the complete shape structure. 
As shown in Fig.~\ref{fig:uptrans}(c), when the input point cloud represents an incomplete shape, it is essential for the seed generator to produce seed points outside the local neighborhood. However, the standard transformer structure represents an intrinsic limitation that the softmax normalization explicitly produces attention weights within a specific range of $(0,1)$.
This may limit the learning ability especially in the seed generator.
To solve this, we simply remove the softmax function in the transformer; that is, $a_{ijm} = \hat{a}_{ijm}$ is applied without normalization. 
The experiments in Sec.~\ref{sec:eval:ablation} show that it is easier to generate better seed points by disabling the softmax function or using other alternatives.

\subsection{Loss Function}
\label{sec:method:loss}
We use Chamfer Distance (CD) as our loss function to measure the distance between two unordered point sets~\cite{fan2017point}. To ensure that the generated seeds $\mathcal{S}$ can cover the complete shape, we down-sample the ground-truth point cloud to the same point number as $\mathcal{S}$ and compute the corresponding CD loss between them. The same loss function is also applied to each output $\mathcal{P}_l$ separately in the upsample layers. Then, we define the sum of all CDs as the completion loss $\mathcal{L}_{comp}$. Besides, following \cite{wen2021cycle4completion}, we compute the partial matching loss $\mathcal{L}_{part}$ on the final predicted result to preserve the shape structure of the input point cloud. The total training loss is defined as:
\begin{equation}
    \mathcal{L} = \mathcal{L}_{comp} + \mathcal{L}_{part}.
\end{equation}
\label{sec:method}

\section{Evaluation}
\subsection{Implementation Details}
The SeedFormer encoder applies two layers of set abstraction \cite{qi2017pointnet++} and obtain $N_p = 128$ patches of the incomplete point cloud. Then, in the seed generator, we produce a set of seed features $\mathcal{F} \in \mathbb{R}^{N_s \times C_s}$ where $N_s=256$ and $C_s=128$. The coarse point cloud $\mathcal{P}_0$ contains $N_0=512$ points which is obtained by merging $\mathcal{S}$ and $\mathcal{P}$ using FPS. Three upsample layers are used in the following coarse-to-fine generation procedure which output dense point clouds $\{\mathcal{P}_1, \mathcal{P}_2, \mathcal{P}_3\}$ and $\mathcal{P}_3$ corresponds to the final predicted result. Channel number of generated features is set to $C=128$ for all upsample layers.

We train our network end-to-end using pytorch~\cite{paszke2019pytorch} implementation. We use Adam~\cite{kingma2014adam} optimizer with $\beta_1=0.9$ and $\beta_2=0.999$. All the models are trained with a batch size of 48 on two NVIDIA TITAN Xp GPUs. The initial learning rate is set to 0.001 with continuous decay of 0.1 for every 100 epochs.

\begin{table*}[t]
	\centering
	\footnotesize
	\setlength{\tabcolsep}{2.5pt}
	\caption{Completion results on PCN dataset in terms of per-point L1 Chamfer Distance $\times 1000$ (lower is better).}
	\begin{tabular}{c|c|cccccccc}
		\toprule[1pt]
		Methods & Average & Plane & Cabinet & Car & Chair & Lamp & Couch & Table & Boat  \\
		\midrule[0.3pt]
		FoldingNet \cite{yang2018foldingnet} & 14.31 & 9.49 & 15.80 & 12.61 & 15.55 & 16.41 & 15.97 & 13.65 & 14.99 \\
		TopNet \cite{tchapmi2019topnet} & 12.15 & 7.61 & 13.31 & 10.90 & 13.82 & 14.44 & 14.78 & 11.22 & 11.12 \\
		AtlasNet \cite{groueix2018papier} & 10.85 & 6.37 & 11.94 & 10.10 & 12.06 & 12.37 & 12.99 & 10.33 & 10.61 \\
		PCN \cite{yuan2018pcn} & 9.64 & 5.50 & 22.70 & 10.63 & 8.70 & 11.00 & 11.34 & 11.68 & 8.59 \\
		GRNet \cite{xie2020grnet} & 8.83 & 6.45 & 10.37 & 9.45 & 9.41 & 7.96 & 10.51 & 8.44 & 8.04 \\
		CRN \cite{wang2020cascaded} & 8.51 & 4.79 & 9.97 & 8.31 & 9.49 & 8.94 & 10.69 & 7.81 & 8.05 \\
		NSFA \cite{zhang2020detail} & 8.06 & 4.76 & 10.18 & 8.63 & 8.53 & 7.03 & 10.53 & 7.35 & 7.48 \\
		PMP-Net \cite{wen2021pmp} & 8.73 & 5.65 & 11.24 & 9.64 & 9.51 & 6.95 & 10.83 & 8.72 & 7.25 \\
		PoinTr \cite{yu2021pointr} & 8.38 & 4.75 & 10.47 & 8.68 & 9.39 & 7.75 & 10.93 & 7.78 & 7.29 \\
		SnowflakeNet \cite{xiang2021snowflakenet} & 7.21 & 4.29 & 9.16 & 8.08 & 7.89 & 6.07 & 9.23 & 6.55 & 6.40 \\
		\midrule[0.3pt]
		SeedFormer & \textbf{6.74} & \textbf{3.85} & \textbf{9.05} & \textbf{8.06} & \textbf{7.06} & \textbf{5.21} & \textbf{8.85} & \textbf{6.05} & \textbf{5.85} \\
		\bottomrule[1pt]
	\end{tabular}

	\label{table:pcn}
\end{table*}

\begin{figure*}[t]
	\newlength{\unit}
	\setlength{\unit}{0.165\linewidth}
	\newlength{\unity}
	\setlength{\unity}{0.11\linewidth}
	\newlength{\unitx}
	\setlength{\unitx}{0.08\linewidth}
	\newlength{\unita}
	\setlength{\unita}{0.135\linewidth}
	
	\centering
	
	\begin{subfigure}{\unit}
		\centering
		\includegraphics[width=\unity]{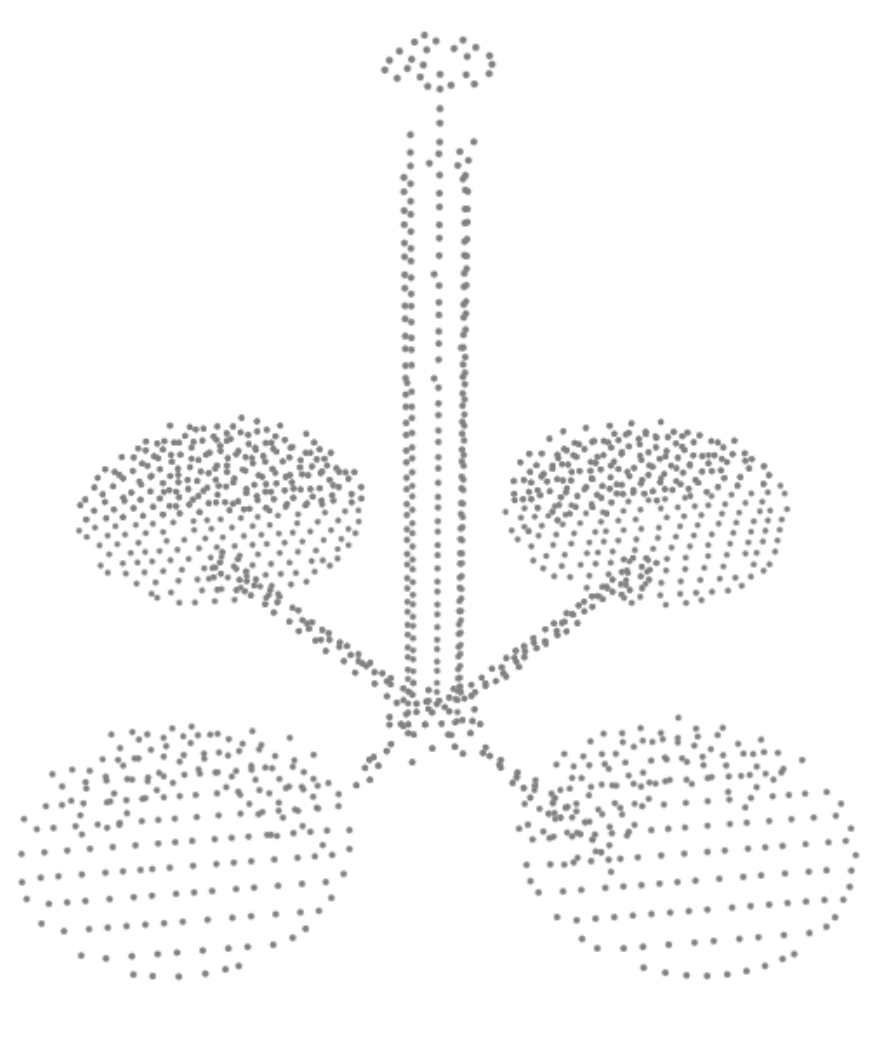}
	\end{subfigure}\hfill%
	\begin{subfigure}{\unit}
		\centering
		\includegraphics[width=\unity]{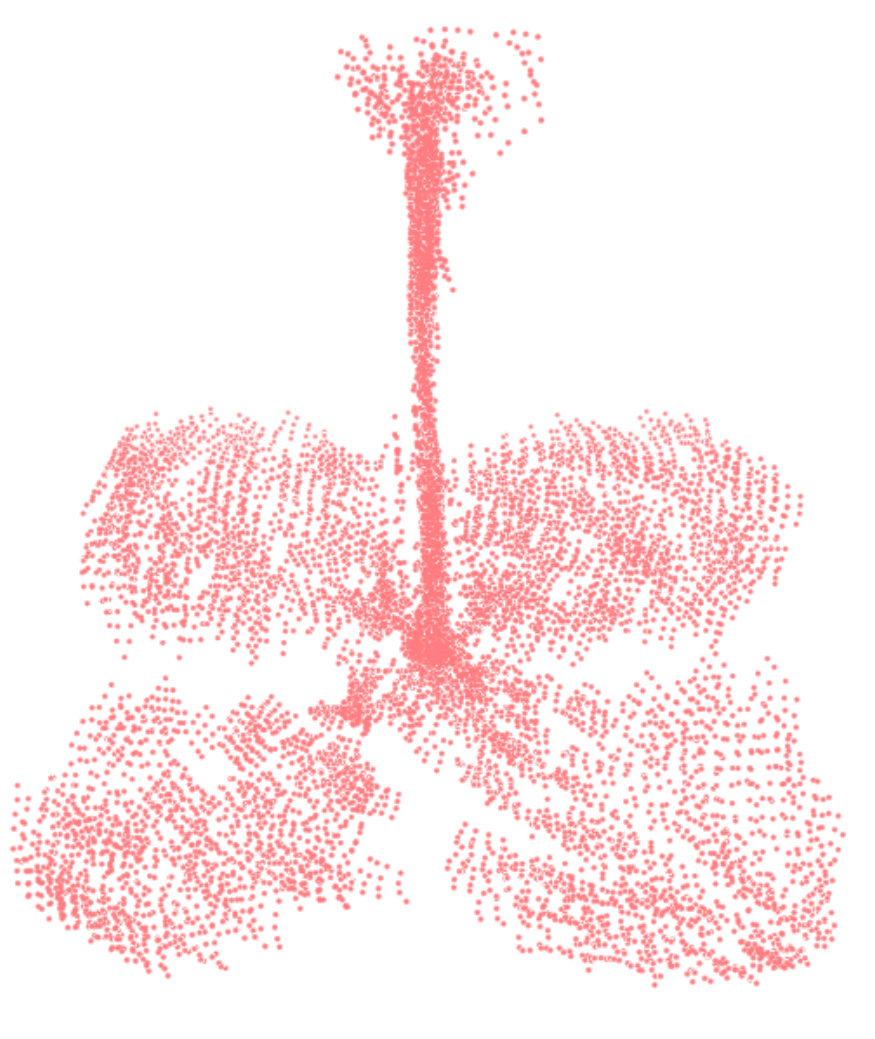}
	\end{subfigure}\hfill%
	\begin{subfigure}{\unit}
		\centering
		\includegraphics[width=\unity]{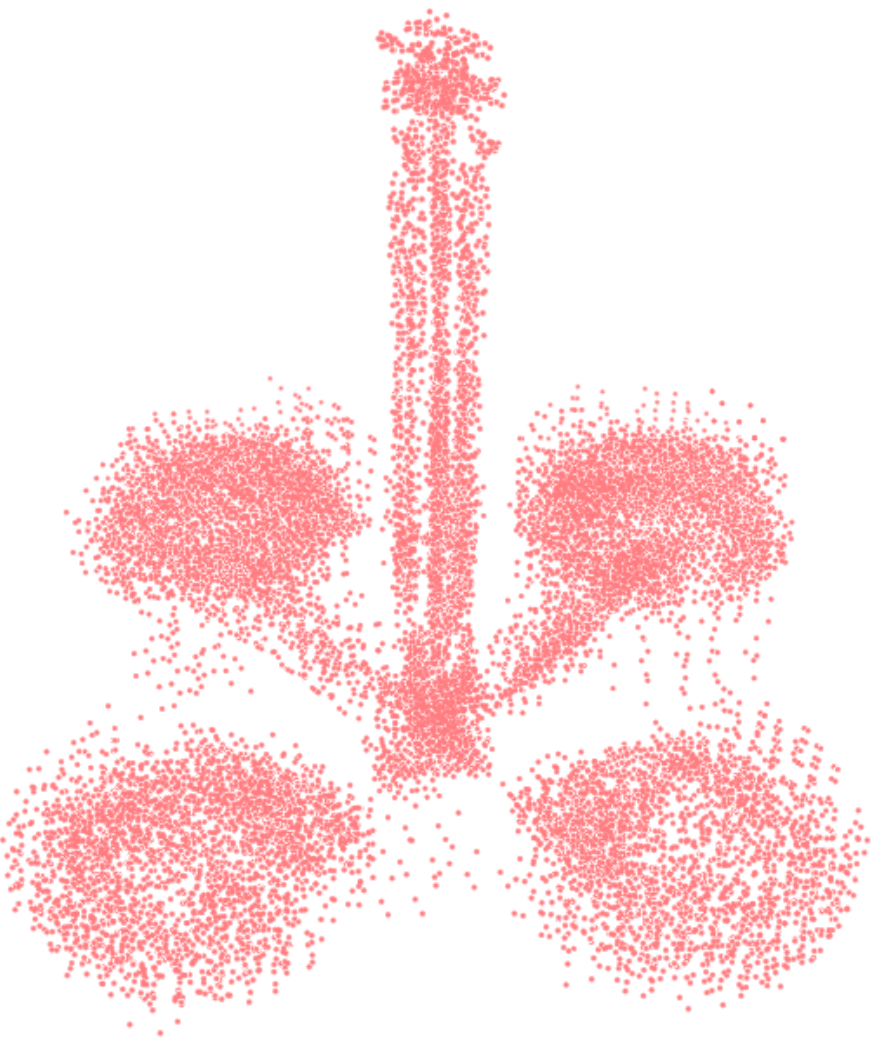}
	\end{subfigure}\hfill%
	\begin{subfigure}{\unit}
		\centering
		\includegraphics[width=\unity]{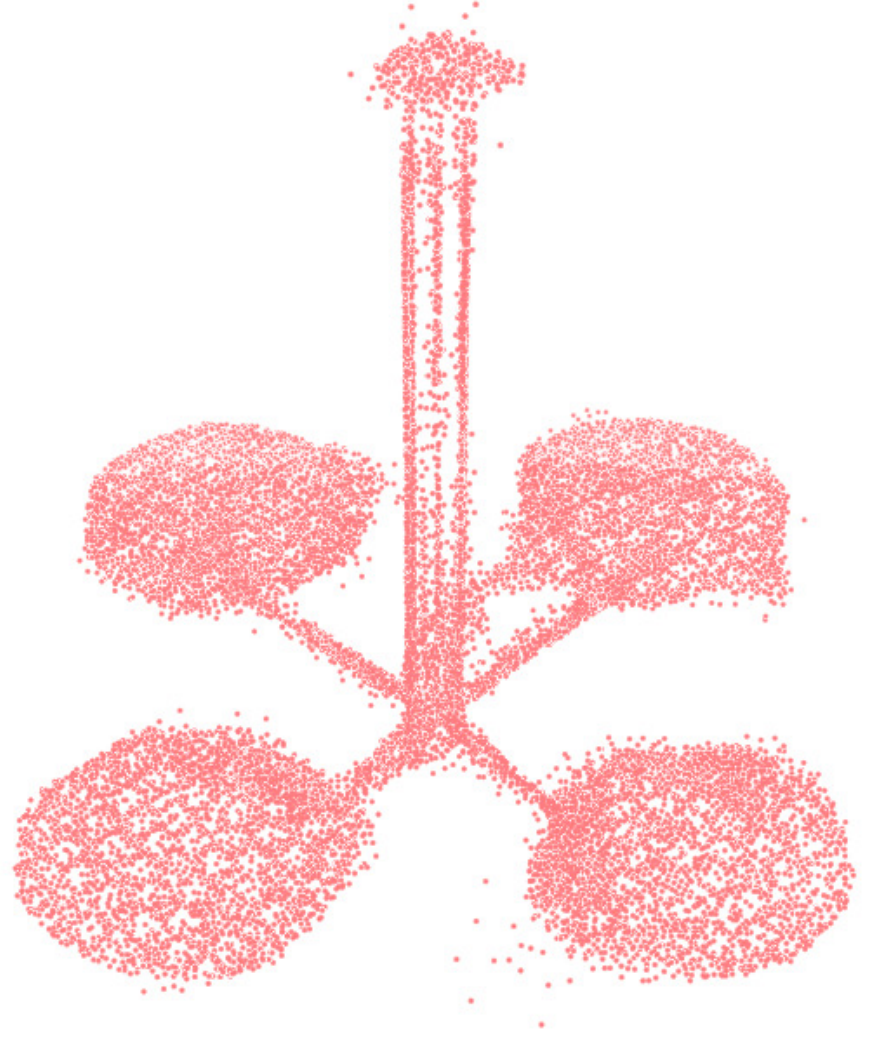}
	\end{subfigure}\hfill%
	\begin{subfigure}{\unit}
		\centering
		\includegraphics[width=\unity]{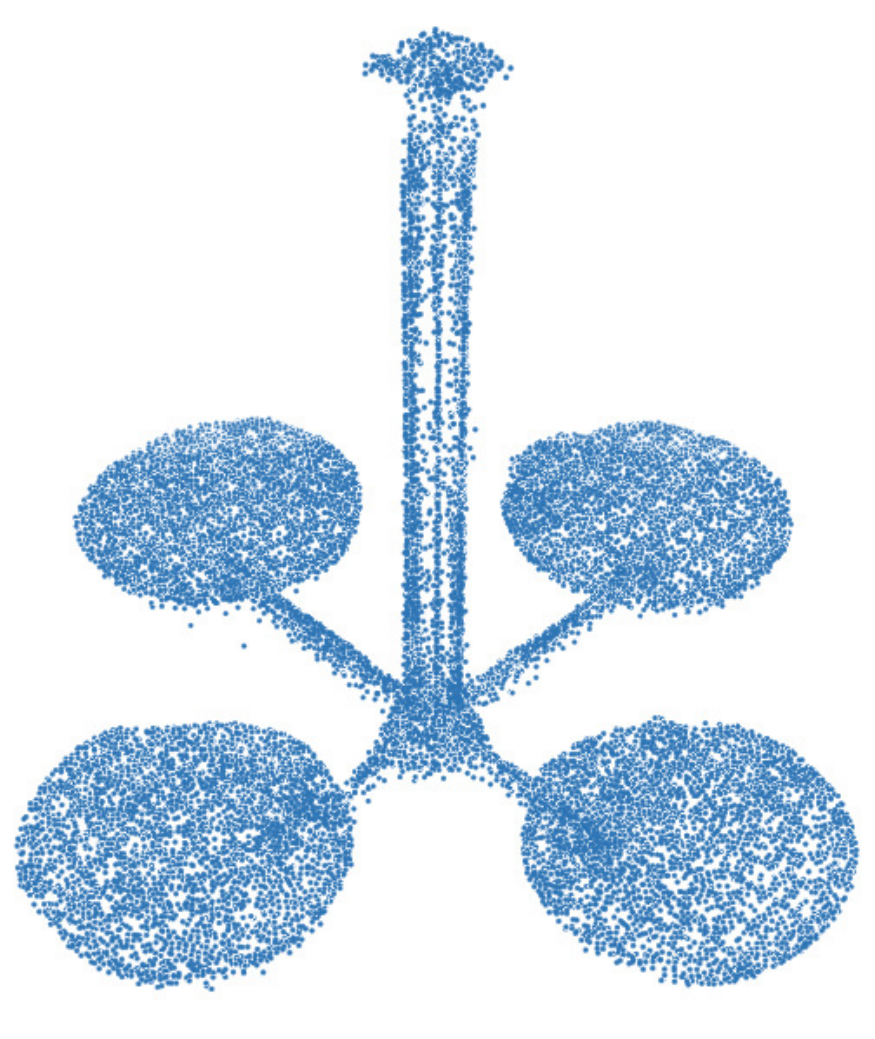}
	\end{subfigure}\hfill%
	\begin{subfigure}{\unit}
		\centering
		\includegraphics[width=\unity]{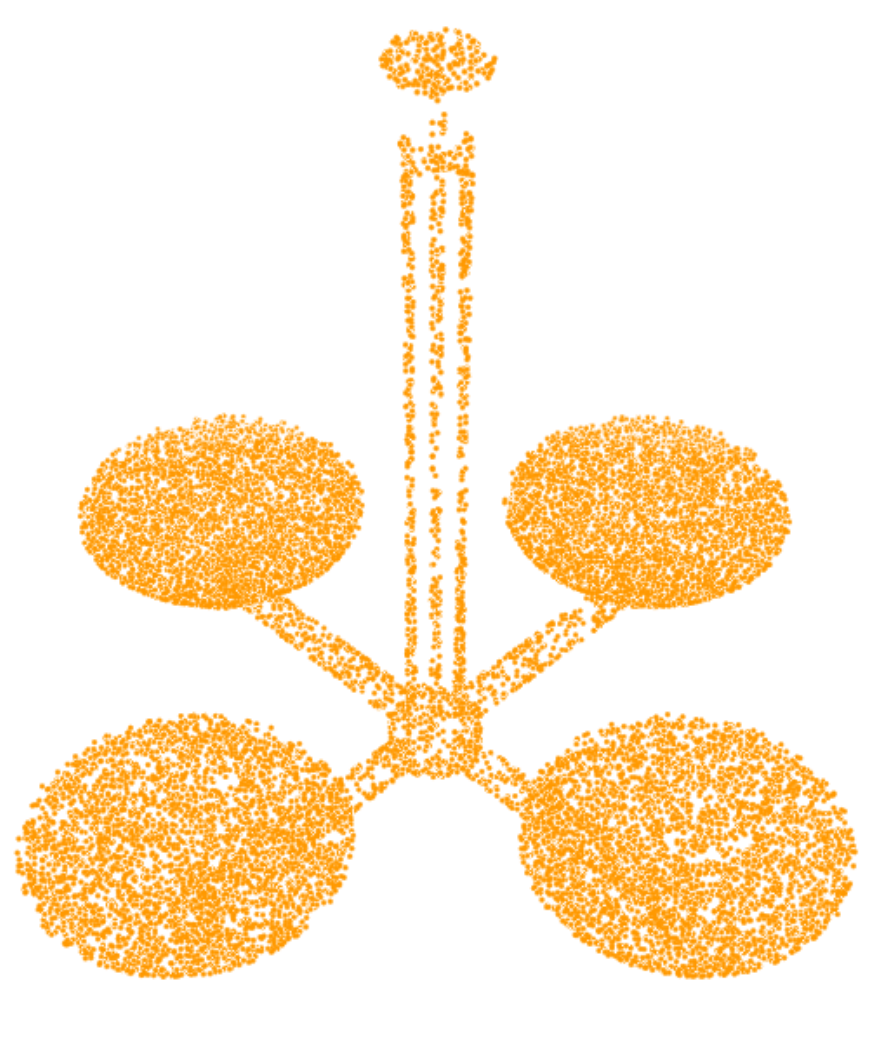}
	\end{subfigure}

	\begin{subfigure}{\unit}
		\centering
		\includegraphics[width=\unitx]{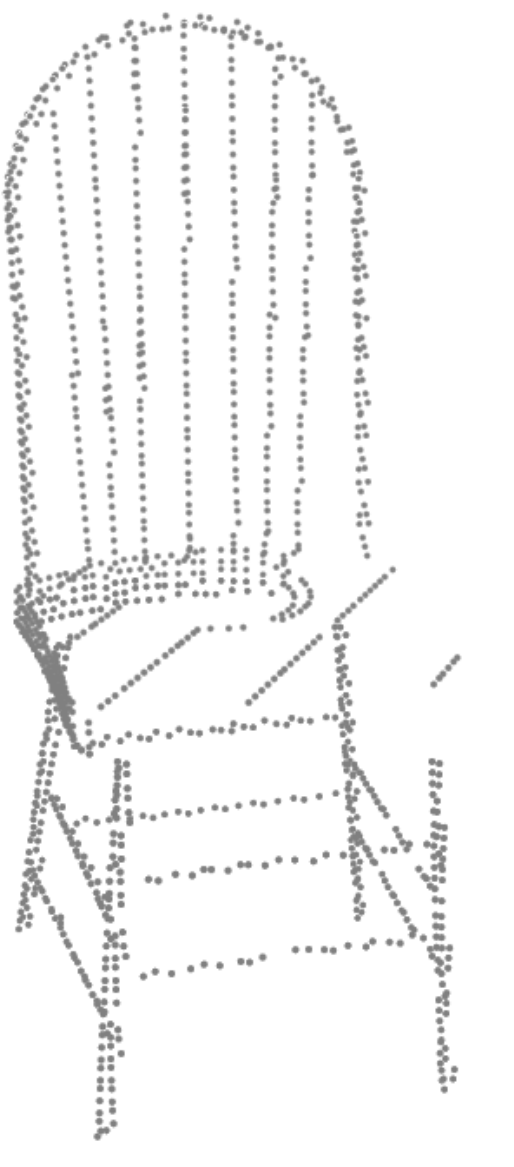}
	\end{subfigure}\hfill%
	\begin{subfigure}{\unit}
		\centering
		\includegraphics[width=\unitx]{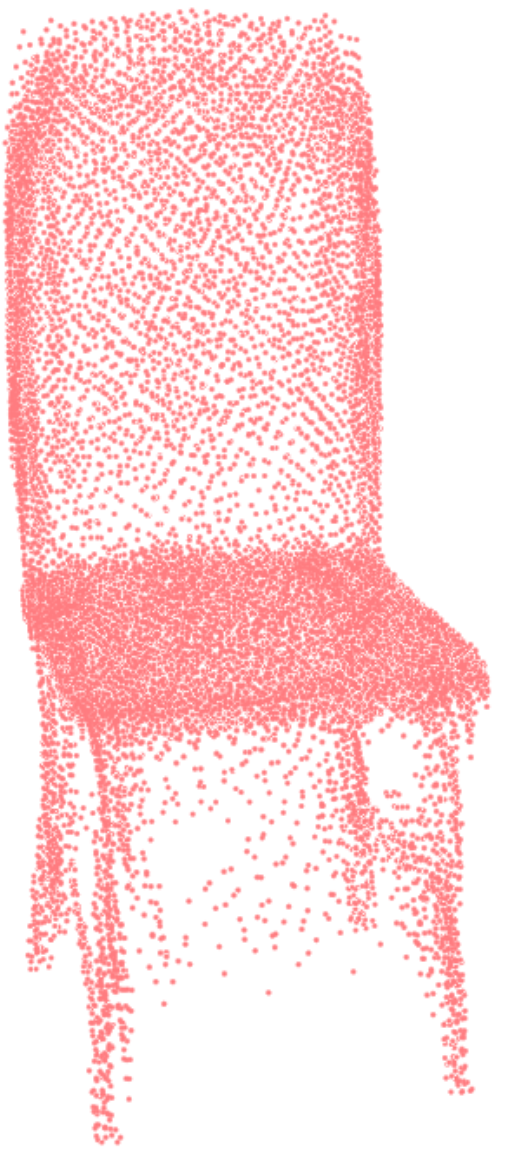}
	\end{subfigure}\hfill%
	\begin{subfigure}{\unit}
		\centering
		\includegraphics[width=\unitx]{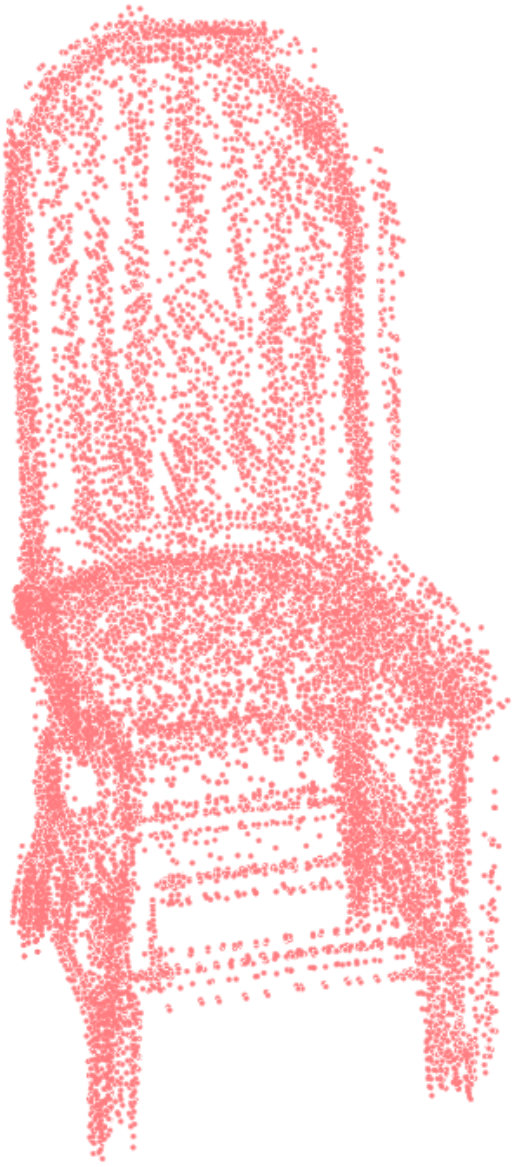}
	\end{subfigure}\hfill%
	\begin{subfigure}{\unit}
		\centering
		\includegraphics[width=\unitx]{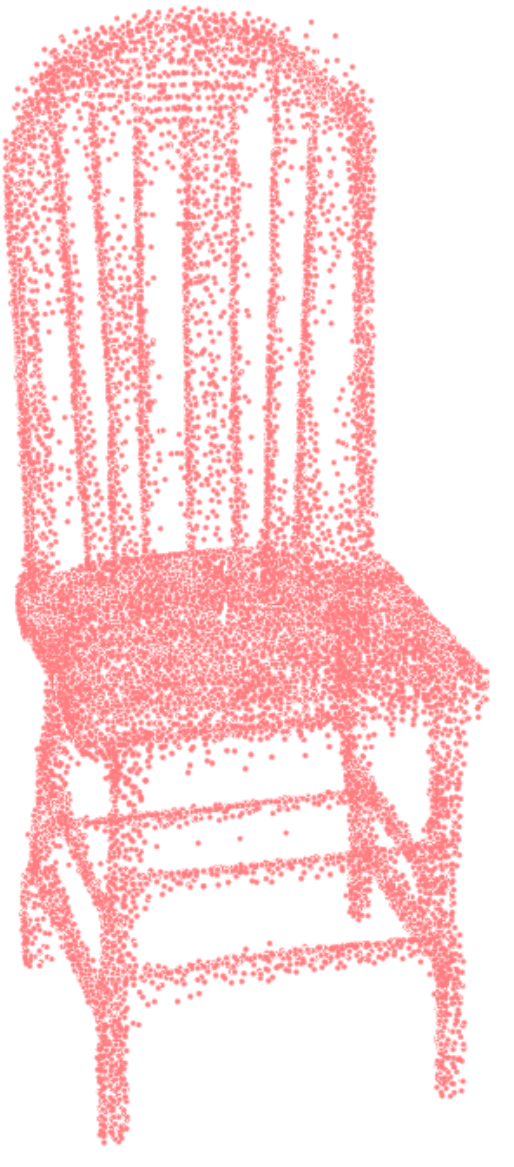}
	\end{subfigure}\hfill%
	\begin{subfigure}{\unit}
		\centering
		\includegraphics[width=\unitx]{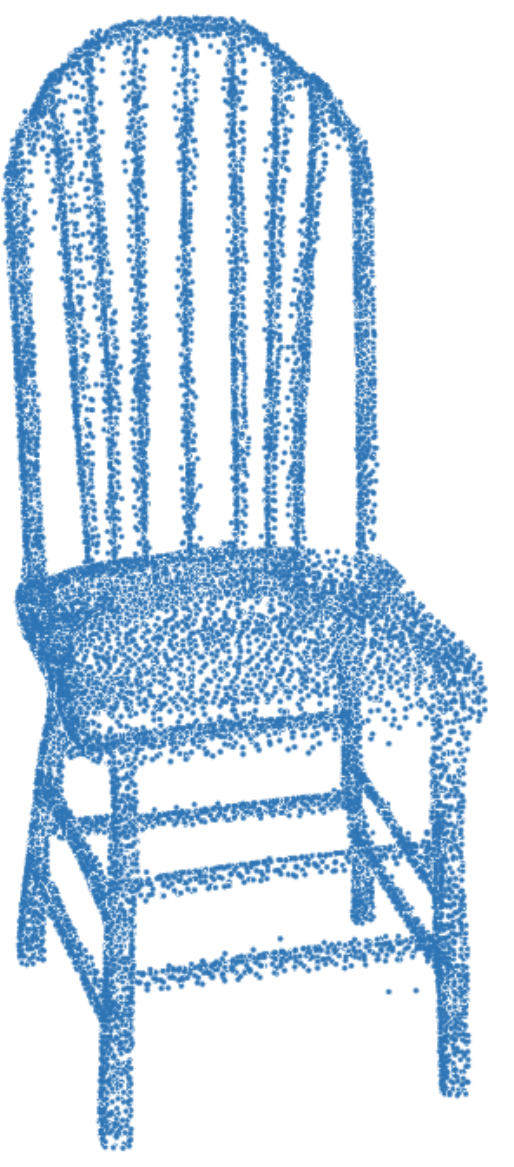}
	\end{subfigure}\hfill%
	\begin{subfigure}{\unit}
		\centering
		\includegraphics[width=\unitx]{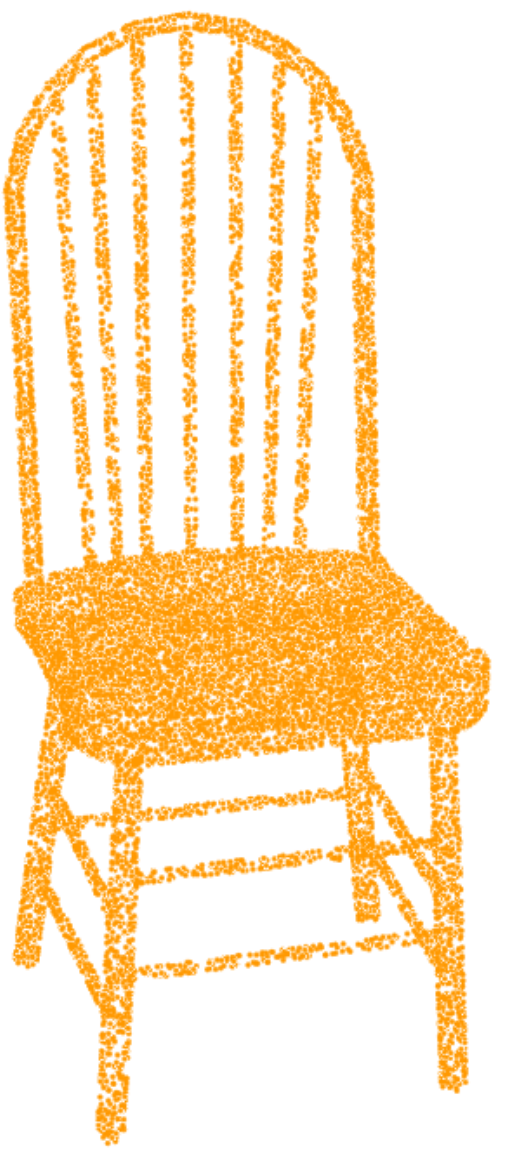}
	\end{subfigure}

	\captionsetup[subfigure]{font=small,labelfont=small}
	\begin{subfigure}{\unit}
		\centering
		\includegraphics[width=\unita]{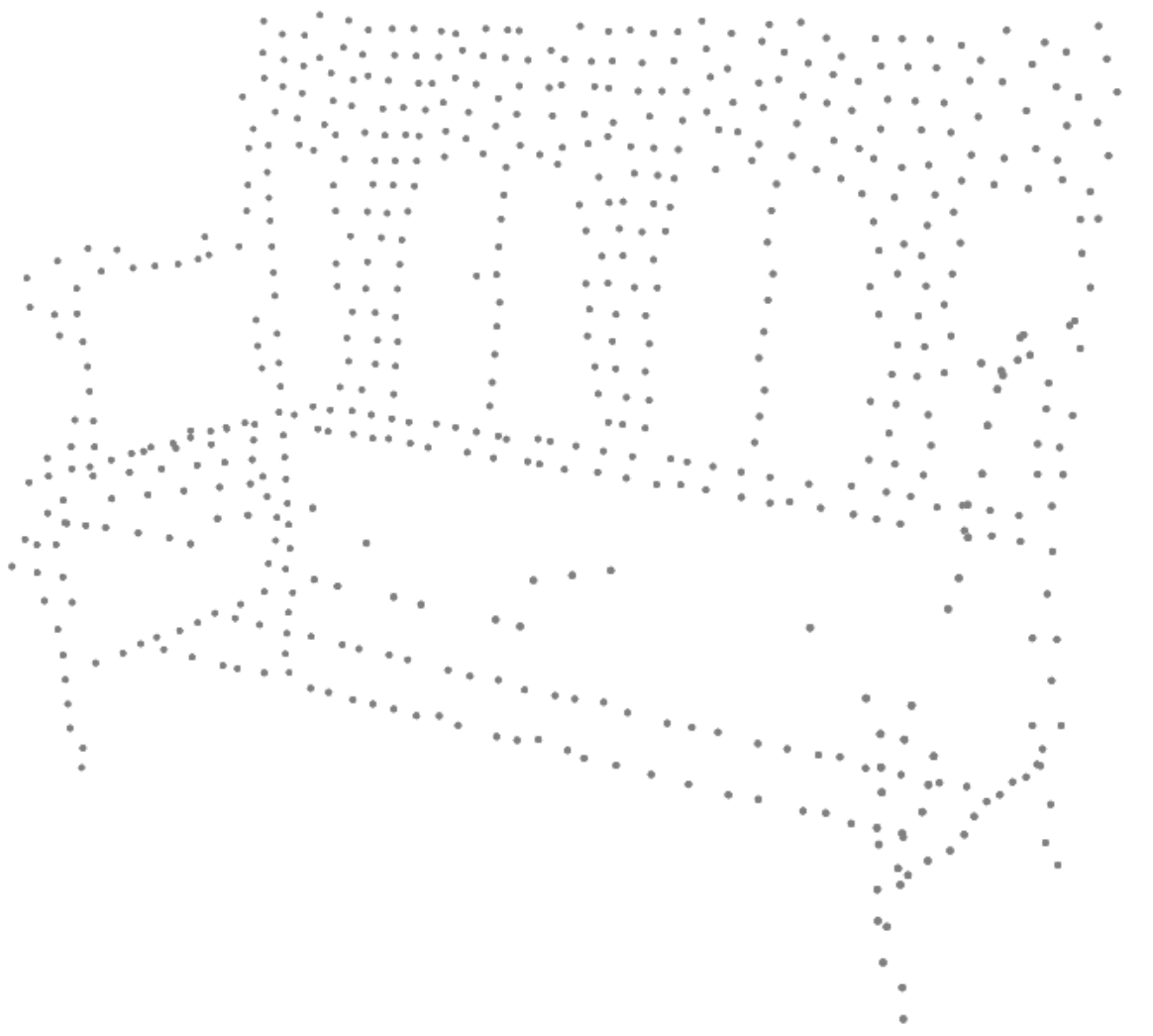}
		\caption{Input}
	\end{subfigure}\hfill%
	\begin{subfigure}{\unit}
		\centering
		\includegraphics[width=\unita]{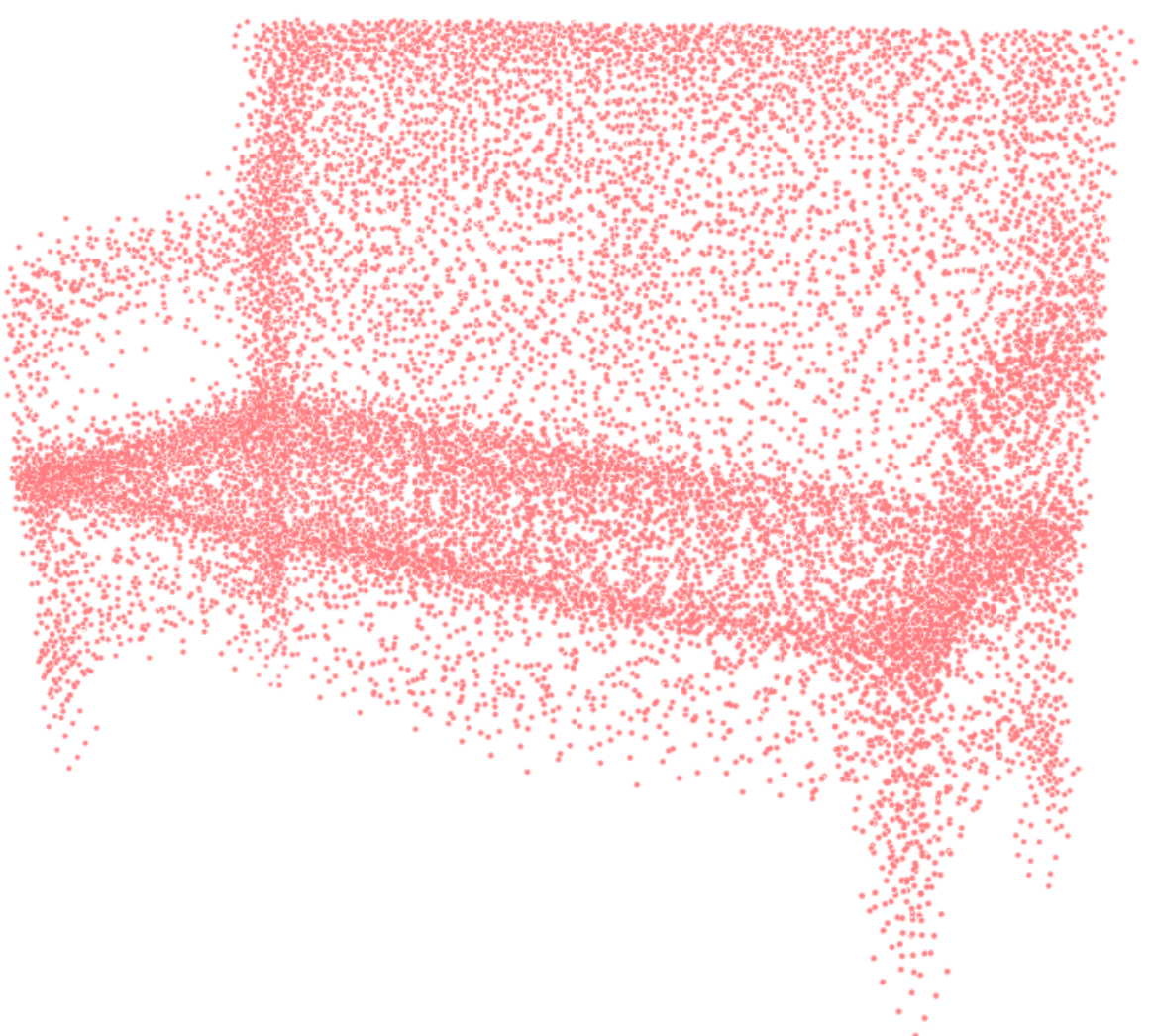}
		\caption{PCN}
	\end{subfigure}\hfill%
	\begin{subfigure}{\unit}
		\centering
		\includegraphics[width=\unita]{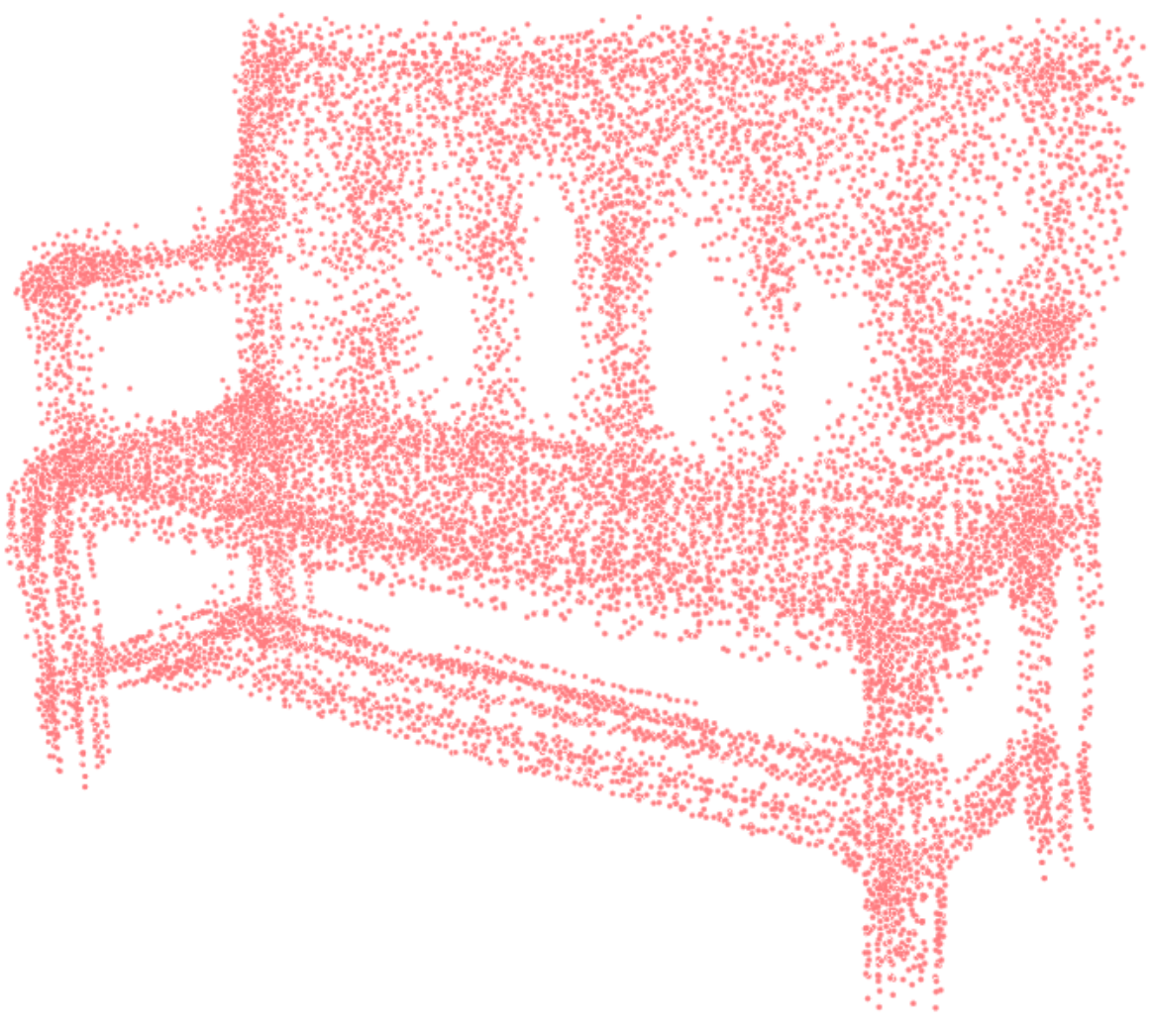}
		\caption{GRNet}
	\end{subfigure}\hfill%
	\begin{subfigure}{\unit}
		\centering
		\includegraphics[width=\unita]{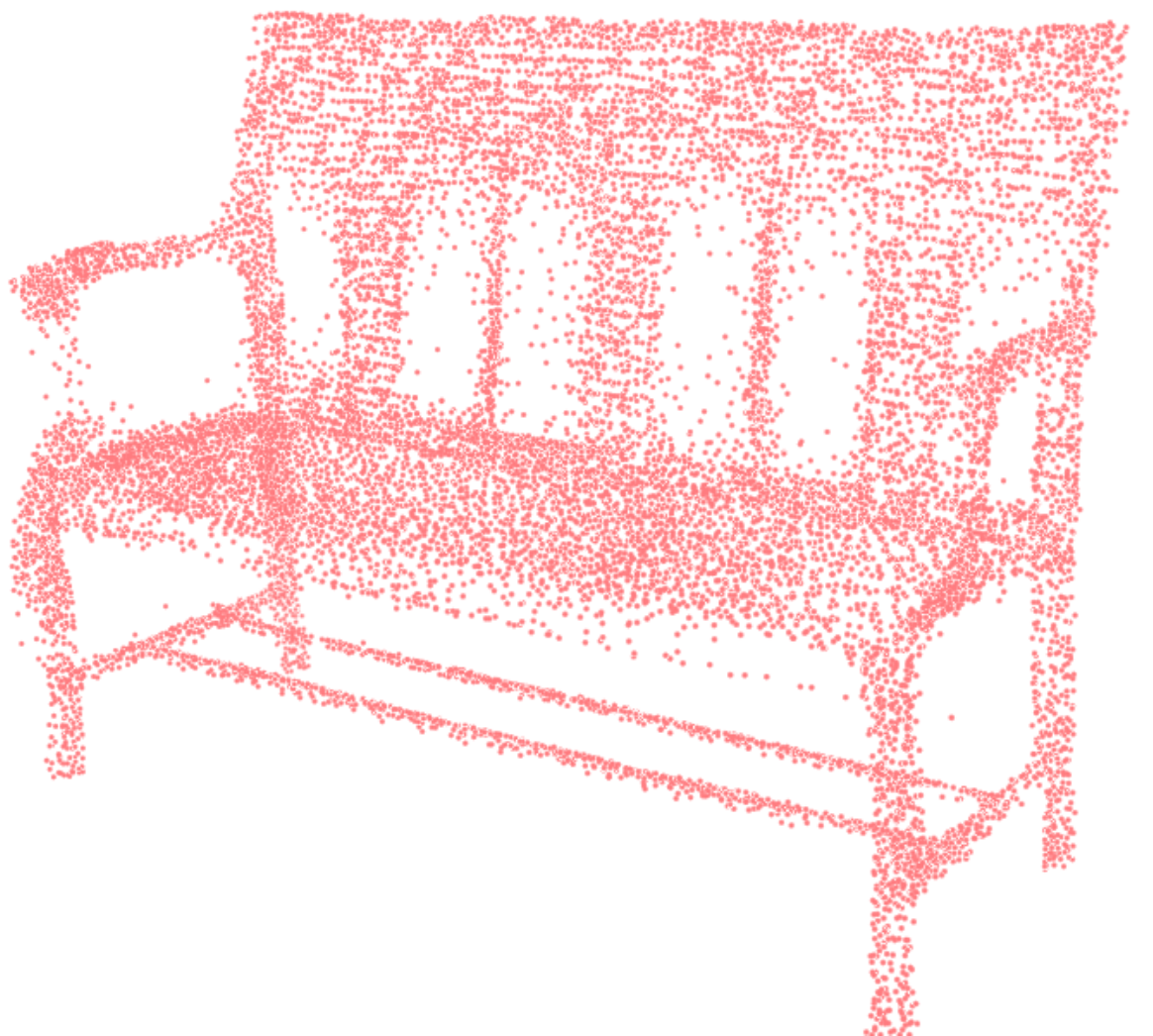}
		\caption{Snowflake}
	\end{subfigure}\hfill%
	\begin{subfigure}{\unit}
		\centering
		\includegraphics[width=\unita]{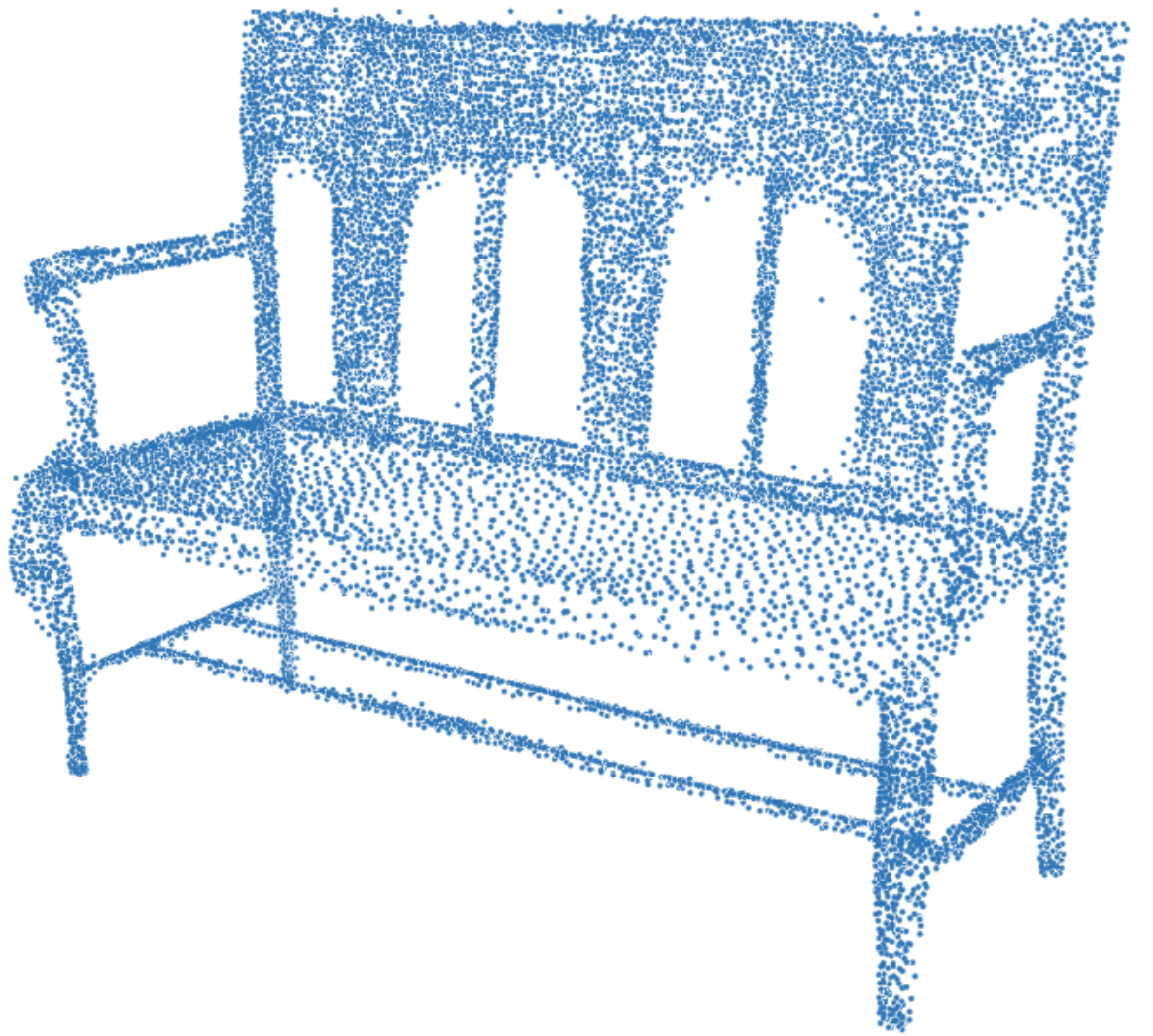}
		\caption{Ours}
	\end{subfigure}\hfill%
	\begin{subfigure}{\unit}
		\centering
		\includegraphics[width=\unita]{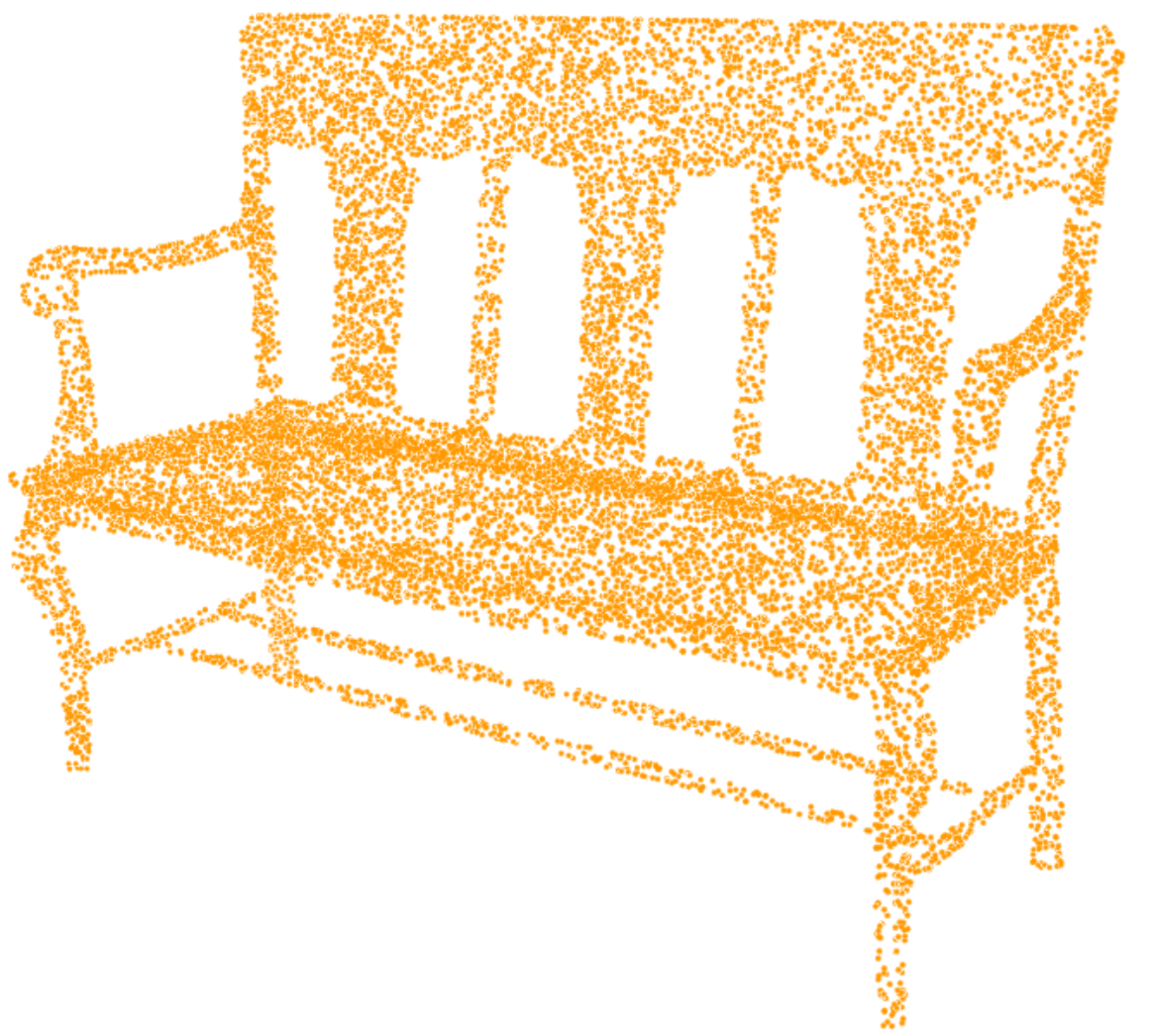}
		\caption{GT}
	\end{subfigure}
	
	\caption{Visual comparisons on PCN dataset.}
	\label{fig:pcn}
\end{figure*}

\subsection{Experiments on PCN Dataset}
\label{sec:eval:pcn}
\noindent\textbf{Data.} The PCN dataset~\cite{yuan2018pcn} is one of the most widely used benchmark datasets for point cloud completion. It is a subset of ShapeNet~\cite{chang2015shapenet} with shapes from 8 categories. The incomplete point clouds are generated by back-projecting 2.5D depth images from 8 viewpoints in order to simulate real-world sensor data. For each shape, 16,384 points are uniformly sampled from the mesh surfaces as complete ground-truth, and 2,048 points are sampled as partial input. We follow the same experimental setting with PCN for a fair comparison.

\noindent\textbf{Results.} Following previous methods, we report the Chamfer Distances with L1 norm ($\times 1000$) in Tab.~\ref{table:pcn}. Detailed results of each category are also provided. SeedFormer achieves the best scores on all categories of this dataset, outperforming previous state-of-the-art methods by a large amount. Moreover, in Fig.~\ref{fig:pcn}, we show visual results of shapes from three categories (Lamp, Chair and Couch), compared with PCN~\cite{yuan2018pcn}, GRNet~\cite{xie2020grnet} and SnowflakeNet~\cite{xiang2021snowflakenet}. It shows that SeedFormer can produce clearly better results with more faithful details. 
As shown in the 2-nd and 3-rd rows, our network can preserve complicated details in the regions of chair arms and backs, without introducing undesired components.

\begin{table*}[t]
	\centering
	\footnotesize
	\setlength{\tabcolsep}{2.2pt}
	\caption{Completion results on ShapeNet-55 dataset evaluated as L2 Chamfer Distance $\times 1000$ (lower is better) and F-Score@1$\%$ (higher is better).}
	\begin{tabular}{c|ccccc|ccc|cc}
		\toprule[1pt]
		Methods & Table & Chair & Plane & Car & Sofa & CD-S & CD-M & CD-H & CD-Avg & F1  \\
		\midrule[0.3pt]
		FoldingNet \cite{yang2018foldingnet}  	& 2.53 & 2.81 & 1.43 & 1.98 & 2.48 & 2.67 & 2.66 & 4.05 & 3.12 & 0.082 \\
        PCN \cite{yuan2018pcn}  				& 2.13 & 2.29 & 1.02 & 1.85 & 2.06 & 1.94 & 1.96 & 4.08 & 2.66 & 0.133 \\
        TopNet \cite{tchapmi2019topnet}  		& 2.21 & 2.53 & 1.14 & 2.18 & 2.36 & 2.26 & 2.16 & 4.3 & 2.91 & 0.126  \\
        PFNet \cite{huang2020pf}  				& 3.95 & 4.24 & 1.81 & 2.53 & 3.34 & 3.83 & 3.87 & 7.97 & 5.22 & 0.339 \\ 
        GRNet \cite{xie2020grnet}  				& 1.63 & 1.88 & 1.02 & 1.64 & 1.72 & 1.35 & 1.71 & 2.85 & 1.97 & 0.238 \\
        PoinTr \cite{yu2021pointr}  			& 0.81 & 0.95 & 0.44 & 0.91 & 0.79 & 0.58 & 0.88 & 1.79 & 1.09 & 0.464 \\
		\midrule[0.3pt]
		SeedFormer & \textbf{0.72} & \textbf{0.81} & \textbf{0.40} & \textbf{0.89} & \textbf{0.71} & \textbf{0.50} & \textbf{0.77} & \textbf{1.49} & \textbf{0.92} & \textbf{0.472} \\
		\bottomrule[1pt]
	\end{tabular}

	\label{table:shapenet55}
\end{table*}

\begin{table*}[t]
	\centering
	\footnotesize
	\setlength{\tabcolsep}{2.5pt}
	\caption{Completion results on ShapeNet-34 dataset evaluated as L2 Chamfer Distance $\times 1000$ (lower is better) and F-Score@1$\%$ (higher is better).}
	\begin{tabular}{c|ccccc|ccccc}
		\toprule[1pt]
		Methods & \multicolumn{5}{c|}{34 seen categories} & \multicolumn{5}{c}{21 unseen categories} \\
		        & CD-S & CD-M & CD-H & Avg & F1 & CD-S & CD-M & CD-H & Avg & F1 \\
		\midrule[0.3pt]
		FoldingNet \cite{yang2018foldingnet}  	 & 1.86 & 1.81 & 3.38 & 2.35 & 0.139 & 2.76 & 2.74 & 5.36 & 3.62 & 0.095 \\
        PCN \cite{yuan2018pcn}  				 & 1.87 & 1.81 & 2.97 & 2.22 & 0.154 & 3.17 & 3.08 & 5.29 & 3.85 & 0.101 \\
        TopNet \cite{tchapmi2019topnet}  		 & 1.77 & 1.61 & 3.54 & 2.31 & 0.171 & 2.62 & 2.43 & 5.44 & 3.50 & 0.121 \\ 
        PFNet \cite{huang2020pf}  				 & 3.16 & 3.19 & 7.71 & 4.68 & 0.347 & 5.29 & 5.87 & 13.33 & 8.16 & 0.322 \\
        GRNet \cite{xie2020grnet}  				 & 1.26 & 1.39 & 2.57 & 1.74 & 0.251 & 1.85 & 2.25 & 4.87 & 2.99 & 0.216 \\
        PoinTr \cite{yu2021pointr}   			 & 0.76 & 1.05 & 1.88 & 1.23 & 0.421 & 1.04 & 1.67 & 3.44 & 2.05 & 0.384 \\
		\midrule[0.3pt]
		SeedFormer & \textbf{0.48} & \textbf{0.70} & \textbf{1.30} & \textbf{0.83} & \textbf{0.452} & \textbf{0.61} & \textbf{1.07} & \textbf{2.35} & \textbf{1.34} & \textbf{0.402} \\
		\bottomrule[1pt]
	\end{tabular}

	\label{table:shapenet34}
\end{table*}

\subsection{Experiments on ShapeNet-55/34}
\noindent\textbf{Data.} We further evaluate our model on ShapeNet-55 and ShapeNet-34 datasets from~\cite{yu2021pointr}. These two datasets are also generated from the synthetic ShapeNet~\cite{chang2015shapenet} dataset while they contain more object categories and incomplete patterns. All 55 categories in ShapeNet are included in ShapeNet-55 with 41,952 shapes for training and 10,518 shapes for testing. ShapeNet-34 uses a subset of 34 categories for training and leaves 21 unseen categories for testing where 46,765 object shapes are used for training, 3,400 for testing on seen categories and 2,305 for testing on novel (unseen) categories. In both datasets, 2,048 points are sampled as input and 8,192 points as ground-truth.  Following the same evaluation strategy with~\cite{yu2021pointr}, 8 fixed viewpoints are selected and the number of points in partial point cloud is set to 2,048, 4,096 or 6,144 (25$\%$, 50$\%$ or 75$\%$ of the complete point cloud) which corresponds to three difficulty levels of \emph{simple}, \emph{moderate} and \emph{hard} in the test stage.

\noindent\textbf{Results.} The ShapeNet-55 dataset tests the ability of dealing with more diverse objects and incompleteness levels. Tab.~\ref{table:shapenet55} reports the average L2 Chamfer Distances ($\times 1000$) on three difficulty levels and the overall CDs. Additionally, we show results from 5 categories (Table, Chair, Plane, Car and Sofa) with more than 2,500 samples in the training set. Complete results for all 55 categories are available in the supplemental material. We also provide results under the F-Score@1$\%$ metric~\cite{tatarchenko2019single}. Compared with previous methods, SeedFormer achieves the best scores on all categories and evaluation metrics. In particular, our method outperforms the SOTA model PoinTr~\cite{yu2021pointr} by $15.6\%$ in terms of overall CD and $16.8\%$ in terms of average CD on hard difficulty.

On ShapeNet-34, the networks should handle novel objects from unseen categories which do not appear in the training phase. We show results on two test sets in Tab.~\ref{table:shapenet34}. Our method again achieves the best scores. Especially, when dealing with unseen objects, SeedFormer shows better generalization ability achieving an average CD of 1.34 which is 34.6$\%$ lower than PoinTr.

\begin{table*}[t]
	\centering
	\footnotesize
	\setlength{\tabcolsep}{2pt}
	\caption{Completion results on KITTI dataset evaluated as Fidelity Distance and Minimal Matching Distance (MMD). Lower is better.}
	\begin{tabular}{c|ccccc|c}
		\toprule[1pt]
		 & PCN \cite{yuan2018pcn} & FoldingNet \cite{yang2018foldingnet} & TopNet \cite{tchapmi2019topnet} & MSN \cite{liu2020morphing} & GRNet \cite{xie2020grnet} & SeedFormer \\
		\midrule[0.3pt]
        Fidelity & 2.235 & 7.467 & 5.354 & 0.434 & 0.816 & \textbf{0.151}\\
        MMD      & 1.366 & 0.537 & 0.636 & 2.259 & 0.568 & \textbf{0.516} \\
		\bottomrule[1pt]
	\end{tabular}

	\label{table:kitti}
\end{table*}

\begin{figure*}[t]
	\begin{center}
		\includegraphics[width=0.99\linewidth]{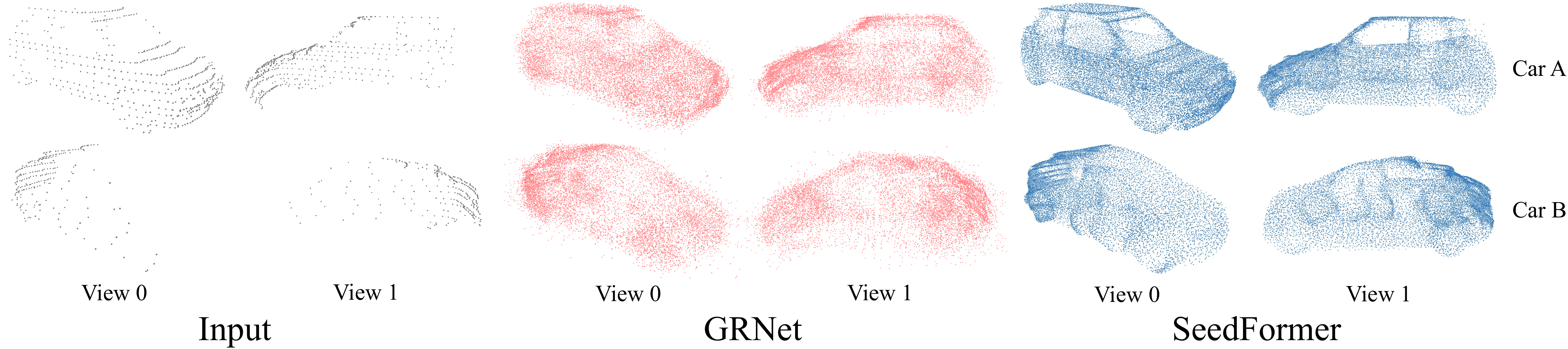}
	\end{center}
	\caption{Visual comparison of point cloud completion results on KITTI dataset. For a clearer comparison, we show two different views of each object.}
	\label{fig:kitti}
\end{figure*}

\subsection{Experiments on KITTI}
\noindent\textbf{Data.} In order to evaluate the proposed model on real-scanned data, we further conduct experiments on the KITTI~\cite{geiger2013vision} dataset for completing sparse point clouds of cars in real-world environments. This dataset consists of a sequence of LiDAR scans from outdoor scenes where car objects are extracted in each frame according to the 3D bounding boxes, resulting in a total of 2,401 partial point clouds. Unlike other datasets which are built from synthetic models, the scanned data in KITTI can be highly sparse and does not have complete point cloud as ground-truth. Thus, we follow the experimental settings of GRNet~\cite{xie2020grnet} and evaluate our method using two metrics: (i) Fidelity Distance, which is the average distance from each point in the input to its nearest neighbour in the output. This measures how well the input is preserved; (ii) Minimal Matching Distance (MMD), which is the Chamfer Distance between the output and the car point cloud from ShapeNet that is closest to the output point cloud in terms of CD. This measures how much the output resembles a typical car. 

\noindent\textbf{Results.} Following GRNet, we fine-tune our model (pretrained on PCN dataset) on ShapeNetCars (the cars from ShapeNet) for a fair comparison. Quantitative evaluation results are shown in Tab.~\ref{table:kitti} compared with previous methods. We also show some visual comparisons of the predicted point clouds in Fig.~\ref{fig:kitti}. Each of the car objects is visualized in two different views for a clearer comparison. We can see that our method performs clearly better on real-scanned data. Even with a very sparse input (see the 2-nd row in Fig.~\ref{fig:kitti}), SeedFormer can produce general structures of the desired object.


\begin{table}[t]
    \newlength{\unitm}
	\setlength{\unitm}{0.475\linewidth}
    \begin{minipage}[t]{\unitm}
        \footnotesize
        \caption{Ablation study on shape representations. We also evaluate the density of Patch Seeds.}
        \label{table:ablation:seed}
        \centering

        \setlength{\tabcolsep}{5pt}
        \begin{tabular}{c|ccc}
    		\toprule[1pt]
    		Methods & CD-Avg \\
    		\midrule[0.3pt]
            global feature & 6.97 \\
    		seed number = 128 & 7.00 \\ 
    		\rowcolor{Gray}
    		seed number = 256 & 6.74 \\
    		seed number = 512 & 6.75 \\
    		\bottomrule[1pt]
	    \end{tabular}
    \end{minipage}\hfill%
    \begin{minipage}[t]{\unitm}
        \footnotesize
        \caption{Ablation study on different generator designs.\\}
        \label{table:ablation:uptrans}
        \centering

        \setlength{\tabcolsep}{5pt}
        \begin{tabular}{c|ccc}
    		\toprule[1pt]
    		Methods & CD-Avg \\
    		\midrule[0.3pt]
    		folding operation & 6.93\\
            deconvolution & 6.90 \\
    		graph convolution & 6.88 \\ 
    		point-wise attention & 6.85 \\
    		\rowcolor{Gray}
    		Upsample Transformer & 6.74 \\
    		\bottomrule[1pt]
	    \end{tabular}
    \end{minipage}

    \begin{minipage}[t]{\unitm}
        \footnotesize
        \caption{Alternatives to softmax function in seed generator.\\}
        \label{table:ablation:softmax}
        \centering

        \setlength{\tabcolsep}{5pt}
        \begin{tabular}{c|ccc}
    		\toprule[1pt]
    		Methods & CD-Avg \\
    		\midrule[0.3pt]
            w/ softmax & 6.83 \\
            \rowcolor{Gray}
    	    w/o softmax & 6.74 \\ 
    	    w/ scaled-softmax & 6.80 \\
    		w/ log-softmax & 6.80 \\
    		\bottomrule[1pt]
	    \end{tabular}
    \end{minipage}\hfill%
    \begin{minipage}[t]{\unitm}
        \footnotesize
        \caption{Ablation study on different positional encoding in Upsample Transformer.}
        \label{table:ablation:encoding}
        \centering

        \setlength{\tabcolsep}{5pt}
        \begin{tabular}{c|ccc}
    		\toprule[1pt]
    		Methods & CD-Avg \\
    		\midrule[0.3pt]
            none &  6.88 \\
    		positional & 6.80 \\ 
    		\rowcolor{Gray}
    		positional + regional & 6.74 \\
    		combined & 6.78 \\
    		\bottomrule[1pt]
	    \end{tabular}
    \end{minipage}
\end{table}

\subsection{Ablation Studies} 
\label{sec:eval:ablation}
In this section, we demonstrate the effectiveness of several architecture designs in SeedFormer and provide some optional choices that can be applied in the network. All the networks are trained on PCN dataset with identical settings.

\noindent\textbf{Patch Seeds.}
The Patch Seeds representation attempts to preserve geometric features locally, which shows clear superiority of detail recovery over previous global feature designs. We first ablate these two network architectures by replacing Patch Seeds with a global feature in the SeedFormer network. Specifically, in the seed generator, the global feature is obtained through max-pooling from input patch features, and is used to produce a coarse point cloud by a deconvolution layer~\cite{xiang2021snowflakenet}. Then, in the subsequent upsample layers, this global feature is concatenated with each input point which is fed to the Upsample Transformer similarly. All other architectures are identical. Results in Tab.~\ref{table:ablation:seed} show clear improvement of our Patch Seeds design (6.74 vs 6.97). In addition, we ablate the number of seeds in the network and show that 256 seed points are suitable for covering an input 3D object with proper density.

\noindent\textbf{Upsample Transformer and other generator designs.} We compare Upsample Transformer with other point cloud generators. Among previous methods for point cloud completion, folding-based operation~\cite{yang2018foldingnet} and deconvolution~\cite{xiang2021snowflakenet} are commonly-used. Unlike our generator which considers semantic relationships between points, these two designs are applied to process each point independently without using contextual information of point clouds. 

The information from local areas is vital for point cloud completion, and we provide two optional choices for point generation operators. As discussed in Sec.~\ref{sec:method:uptrans}, we adopt graph convolution~\cite{wang2019dynamic} and vanilla Transformer (point-wise attention) to further demonstrate the effectiveness of local aggregation. In this experiment, we replace the Upsample Transformers in both seed generator and upsample layers, and keep other architectures the same. All the generators are designed to produce new features with upsampled points. 
Tab.~\ref{table:ablation:uptrans} shows point generation by local aggregation performs better than previous methods, and the performance of Upsample Transformer stands out in similar designs. 

\noindent\textbf{Alternatives to softmax function.} The softmax function can provide balanced weights in the self-attention mechanism. However, in the case of seed generators, it also presents intrinsic limitations on generating new points. In Tab.~\ref{table:ablation:softmax}, we show that the standard transformer structure (w/ softmax) is not the best choice (6.83) in the seed generator. Our improvement (w/o softmax) by applying self-attention without softmax function aims to release the points from limited weights within a specific range of $(0,1)$. Similar results can be achieved by using a log-softmax or scaled-softmax (multiplied by a scale parameter $\lambda$) function, which are also alternatives to the original softmax function.  

\noindent\textbf{Positional encoding.} Upsample Transformer applies both positional encoding from spatial relations and regional encoding from seed feature relations. This extends the original positional encoding in transformers by utilizing information from interpolated seed features. Tab.~\ref{table:ablation:encoding} shows that this design performs better which also demonstrates the effectiveness of incorporating information from Patch Seeds into the generation process of Upsample Transformer. Moreover, we design another ablation which combines both features in one encoding (6.78). We concatenate the two inputs and obtain a joint version through MLPs.


\begin{figure*}[t]
	\newlength{\unitb}
	\setlength{\unitb}{0.19\linewidth}
	\newlength{\unitc}
	\setlength{\unitc}{0.17\linewidth}
	\centering
	
	\begin{subfigure}{\unitb}
		\centering
		\includegraphics[width=\linewidth]{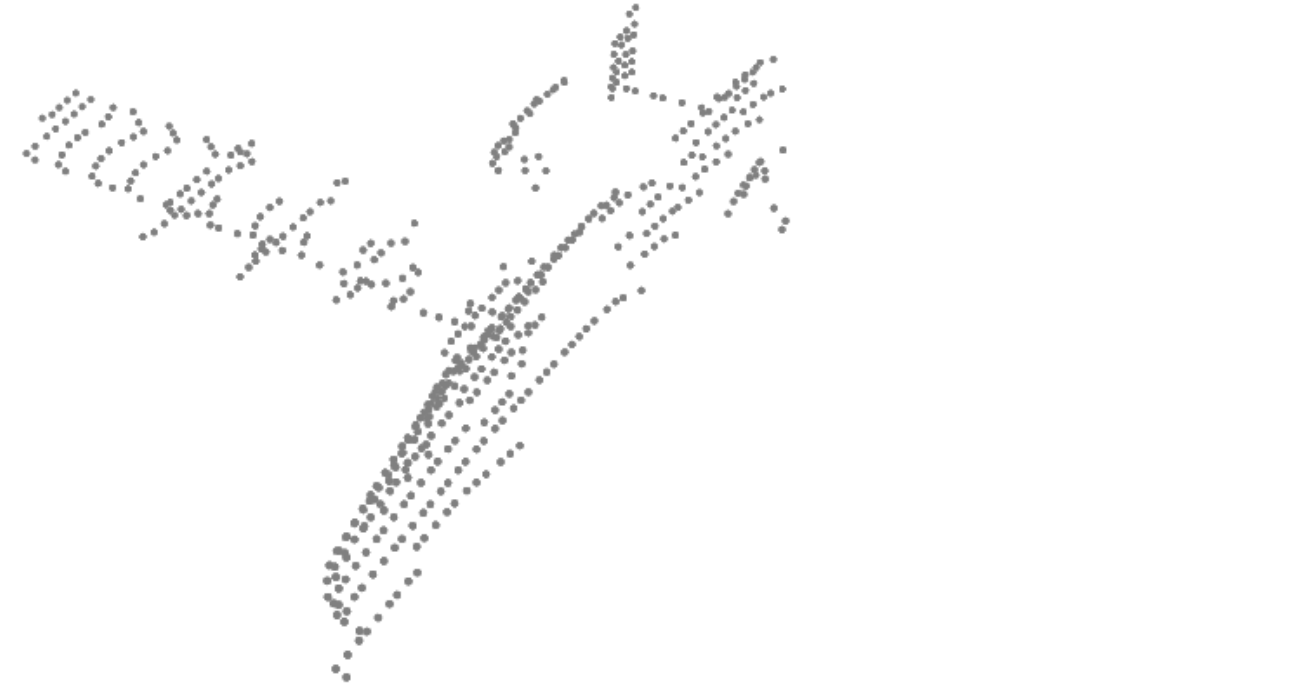}
	\end{subfigure}\hfill%
	\begin{subfigure}{\unitb}
		\centering
		\includegraphics[width=\linewidth]{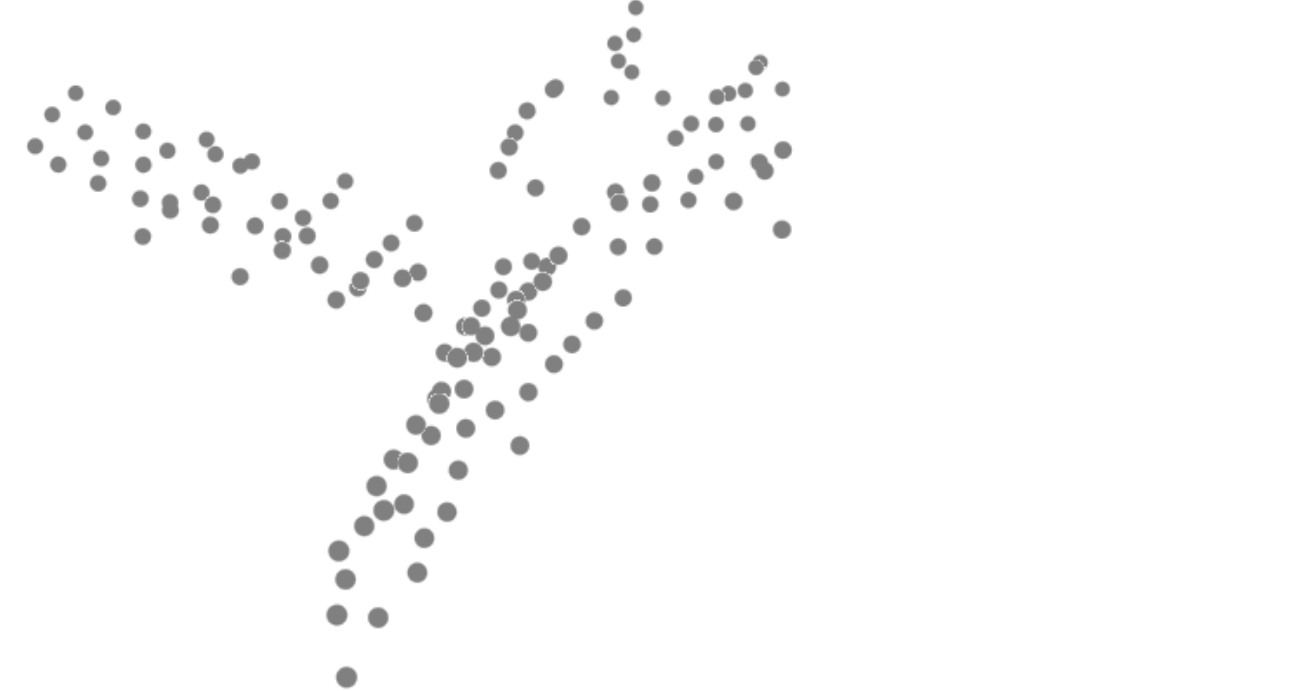}
	\end{subfigure}\hfill%
	\begin{subfigure}{\unitb}
		\centering
		\includegraphics[width=\linewidth]{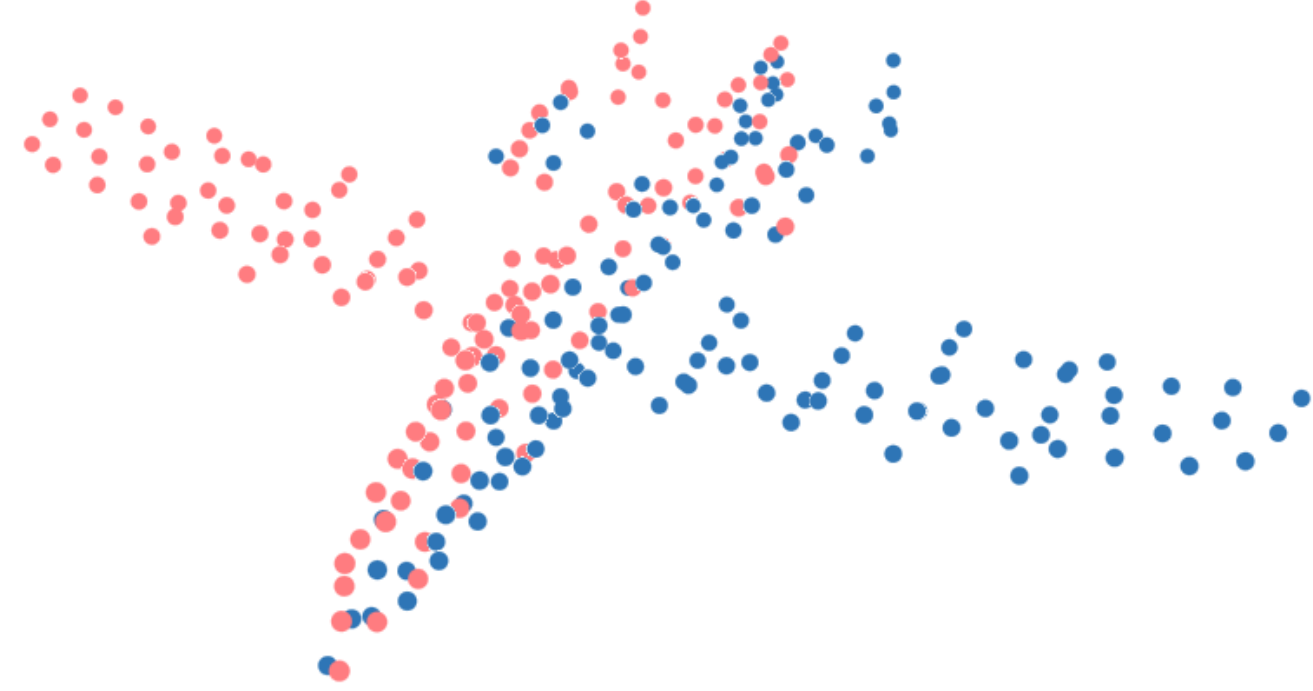}
	\end{subfigure}\hfill%
	\begin{subfigure}{\unitb}
		\centering
		\includegraphics[width=\linewidth]{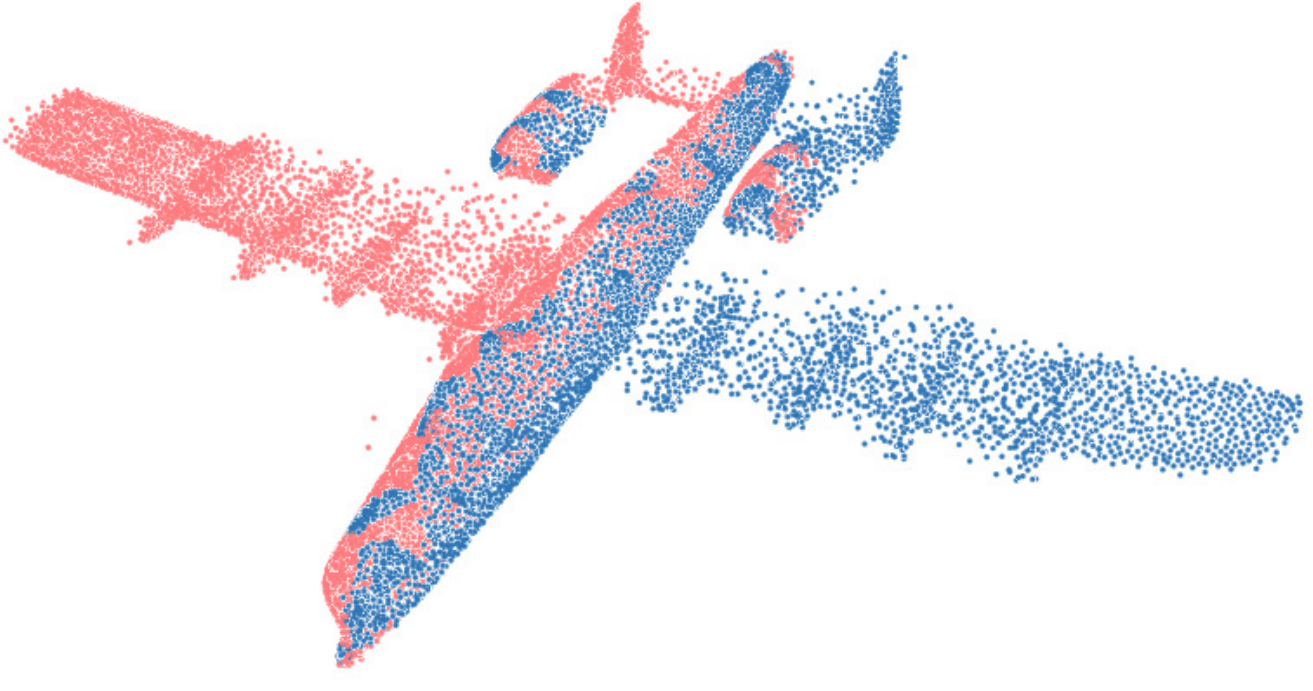}
	\end{subfigure}\hfill%
	\begin{subfigure}{\unitb}
		\centering
		\includegraphics[width=\linewidth]{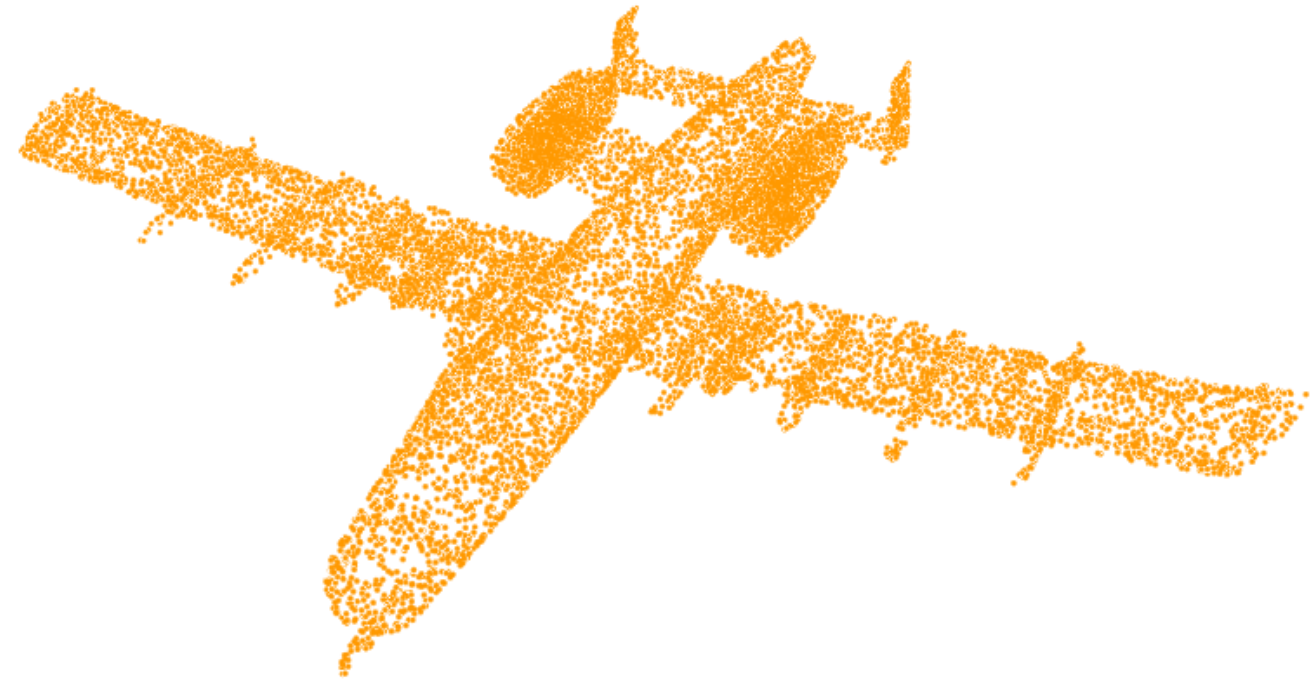}
	\end{subfigure}

	\captionsetup[subfigure]{font=small,labelfont=small}
	\begin{subfigure}{\unitb}
		\centering
		\includegraphics[width=\unitc]{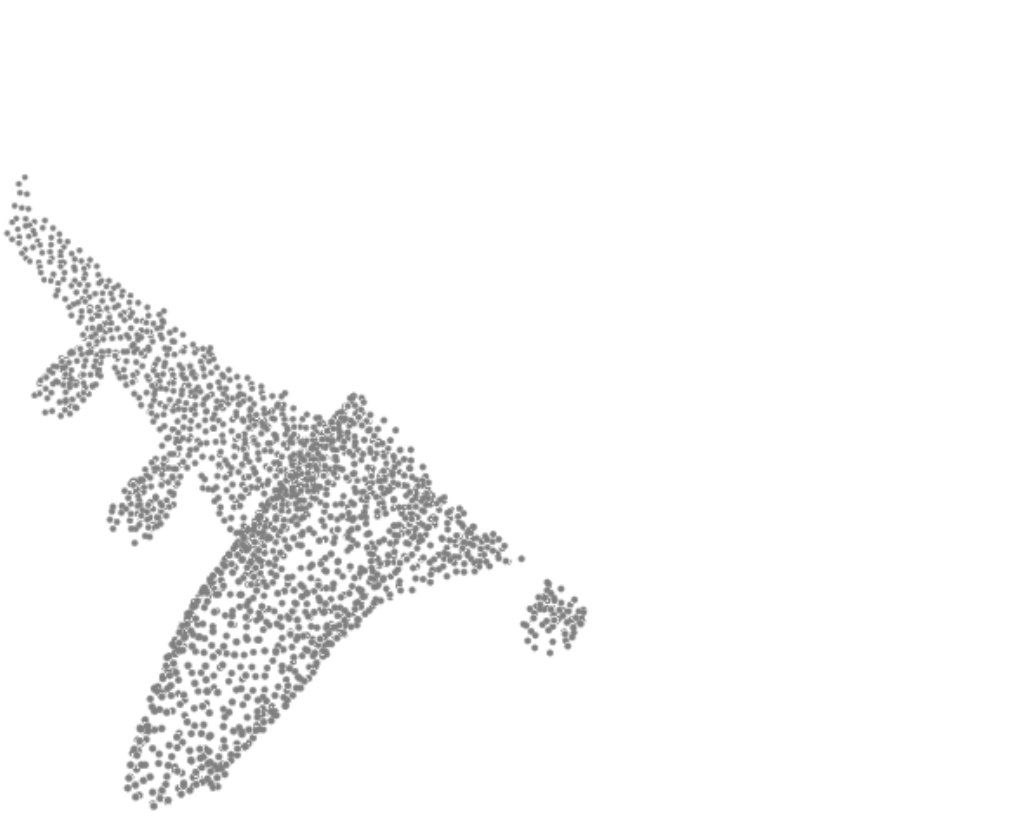}
		\caption{Input}
		\label{fig:insight:partial}
	\end{subfigure}\hfill%
	\begin{subfigure}{\unitb}
		\centering
		\includegraphics[width=\unitc]{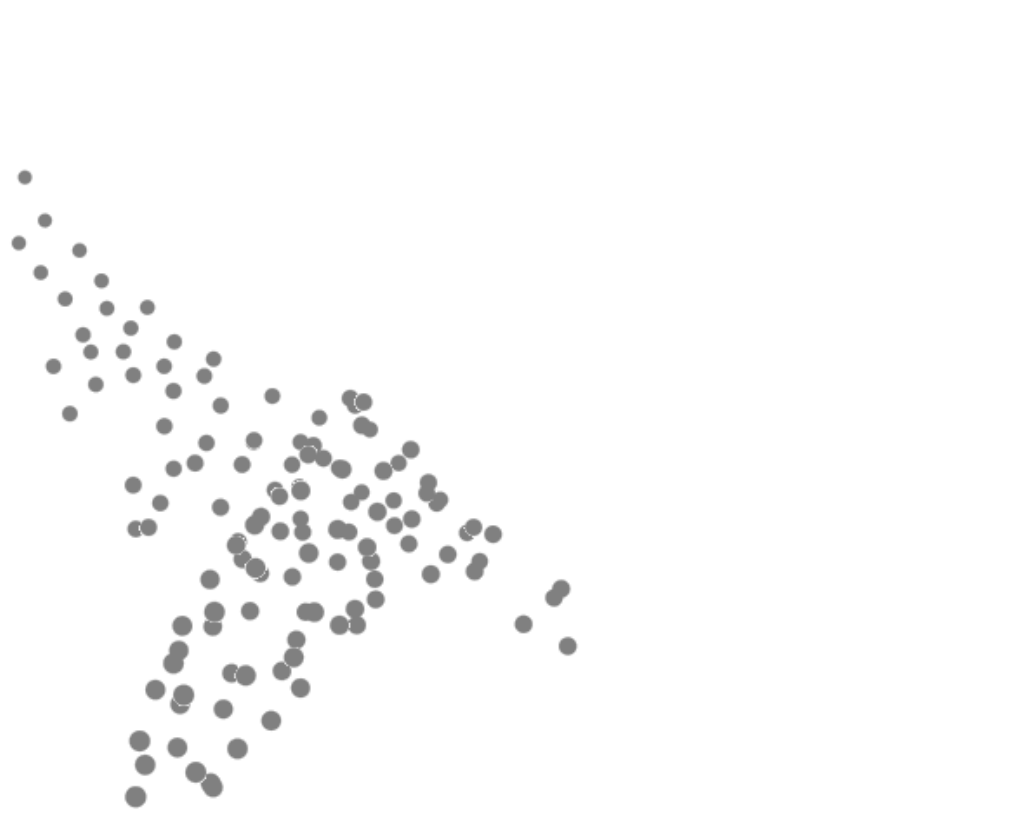}
		\caption{Patch}
		\label{fig:insight:pp}
	\end{subfigure}\hfill%
	\begin{subfigure}{\unitb}
		\centering
		\includegraphics[width=\unitc]{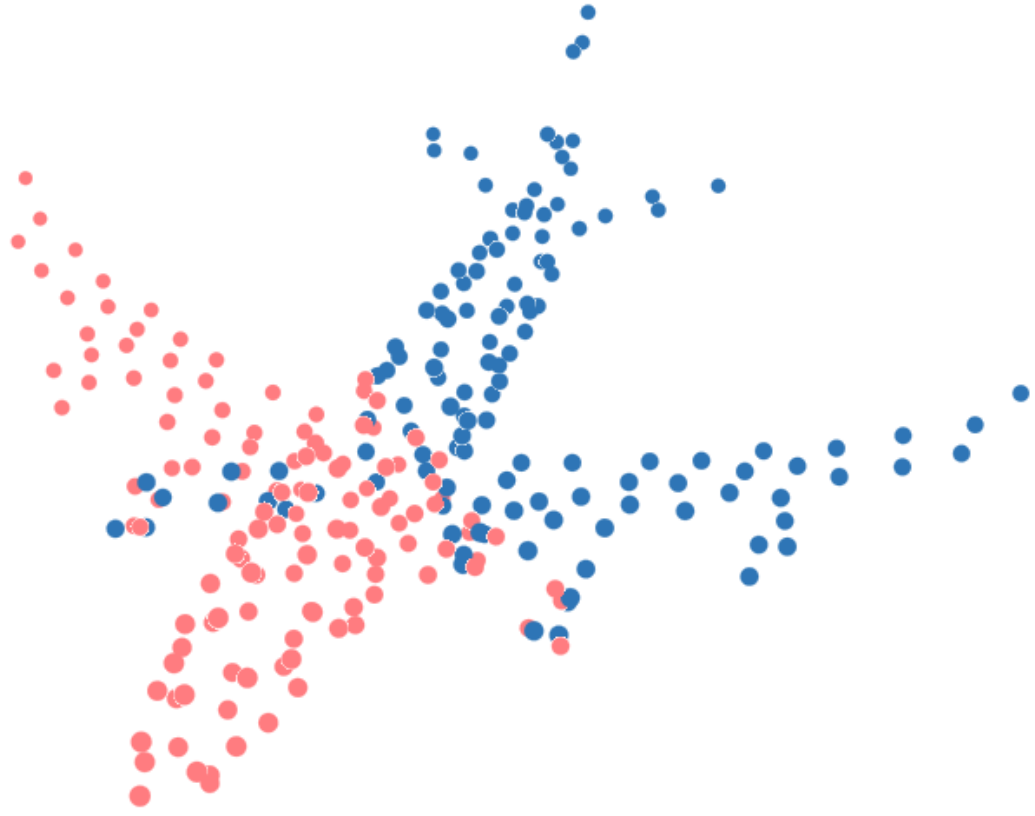}
		\caption{Patch Seeds}
		\label{fig:insight:seed}
	\end{subfigure}\hfill%
	\begin{subfigure}{\unitb}
		\centering
		\includegraphics[width=\unitc]{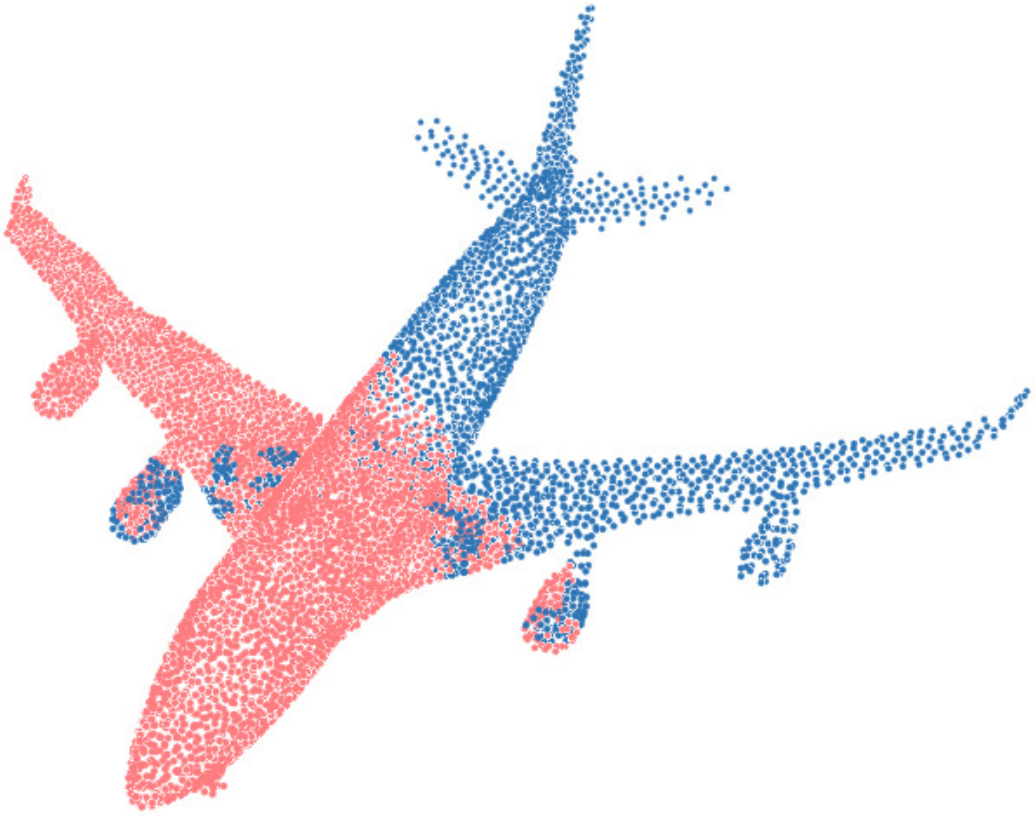}
		\caption{Prediction}
		\label{fig:insight:pred}
	\end{subfigure}\hfill%
	\begin{subfigure}{\unitb}
		\centering
		\includegraphics[width=\unitc]{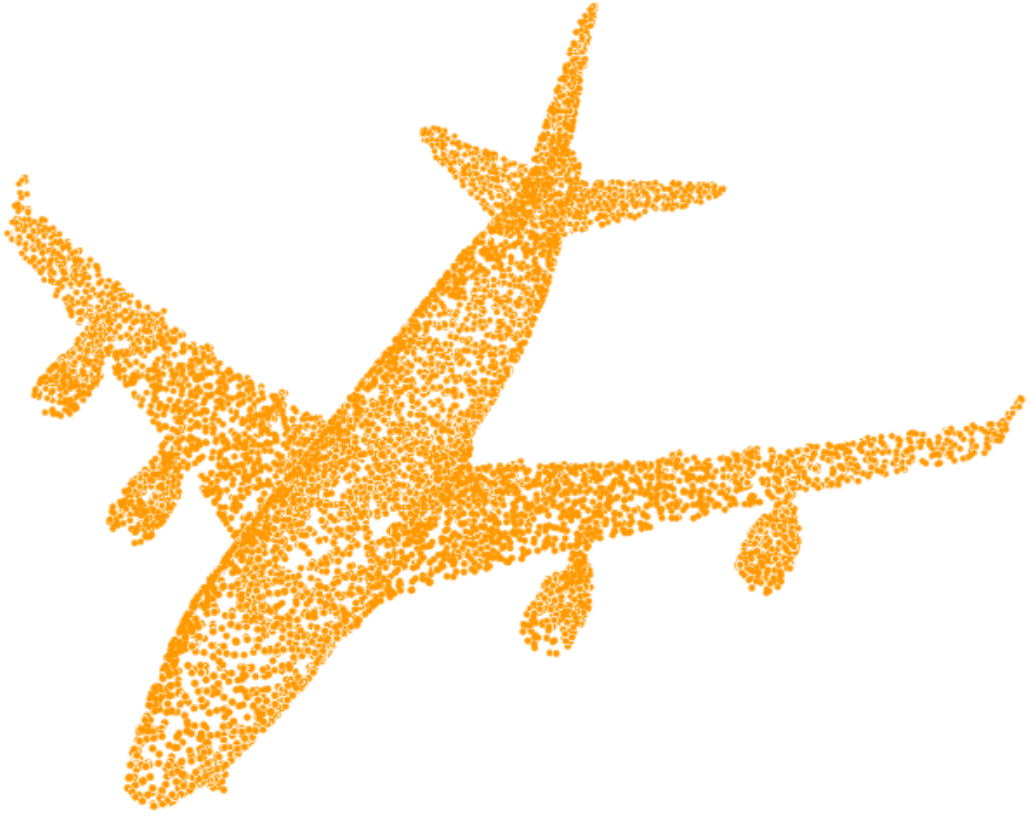}
		\caption{GT}
		\label{fig:insight:gt}
	\end{subfigure}

	\caption{Illustration of Patch Seeds and other intermediate results. (a) Input point cloud. (b) Extracted patch centers from partial input. (c) Generated seeds (red: seed0, blue: seed1) according to the seed generator process. (d) Our predicted results with each point colored according its nearest seed. (e) Ground truth.}
	\label{fig:insight}
\end{figure*}

\section{Visualization of Patch Seeds}
In this section, we give some insights to achieve a deeper understanding of the generation process of Patch Seeds.
The seed generator takes $N_p=128$ patches (Fig.~\ref{fig:insight}(b)) extracted from the input point cloud and produces a new set of $N_s=256$ seed points (Fig.~\ref{fig:insight}(c)). Instead of using a global feature, adopting a patch-to-patch translation allows us to track the paths in the seed generation process. 
Specifically, the generated seeds can be divided into two groups, denoted as seed0 and seed1, since each patch center is split into two seeds according to different groups of self-attention weights in Upsample Transformer (see in Eq.~\ref{equ:att}). Thus, we visualize the obtained Patch Seeds in Fig.~\ref{fig:insight}(c) with specific colors (red: seed0, blue: seed1).
We can see that the red points (seed0) are close to the partial patches where the network tries to preserve the input structures. As for another group (blue points), if the input point cloud is symmetric to the ground-truth (the 1-st row in Fig.~\ref{fig:insight}(a)), SeedFormer learns to duplicate the existing points as well as features, thus it can recover the missiles which are inferred from the seen parts. If the input is asymmetric (in the 2-nd row), our network can also predict missing parts according to the similar observations in the seen point cloud.
For a clearer illustration, we also visualize the predicted complete point clouds in Fig.~\ref{fig:insight}(d) where each point is colored according to its nearest seed. 


\label{sec:eval}

\section{Conclusion}
In this paper, we propose a novel point cloud completion network, termed SeedFormer. As opposed to previous methods, SeedFormer introduce a new shape representation, namely Patch Seeds, in point cloud completion. The idea of Patch Seeds is to capture both global shape structure and fine-grained local details by learning regional features which are stored in several local seeds. This leads to a better performance of shape recovery and detail preservation in the generated point clouds. Furthermore, by extending the transformer structure into point generation, we propose a novel Upsample Transformer to capture useful neighborhood information. 
Comprehensive experiments show that our method achieves clear improvements on several challenging benchmark dataset compared to state-of-the-art competitors.

\noindent\textbf{Acknowledgements.} This work is supported by the National Natural Science Foundation of China (No. 61832008).
\label{sec:conclusion}

%
%
\bibliographystyle{splncs04}
\bibliography{egbib}

\newpage
\renewcommand\thesection{\Alph{section}}
\renewcommand\thesubsection{\Alph{subsection}}

\setcounter{section}{0}

In this supplementary material, we provide additional information to complement the manuscript. First, we present additional implementation details and experimental settings of SeedFormer (Sec.~\ref{sec:supp:detail}). Second, we provide more details about the ablation studies (variants of SeedFormer) in Sec.~\ref{sec:supp:ablation}. Third, we show complexity analysis of our method compared with previous state-of-the-art methods (Sec.~\ref{sec:supp:complex}). At last, we present more qualitative results compared with other methods (Sec.~\ref{sec:supp:experiment}).

\section{Implementation Details}
\label{sec:supp:detail}
In this section, we provide additional implementation details of the proposed SeedFormer.

\vspace{5pt}
\noindent\textbf{Encoder.} We use point transformer~\cite{zhao2021point} ($SA$) and set abstraction~\cite{qi2017pointnet++} ($PT$) layers to extract features from an input point cloud. The point cloud (with 2,048 points) is downsampled using FPS in each layer of set abstraction. Detailed network architecture is as follows: $SA(C=128, N=512) \rightarrow PT(C=128) \rightarrow SA(C=256, N=N_p) \rightarrow PT(C=C_p)$. The outputs are $N_p=128$ patch centers $\mathcal{P}_p$ as well as the corresponding patch features $\mathcal{F}_p$ with $C_p = 256$ channels. 

\vspace{5pt}
\noindent\textbf{Seed Generator.} The inputs to seed generator are the obtained patch centers $\mathcal{P}_p$ and correponding patch features $\mathcal{F}_p$. 
The implementation of Upsample Transformer (Eq.~1 in the main paper) is slightly different from that in an upsample layer.
Since there are no skipped features, we use $\mathcal{F}_p$, applied by linear projections, for both keys and queries in the transformer. Also, the positional encoding of each point is calculated without incorporating seed features. Then, after we obtain seed features $\mathcal{F}$, we concatenate them with a pooled feature from $\mathcal{F}_p$ which is used for encoding global contexts, and apply shared MLPs to produce the seed coordinates $\mathcal{S}$.

\vspace{5pt}
\noindent\textbf{Experimental settings.} For PCN dataset (with 16,384 points in ground-truth), we set the upsampling rate as $r_1=1, r_2=4, r_3=8$ where the first upsample layer is used to refine the input coarse point cloud $\mathcal{P}_0$. For ShapeNet-34/55 (8,192 points), we set $r_1=1, r_2=4, r_3=4$. Each input point cloud from all datasets contains 2,048 points. For those point clouds with less than 2,048 points, we randomly duplicate input points; for those with more than 2,048 points, we pick a subset as input. 

\section{Ablation Studies}
\label{sec:supp:ablation}
We give more details about ablation networks used in Sec.~4.5 in the main paper.
Following the same idea of point generation by local aggregation, we propose two optional generator designs which are both more effective than previous methods. Compared with the default Upsample Transformer, these options can also achieve decent results (Tab.~6 in the main paper) while being more efficient. Similarly, given the features $\{f_i^l\}_{i=1}^{N_l}$ of the input point cloud $\mathcal{P}_l$, we first apply graph convolutions~\cite{wang2019dynamic} to generate new point features by aggregating local neighborhood information:
\begin{equation}
	h_{im} = \max_{j \in \mathcal{N}(i)} \alpha_m(f_j^l).
\end{equation}
Here, $\alpha_m$ is a feature mapping function (MLPs) and $m=1,2,...,r_l$ corresponds to each of the $r_l$ generation processes in this layer. The output features are $\mathcal{H}_l = \{h_{im} | i = 1,2,...,N_l; m = 1,2,...,r_l\} \in \mathbb{R}^{r_l N_l \times C}$ with a upsampling rate of~$r_l$. Then, we obtain the new point cloud by learning displacement offsets from the produced features using shared MLPs.

Another option applies transformer structures in point generation. Different from an Upsample Transformer, we adopt point-wise attention which is more efficient in the computations. Similar to Eq.~3 in the main paper, we compute self-attention weights by:
\begin{equation}
	\hat{a}_{ijm} = \alpha_m(\beta(q_i^l) - \gamma(k_j^l) + \delta), j \in \mathcal{N}(i).
\end{equation}
The difference is that $\alpha_m: \mathbb{R}^C \rightarrow \mathbb{R}$ outputs one-dimensional weights which are assigned to different points in the neighborhood. Then, we obtain the generated point features by combining weights with duplicated values (Eq.~5 in the main paper). To sum up, the point-wise attention is faster (Sec.~\ref{sec:supp:complex}) while it can still outperform state-of-the-art methods.

\begin{table*}[t]
	\centering
	\footnotesize
	\setlength{\tabcolsep}{10pt}
	\caption{Complexity analysis on PCN dataset evaluated as the number of parameters (Params) and theoretical computational cost (FLOPs). We also report the average CDs of all categories as references.}
	\vspace{5pt}
	\begin{tabular}{c|cc|c}
		\toprule[1pt]
		Methods & Params & FLOPs & CD-Avg   \\
		\midrule[0.3pt]
		FoldingNet~\cite{yang2018foldingnet} & \textbf{2.41M} & 27.65G & 14.31 \\
		PCN~\cite{yuan2018pcn} & 6.84M & 14.69G & 9.64\\
		GRNet~\cite{xie2020grnet} & 76.71M & 25.88G & 8.83\\
		SnowflakeNet~\cite{xiang2021snowflakenet} & 19.32M & 10.32G & 7.21 \\
		\midrule[0.3pt]
		point-wise attention & 3.12M & \textbf{7.87G} & 6.85\\
		SeedFormer & 3.20M & 29.61G & \textbf{6.74} \\
		\bottomrule[1pt]
	\end{tabular}
	
	\label{table:complex}
\end{table*}

\section{Complexity Analysis}
\label{sec:supp:complex}
We provide a detailed complexity analysis of our method. Tab.~\ref{table:complex} reports the theoretical computational cost (FLOPs) and number of parameters for different models. The scores are measured on PCN dataset (16,384 points) with a batch size of 1. We also provide the overall Chamfer Distances as references. We can see that the model size of SeedFormer is very favorable compared to previous methods. This is owing to our designs of seed generator and upsample layers which process each point with a shared generator. 
On the other hand, since Upsample Transformer is required to aggregate local points in each generation, the computational cost of SeedFormer is relatively high but it is comparable with GRNet~\cite{xie2020grnet} and FoldingNet~\cite{yang2018foldingnet}. We also provide a more efficient version of SeedFormer using a faster transformer structure (point-wise attention) as discussed in Sec.~4.5 of the main manuscript. This model achieves a decent result (6.85) with lower computational cost. In all, our model can offer a pleasant trade-off between cost and performance.


\section{Additional Experimental Results}
\label{sec:supp:experiment}
\begin{figure*}[t]
	\newlength{\sunit}
	\setlength{\sunit}{0.165\linewidth}
	\newlength{\sunita}
	\setlength{\sunita}{0.15\linewidth}
	\newlength{\sunitb}
	\setlength{\sunitb}{0.09\linewidth}
	\newlength{\sunitc}
	\setlength{\sunitc}{0.12\linewidth}
	\newlength{\sunitd}
	\setlength{\sunitd}{0.12\linewidth}
	
	\centering
	
	\begin{subfigure}{\sunit}
		\centering
		\includegraphics[width=\sunita]{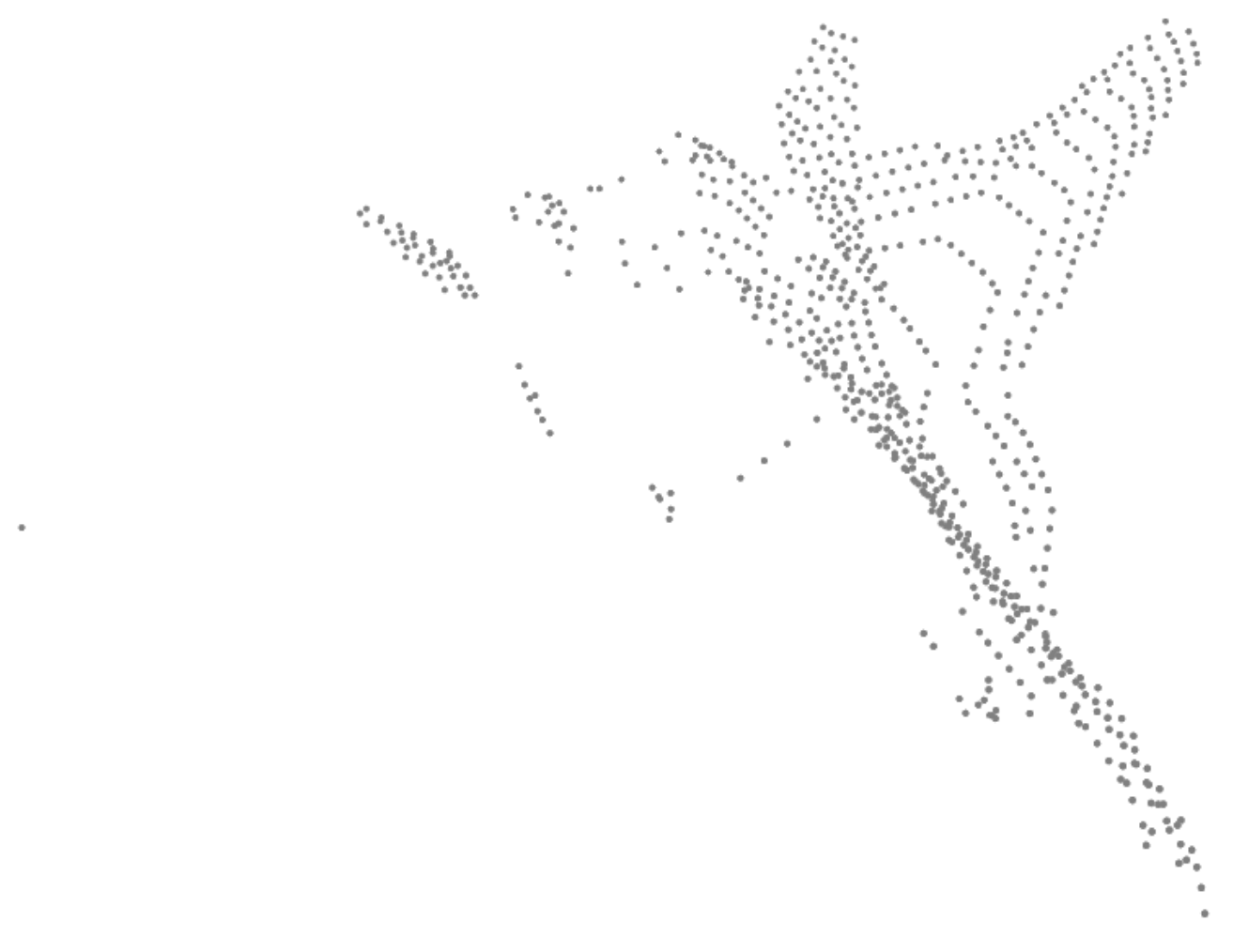}
	\end{subfigure}\hfill%
	\begin{subfigure}{\sunit}
		\centering
		\includegraphics[width=\sunita]{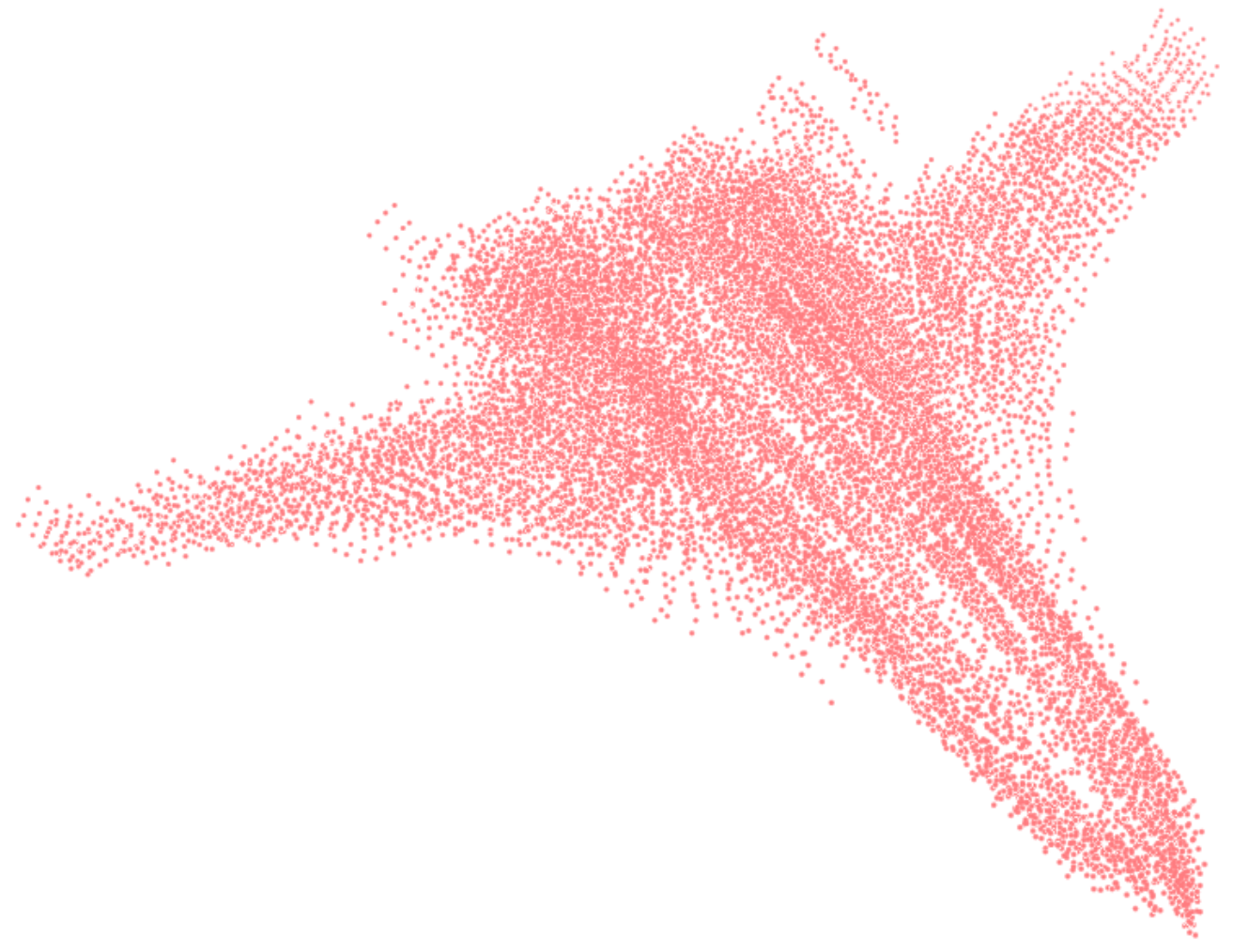}
	\end{subfigure}\hfill%
	\begin{subfigure}{\sunit}
		\centering
		\includegraphics[width=\sunita]{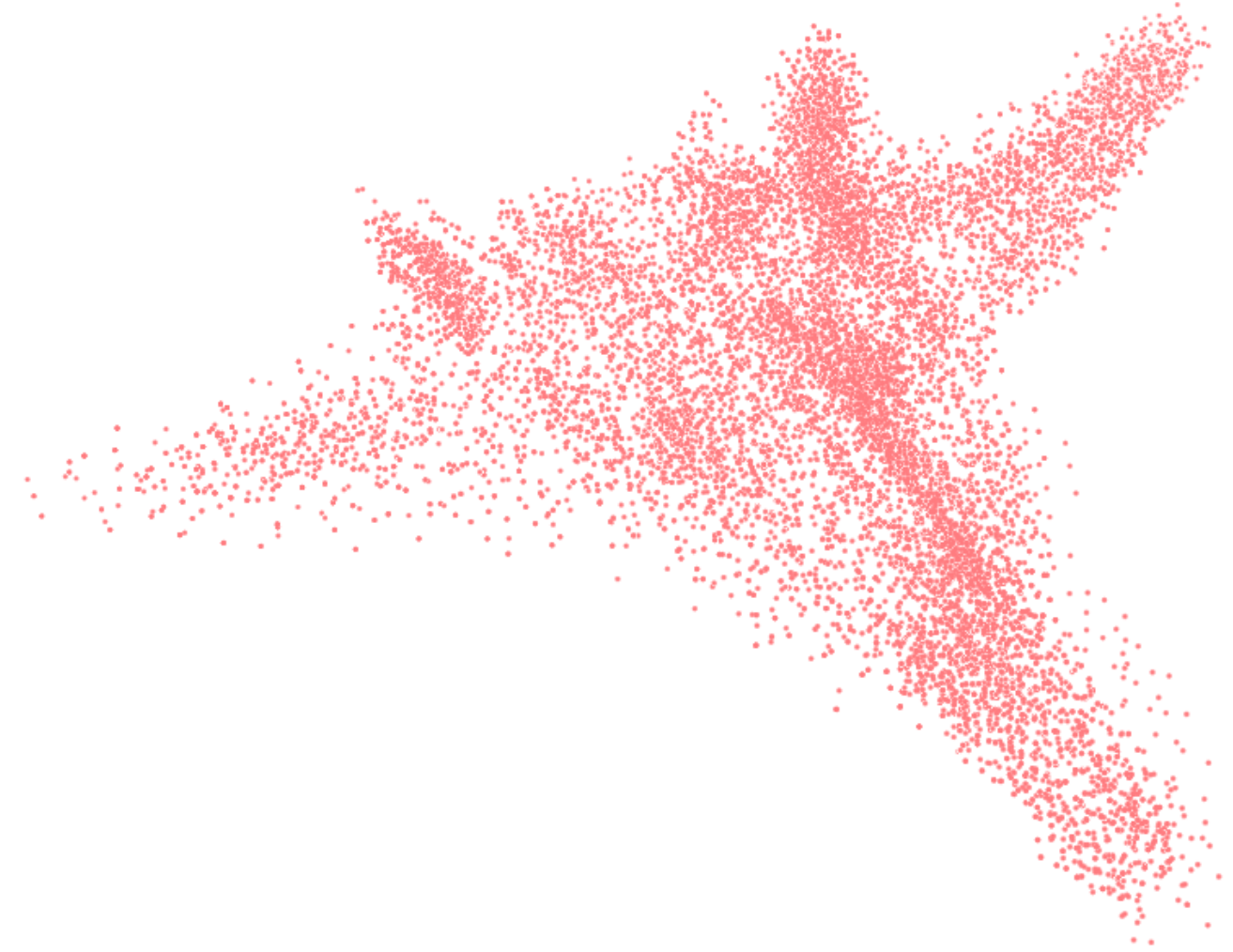}
	\end{subfigure}\hfill%
	\begin{subfigure}{\sunit}
		\centering
		\includegraphics[width=\sunita]{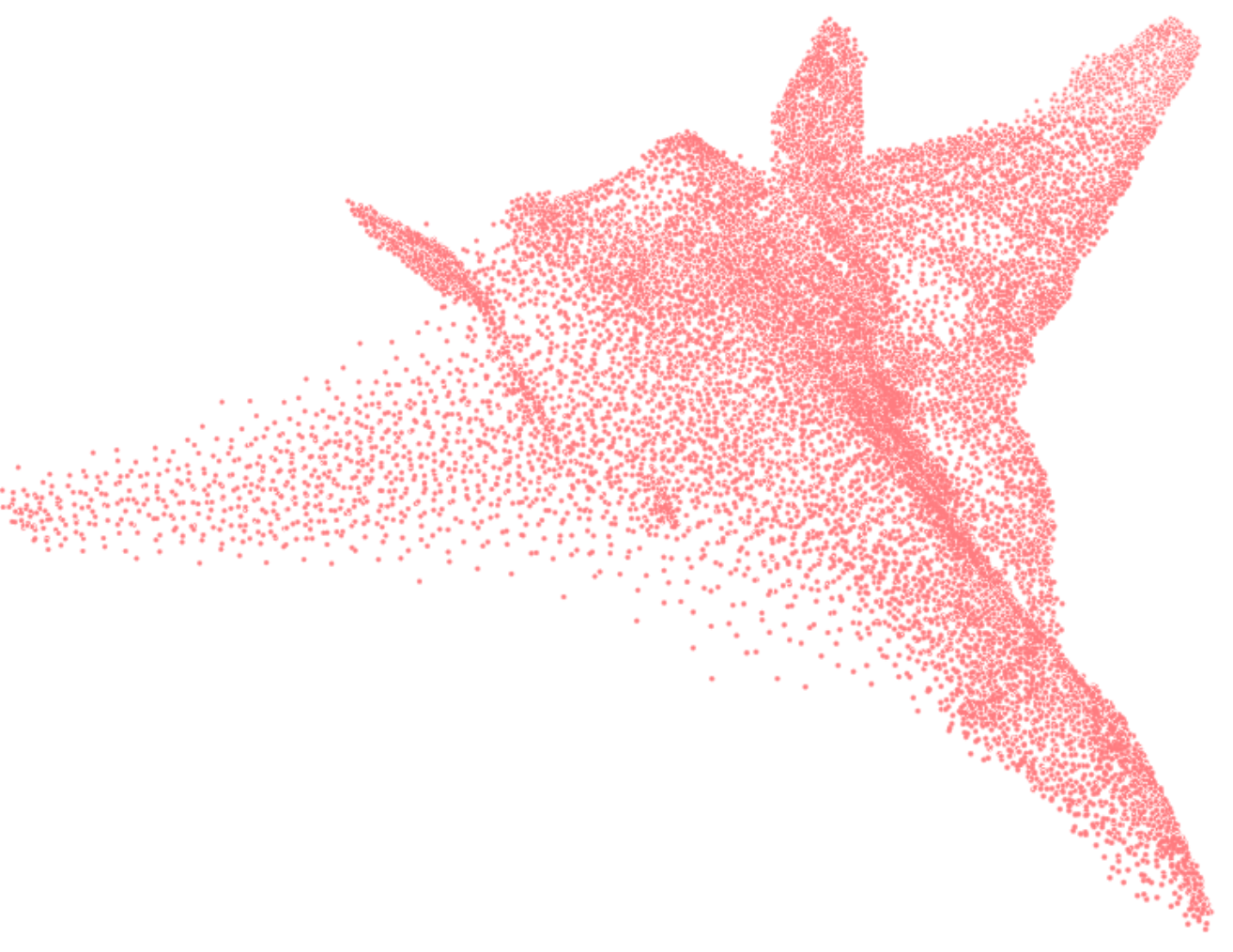}
	\end{subfigure}\hfill%
	\begin{subfigure}{\sunit}
		\centering
		\includegraphics[width=\sunita]{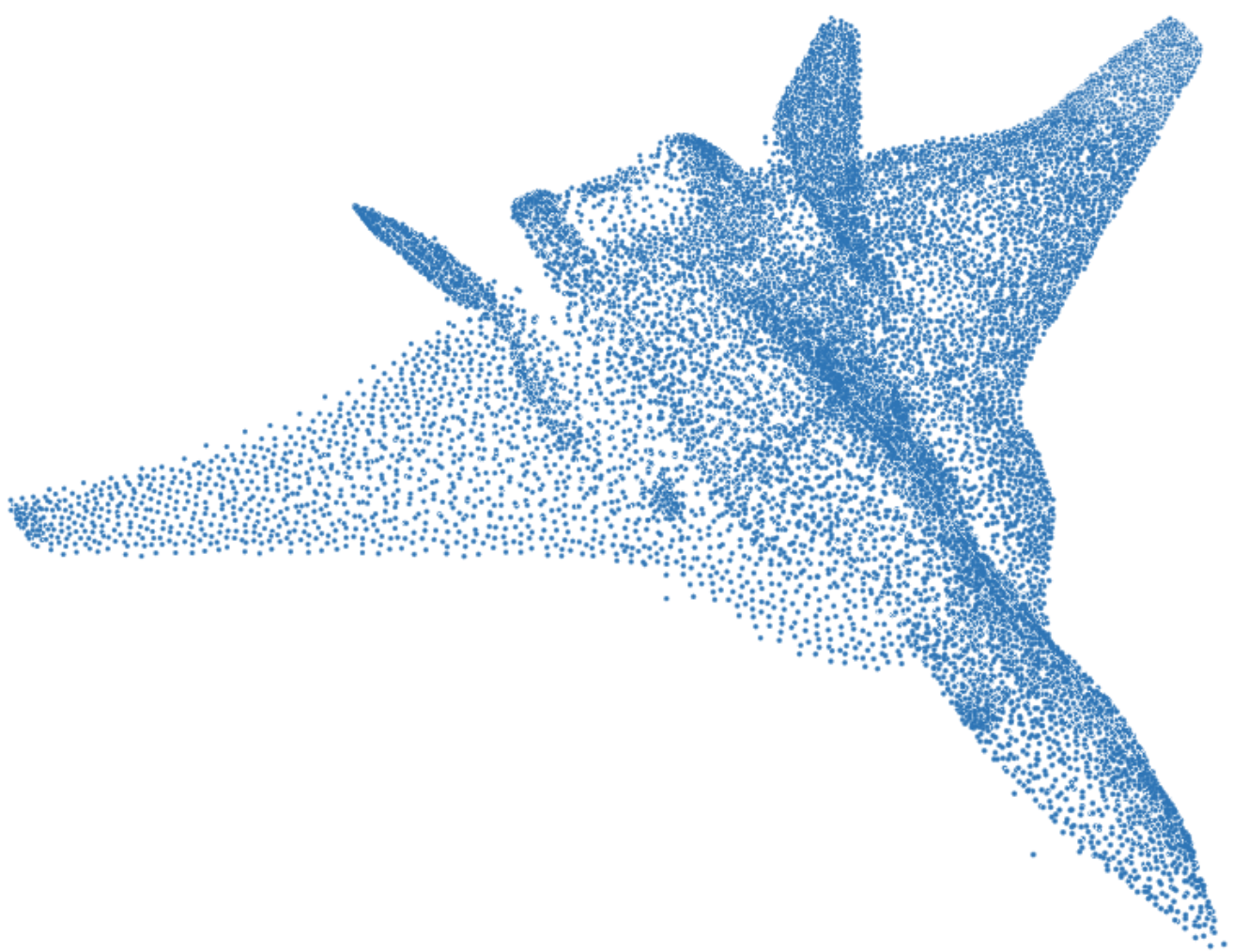}
	\end{subfigure}\hfill%
	\begin{subfigure}{\sunit}
		\centering
		\includegraphics[width=\sunita]{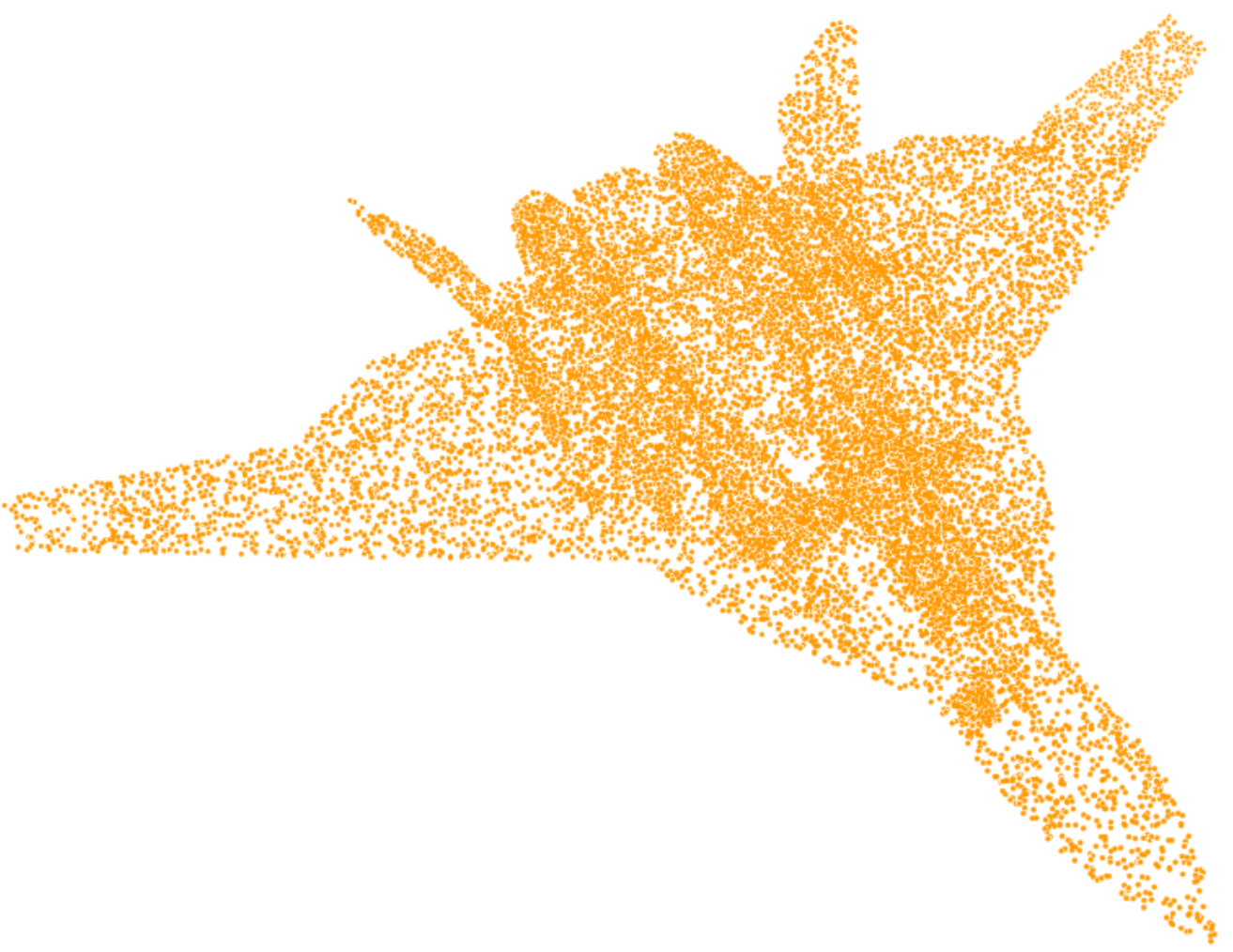}
	\end{subfigure}

	\begin{subfigure}{\sunit}
		\centering
		\includegraphics[width=\sunita]{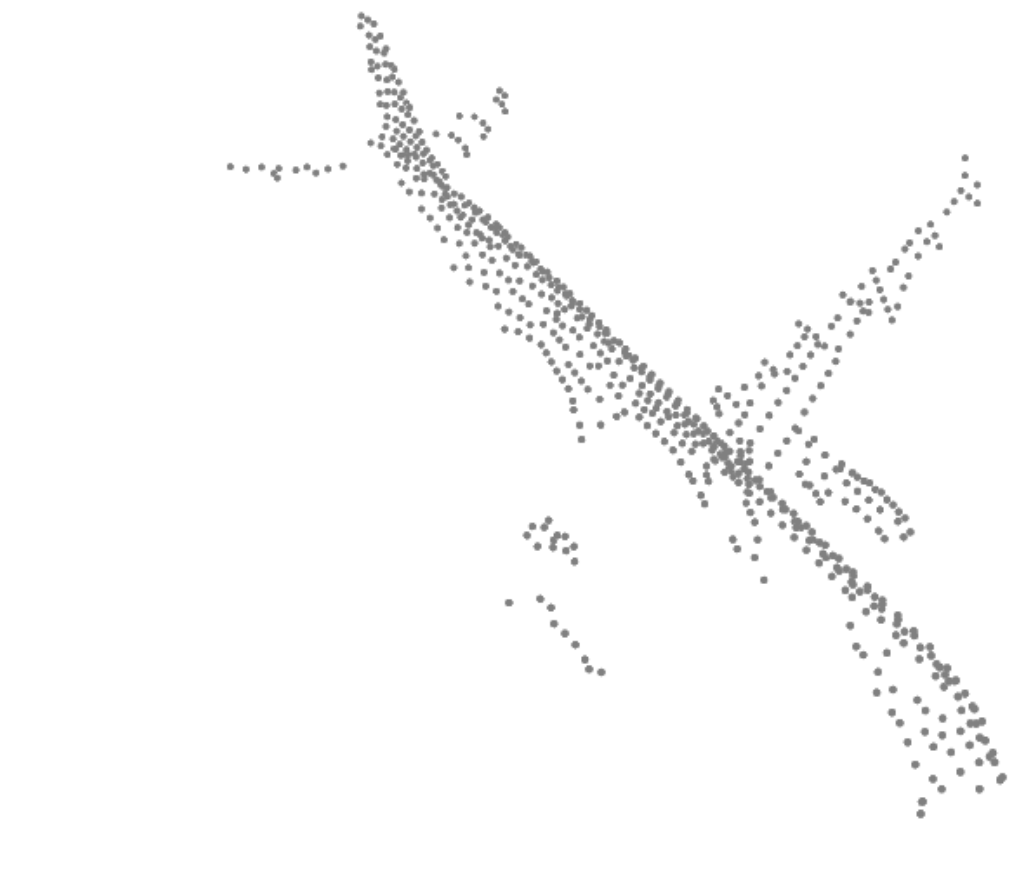}
	\end{subfigure}\hfill%
	\begin{subfigure}{\sunit}
		\centering
		\includegraphics[width=\sunita]{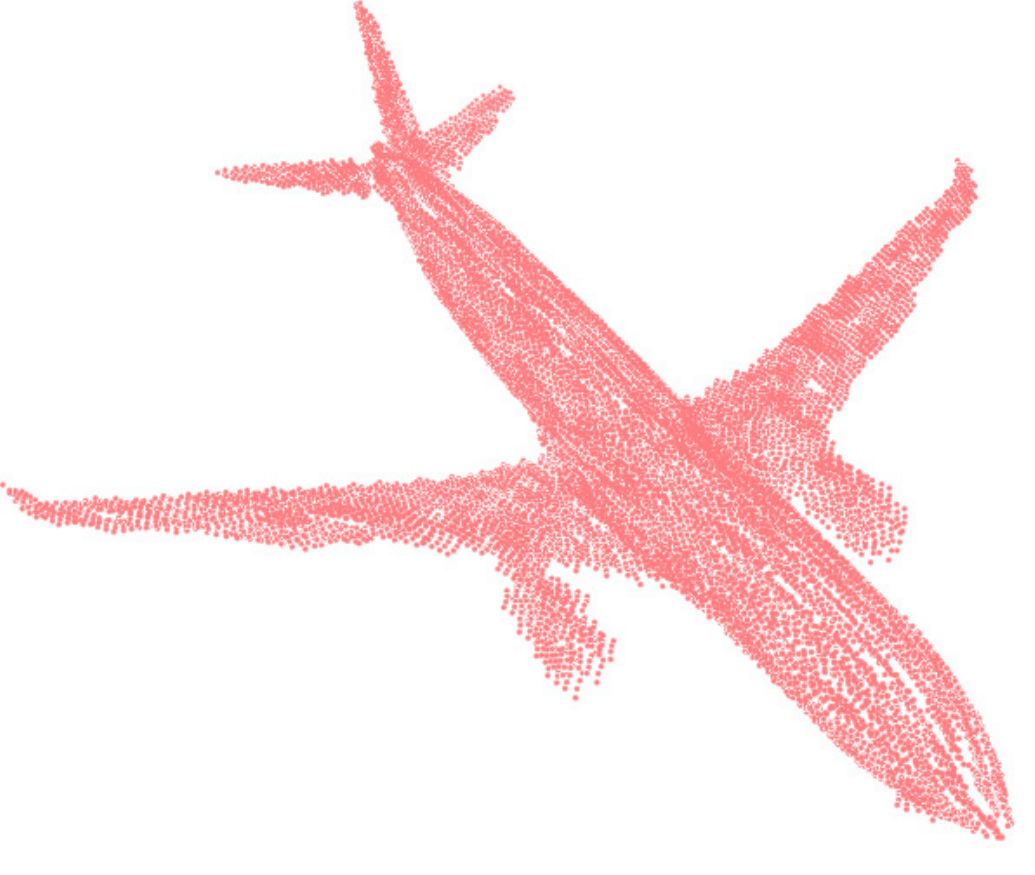}
	\end{subfigure}\hfill%
	\begin{subfigure}{\sunit}
		\centering
		\includegraphics[width=\sunita]{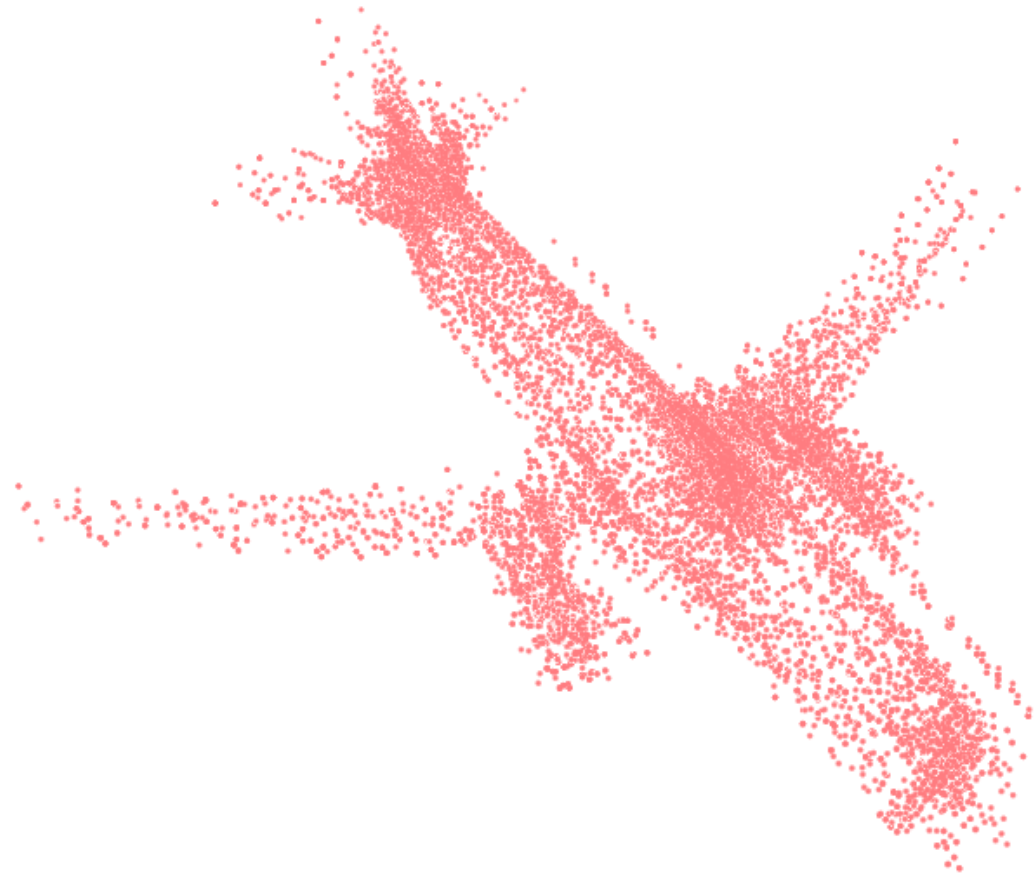}
	\end{subfigure}\hfill%
	\begin{subfigure}{\sunit}
		\centering
		\includegraphics[width=\sunita]{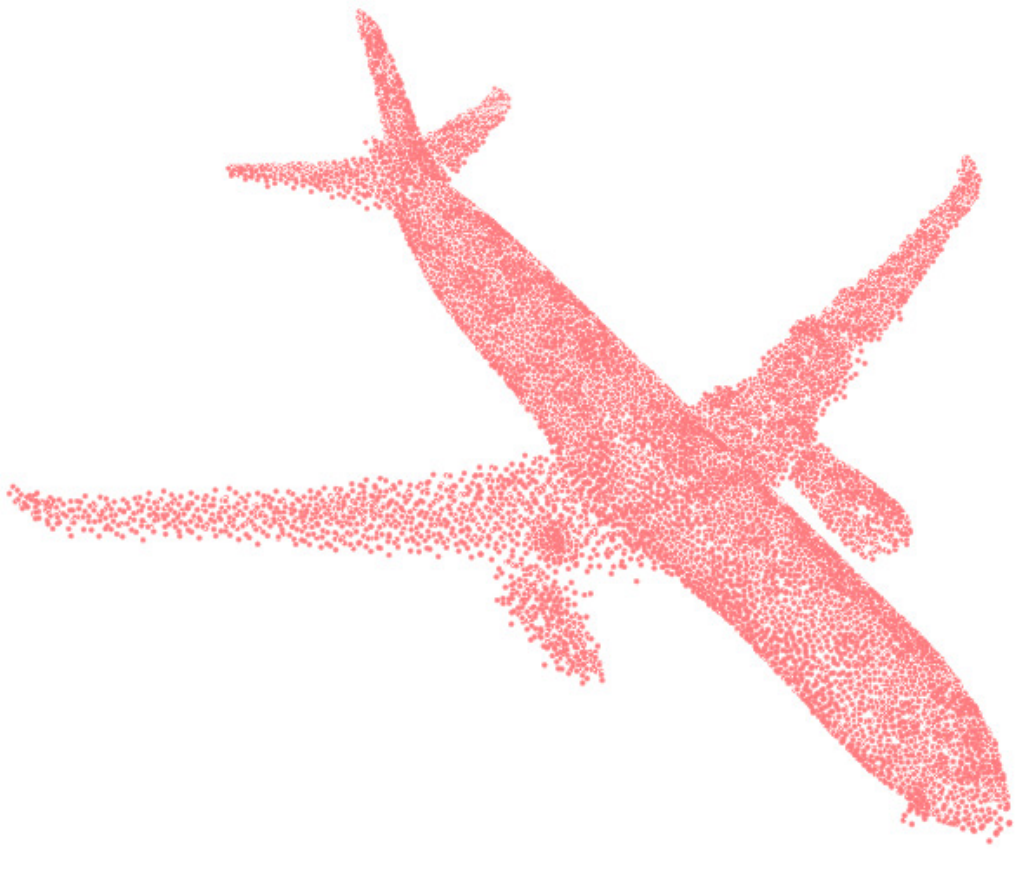}
	\end{subfigure}\hfill%
	\begin{subfigure}{\sunit}
		\centering
		\includegraphics[width=\sunita]{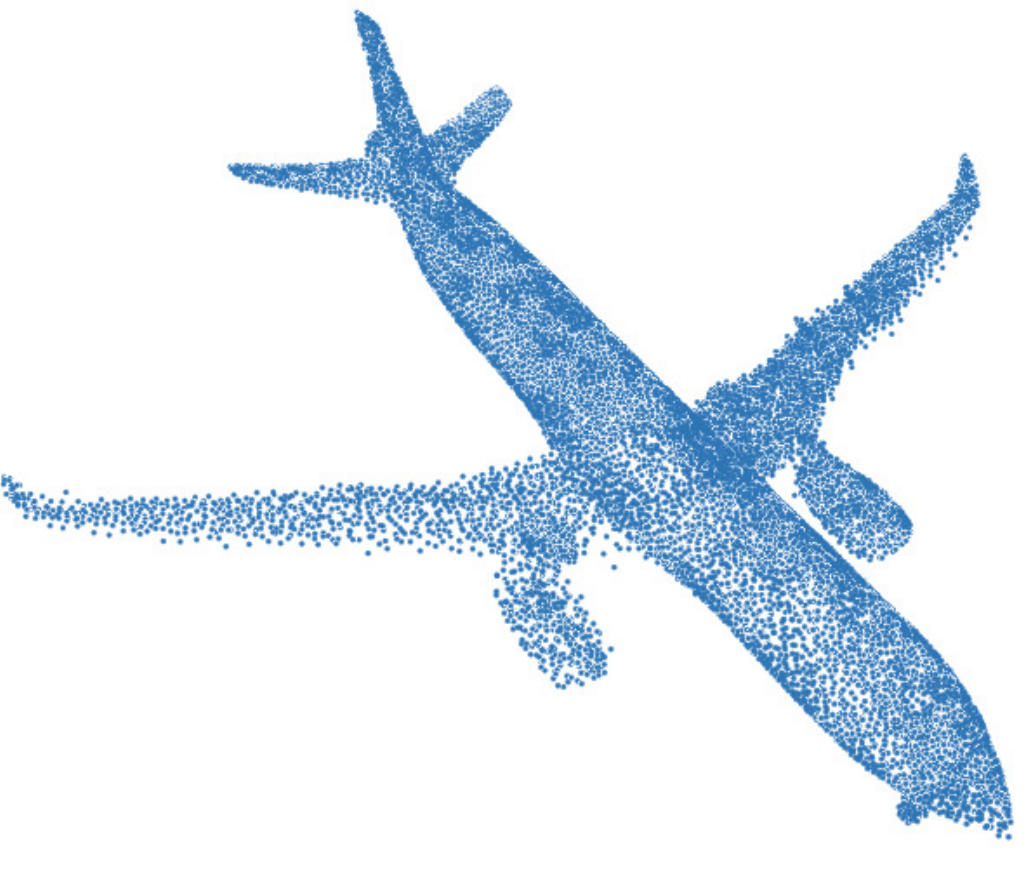}
	\end{subfigure}\hfill%
	\begin{subfigure}{\sunit}
		\centering
		\includegraphics[width=\sunita]{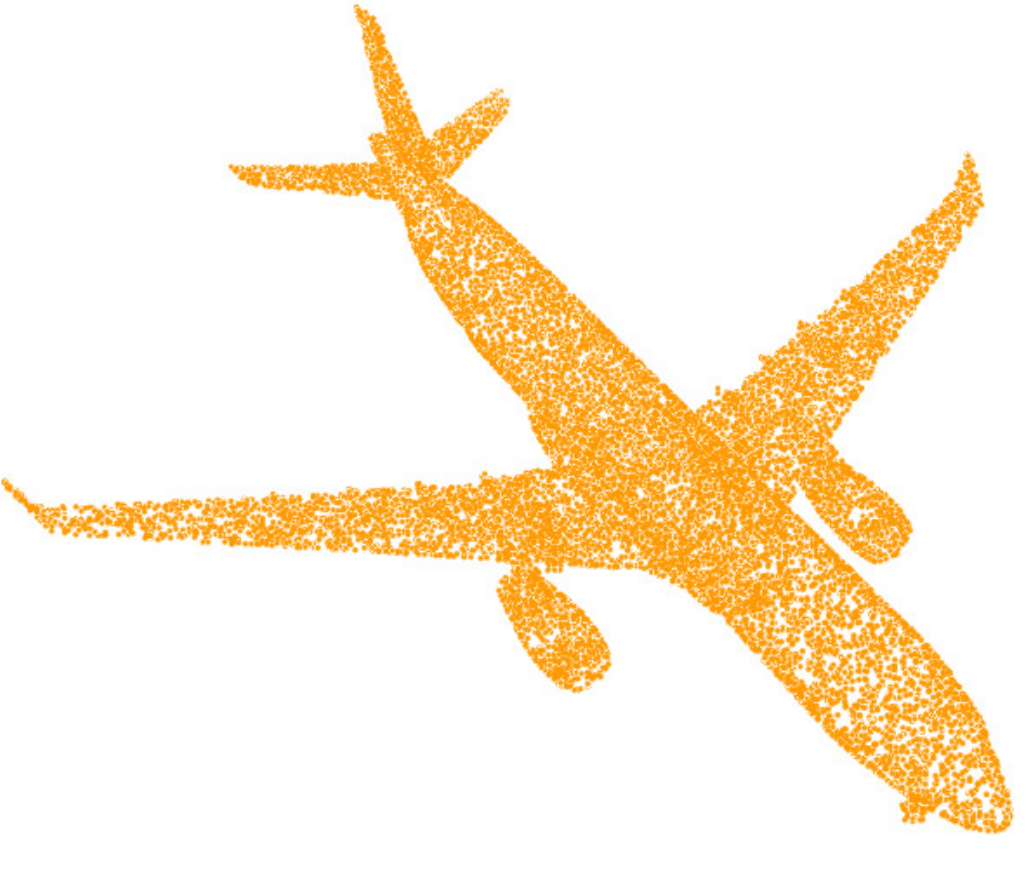}
	\end{subfigure}
	
	
	\begin{subfigure}{\sunit}
		\centering
		\includegraphics[width=\sunitb]{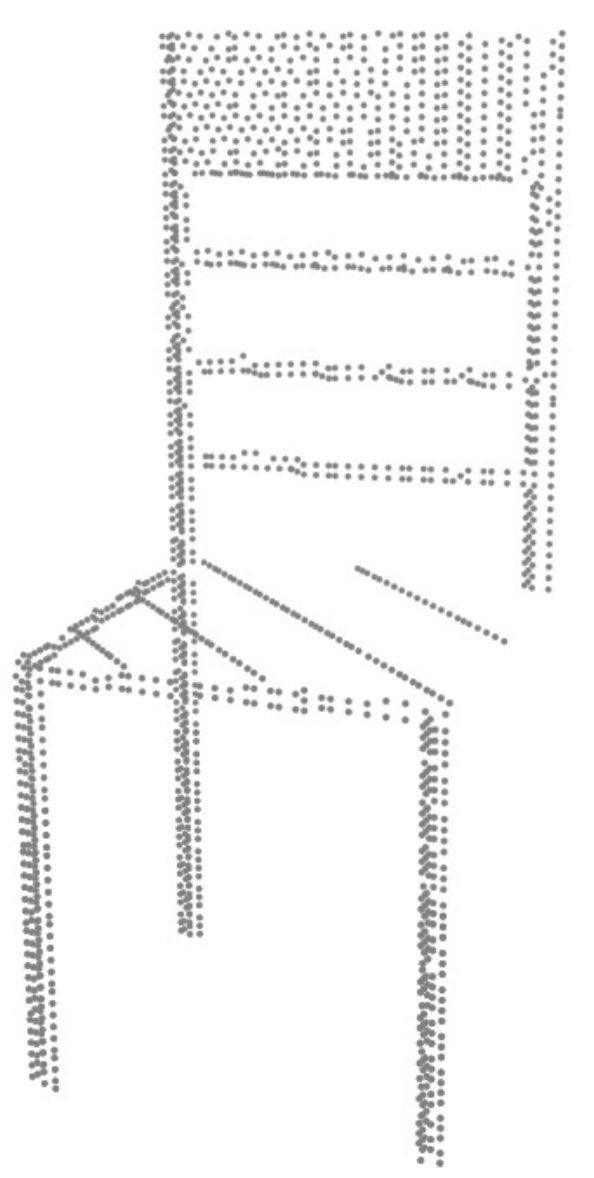}
	\end{subfigure}\hfill%
	\begin{subfigure}{\sunit}
		\centering
		\includegraphics[width=\sunitb]{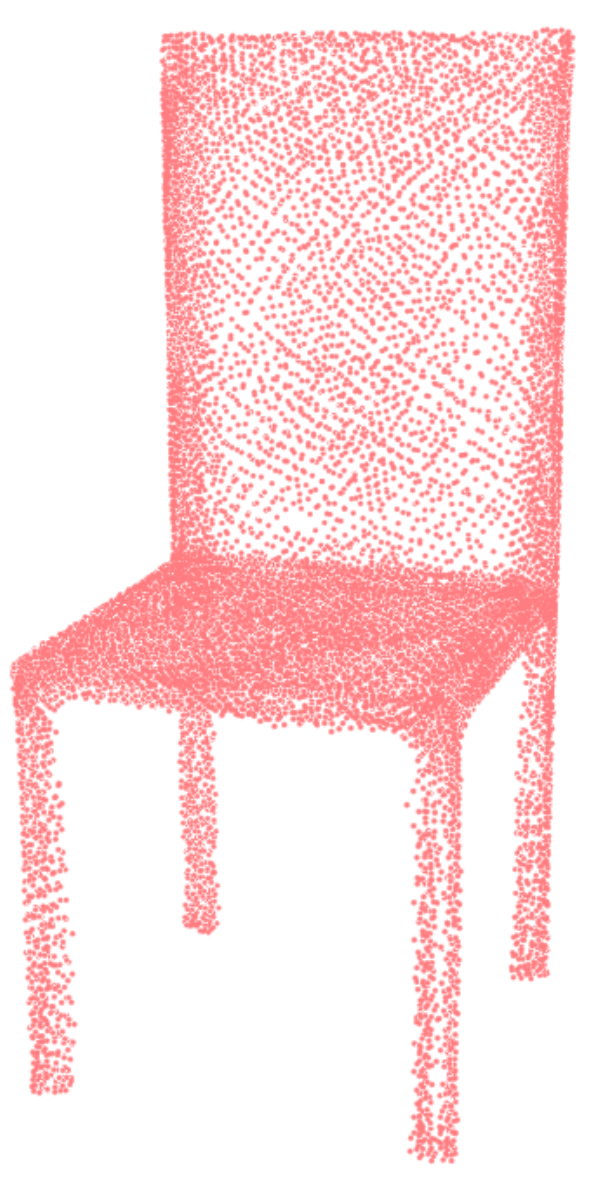}
	\end{subfigure}\hfill%
	\begin{subfigure}{\sunit}
		\centering
		\includegraphics[width=\sunitb]{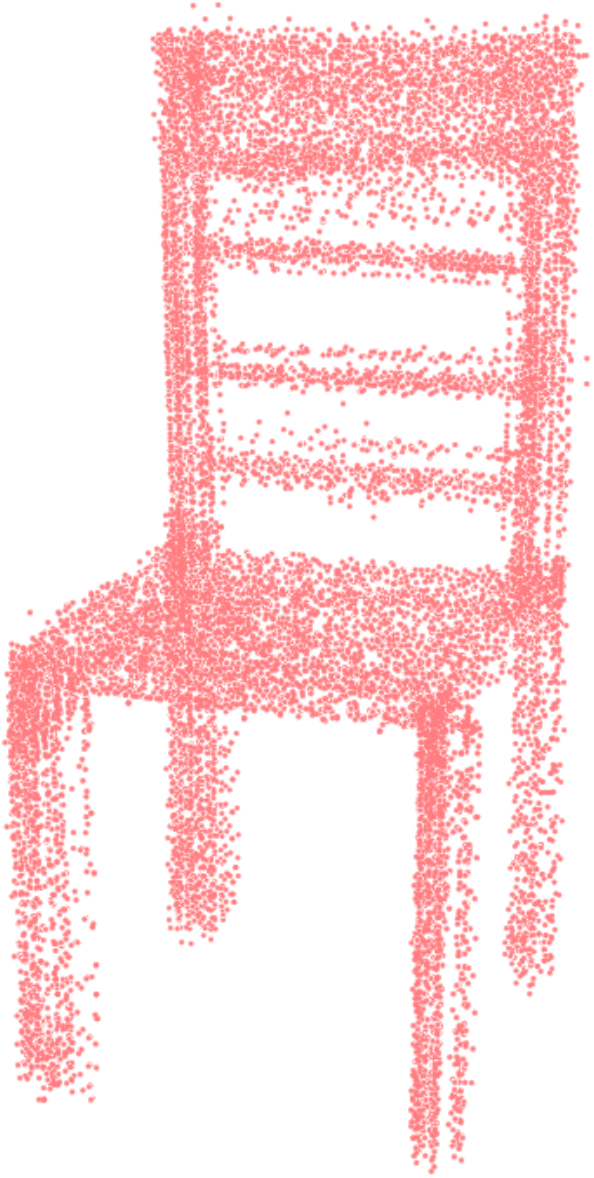}
	\end{subfigure}\hfill%
	\begin{subfigure}{\sunit}
		\centering
		\includegraphics[width=\sunitb]{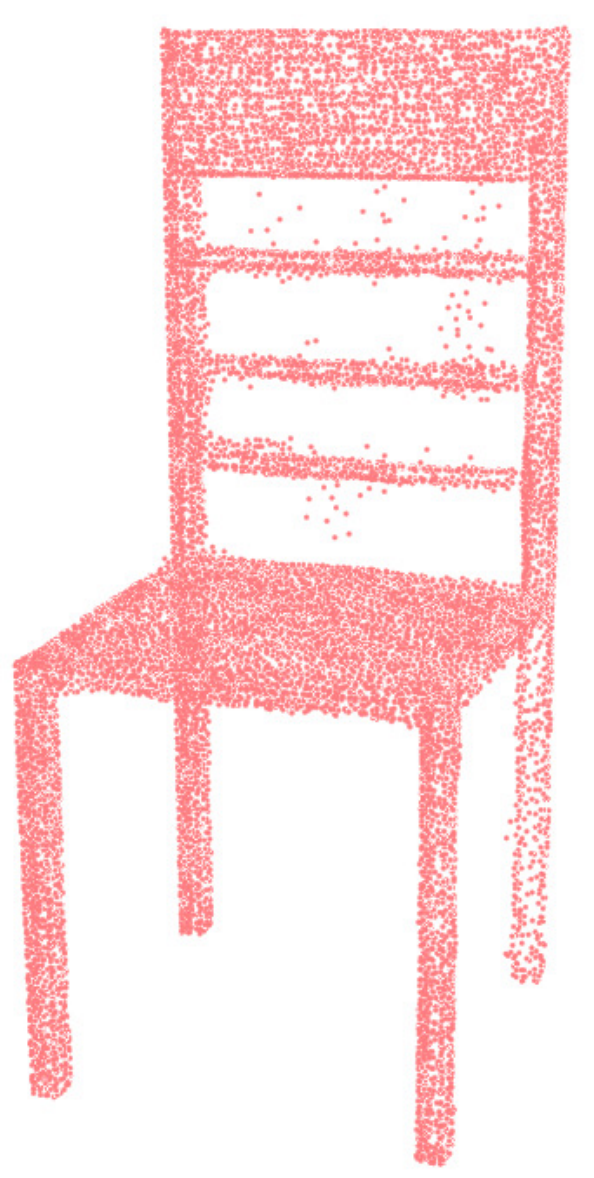}
	\end{subfigure}\hfill%
	\begin{subfigure}{\sunit}
		\centering
		\includegraphics[width=\sunitb]{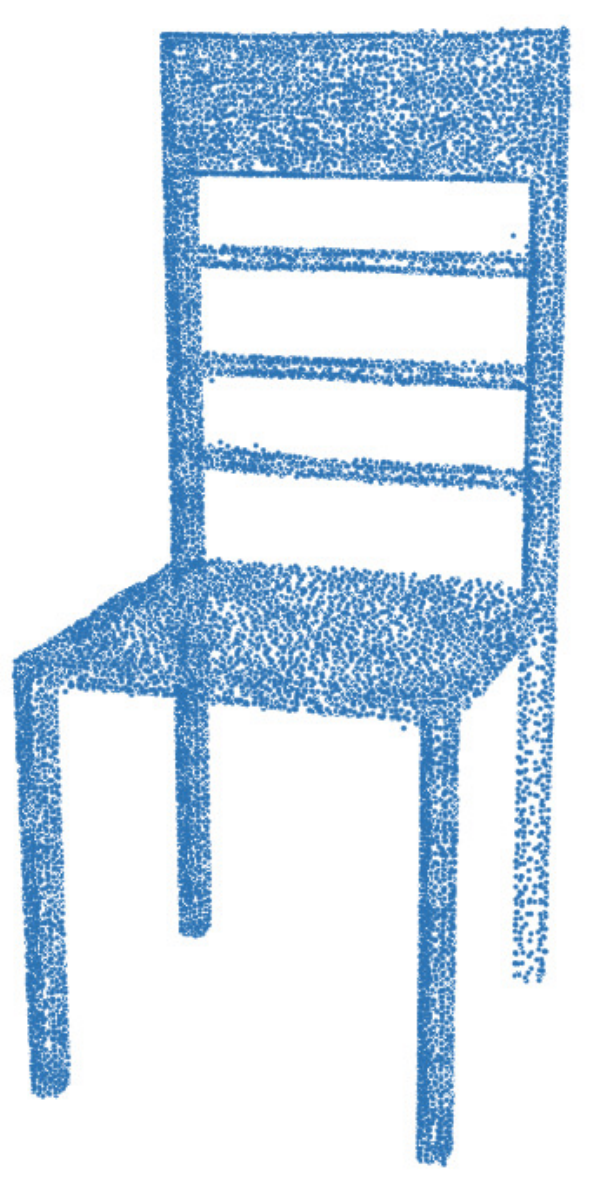}
	\end{subfigure}\hfill%
	\begin{subfigure}{\sunit}
		\centering
		\includegraphics[width=\sunitb]{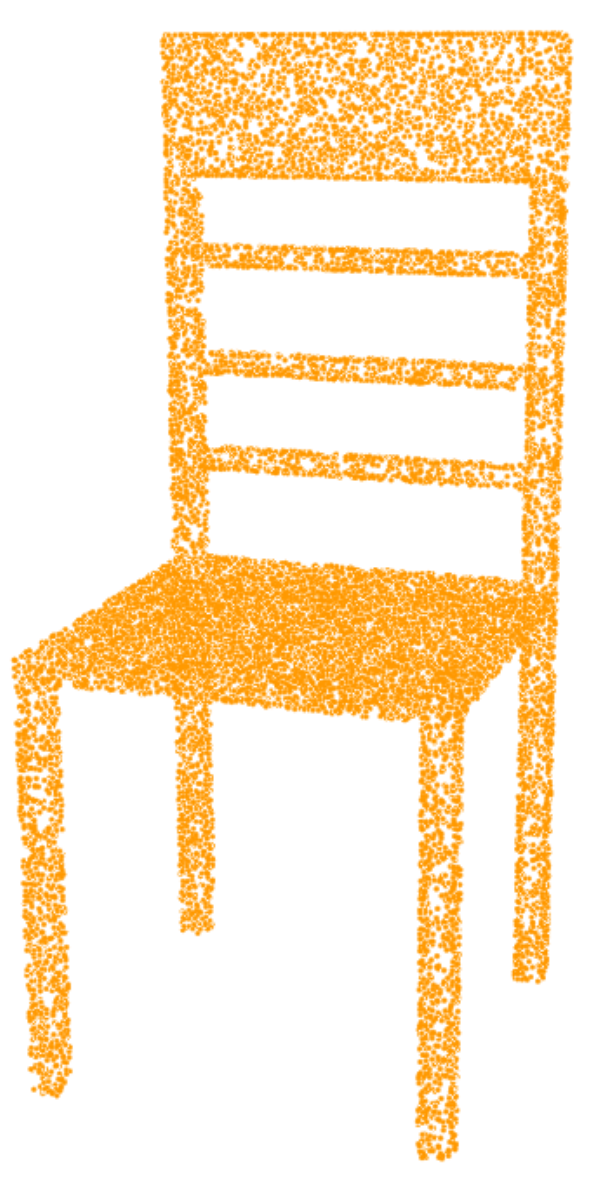}
	\end{subfigure}

	\begin{subfigure}{\sunit}
		\centering
		\includegraphics[width=\linewidth]{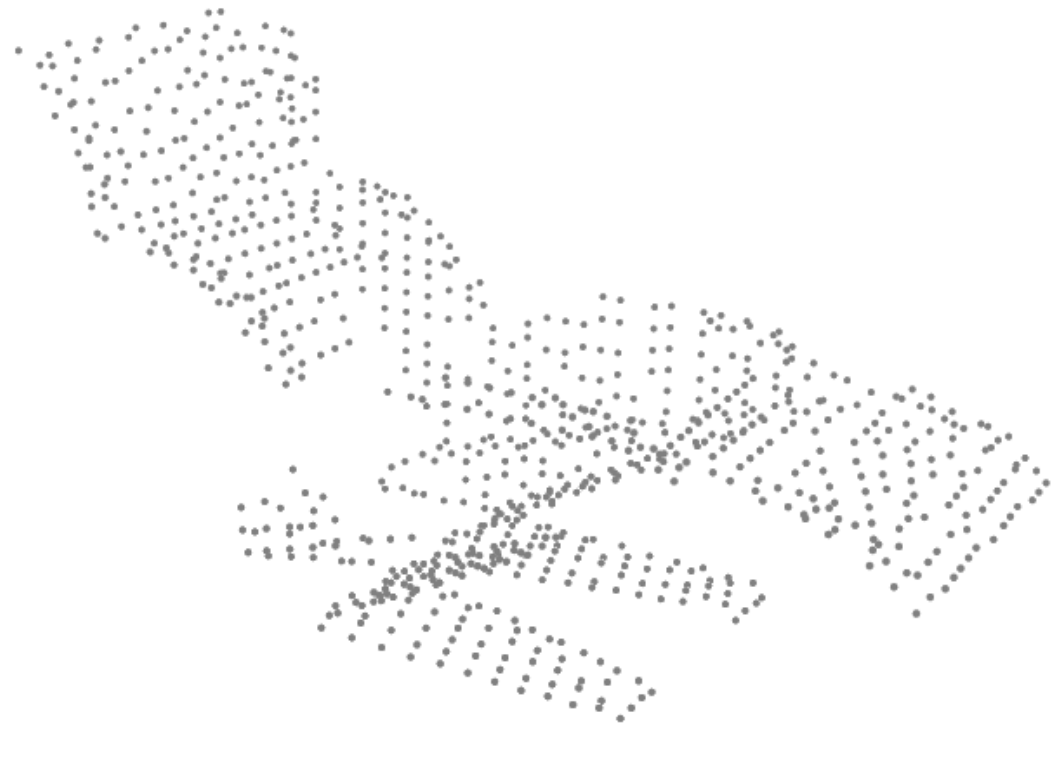}
	\end{subfigure}\hfill%
	\begin{subfigure}{\sunit}
		\centering
		\includegraphics[width=\linewidth]{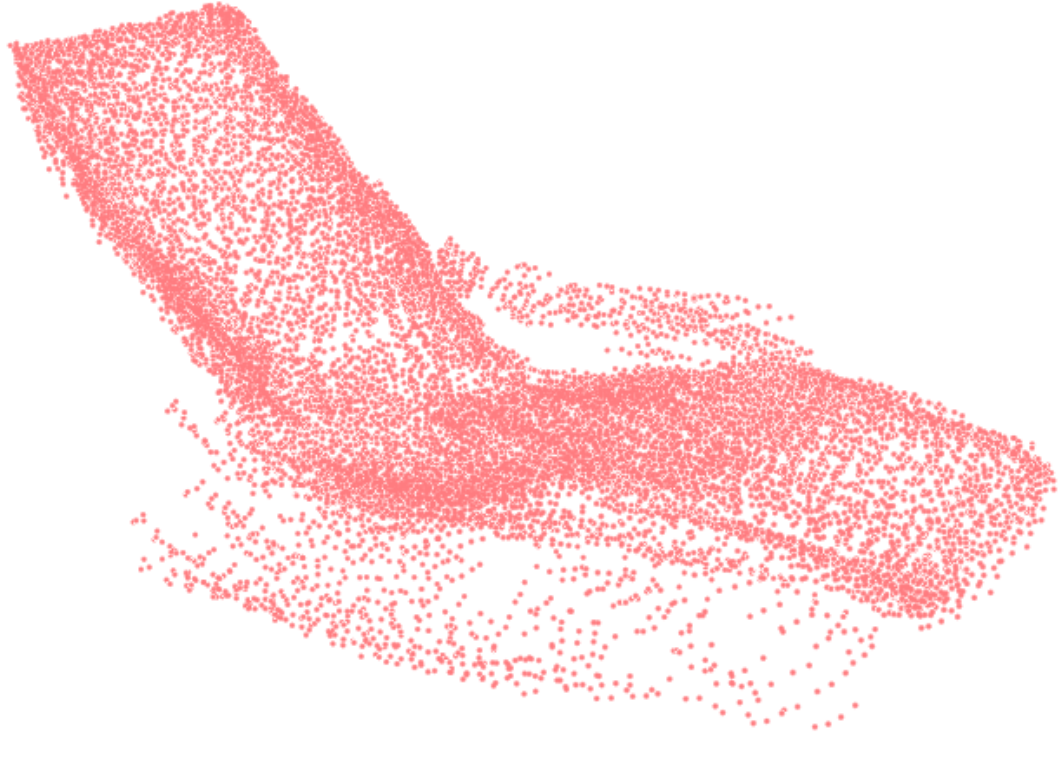}
	\end{subfigure}\hfill%
	\begin{subfigure}{\sunit}
		\centering
		\includegraphics[width=\linewidth]{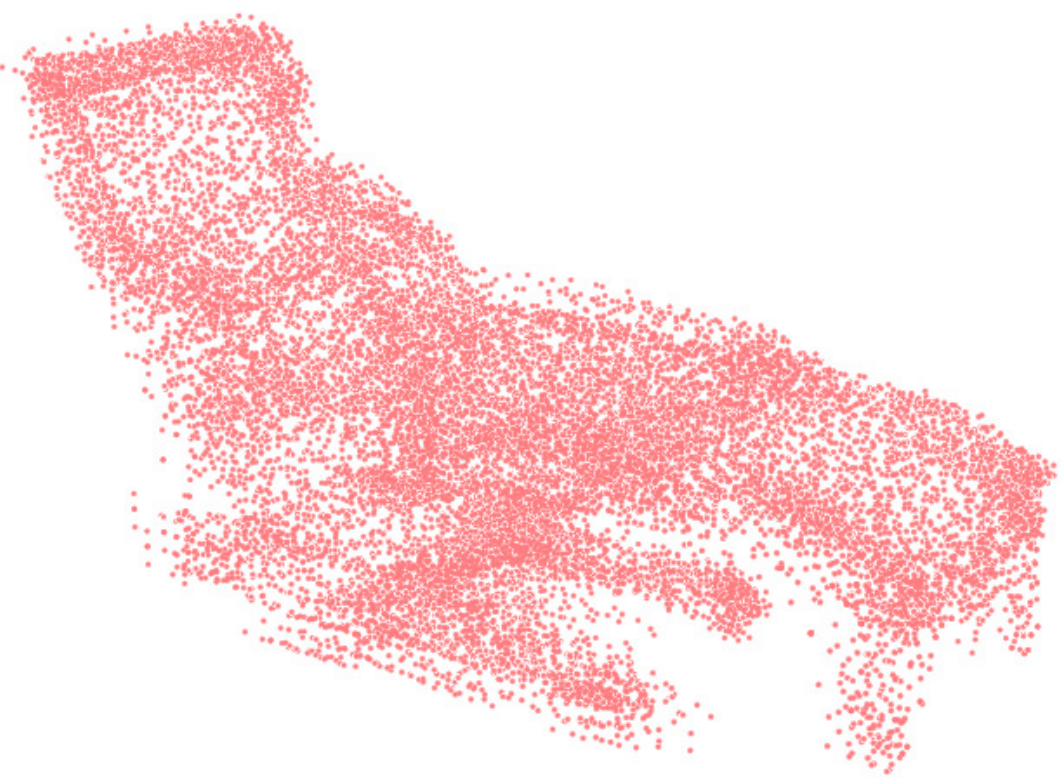}
	\end{subfigure}\hfill%
	\begin{subfigure}{\sunit}
		\centering
		\includegraphics[width=\linewidth]{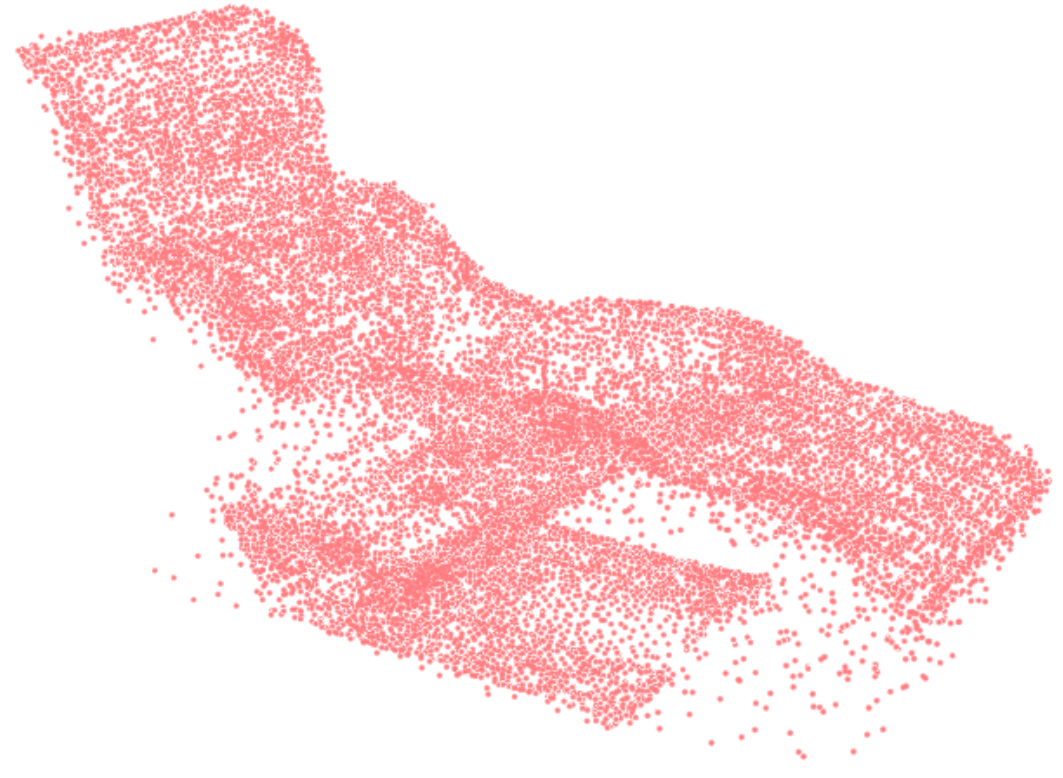}
	\end{subfigure}\hfill%
	\begin{subfigure}{\sunit}
		\centering
		\includegraphics[width=\linewidth]{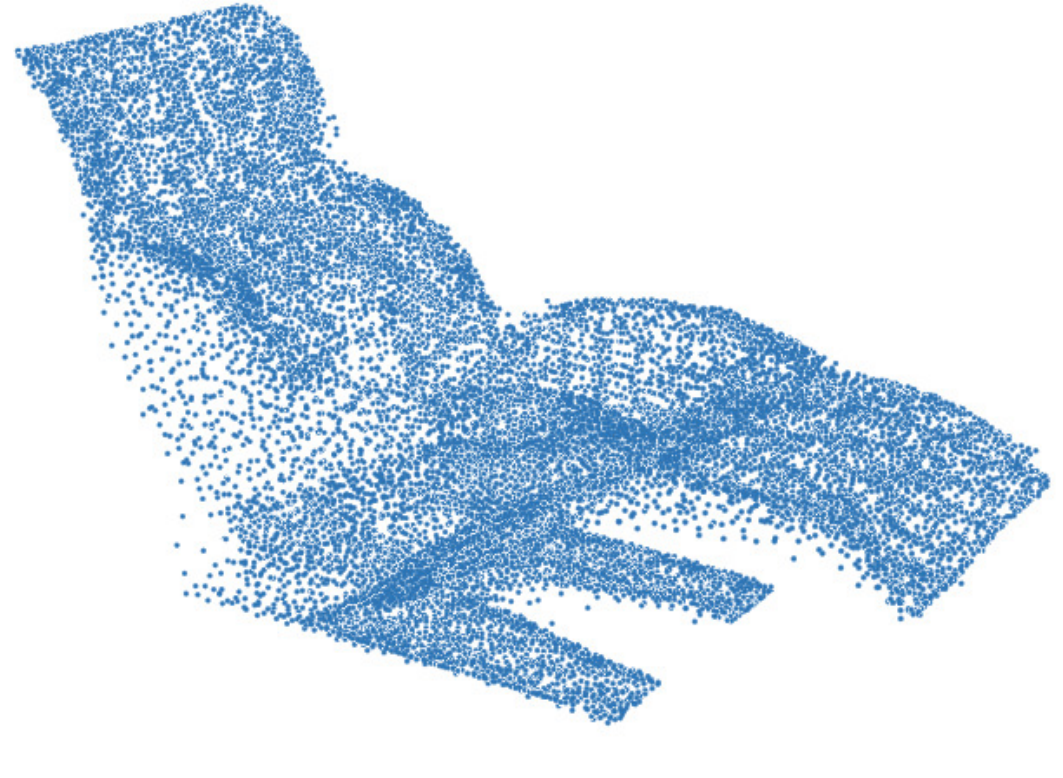}
	\end{subfigure}\hfill%
	\begin{subfigure}{\sunit}
		\centering
		\includegraphics[width=\linewidth]{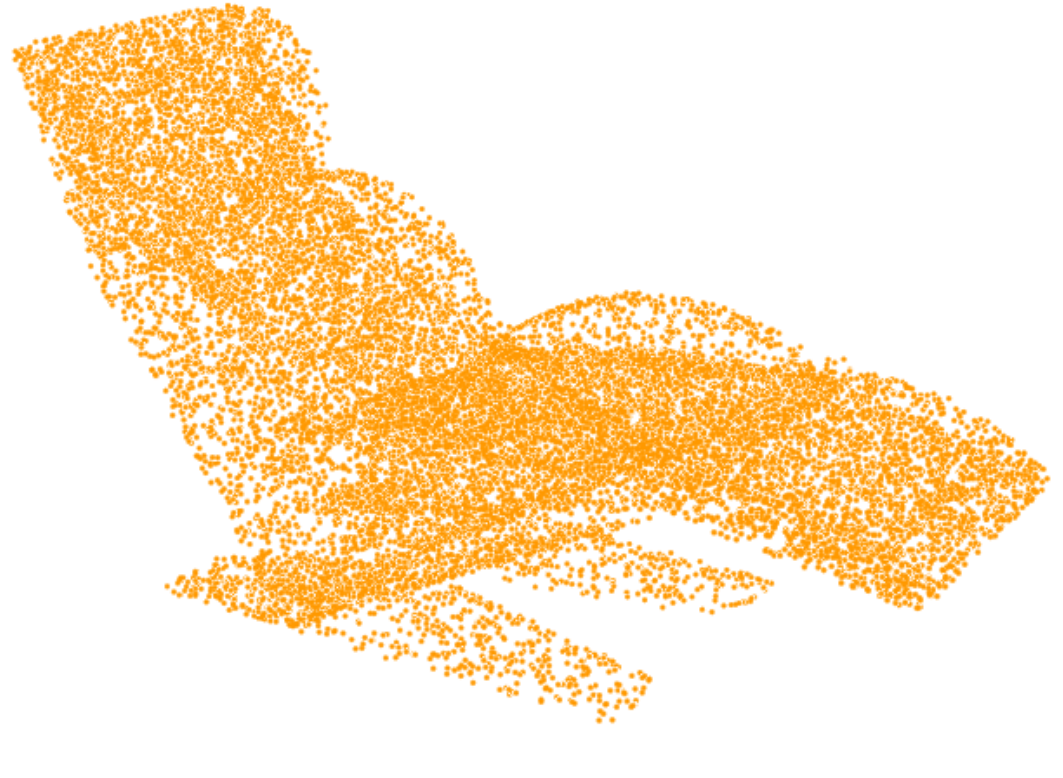}
	\end{subfigure}

	
	\begin{subfigure}{\sunit}
		\centering
		\includegraphics[width=\sunitc]{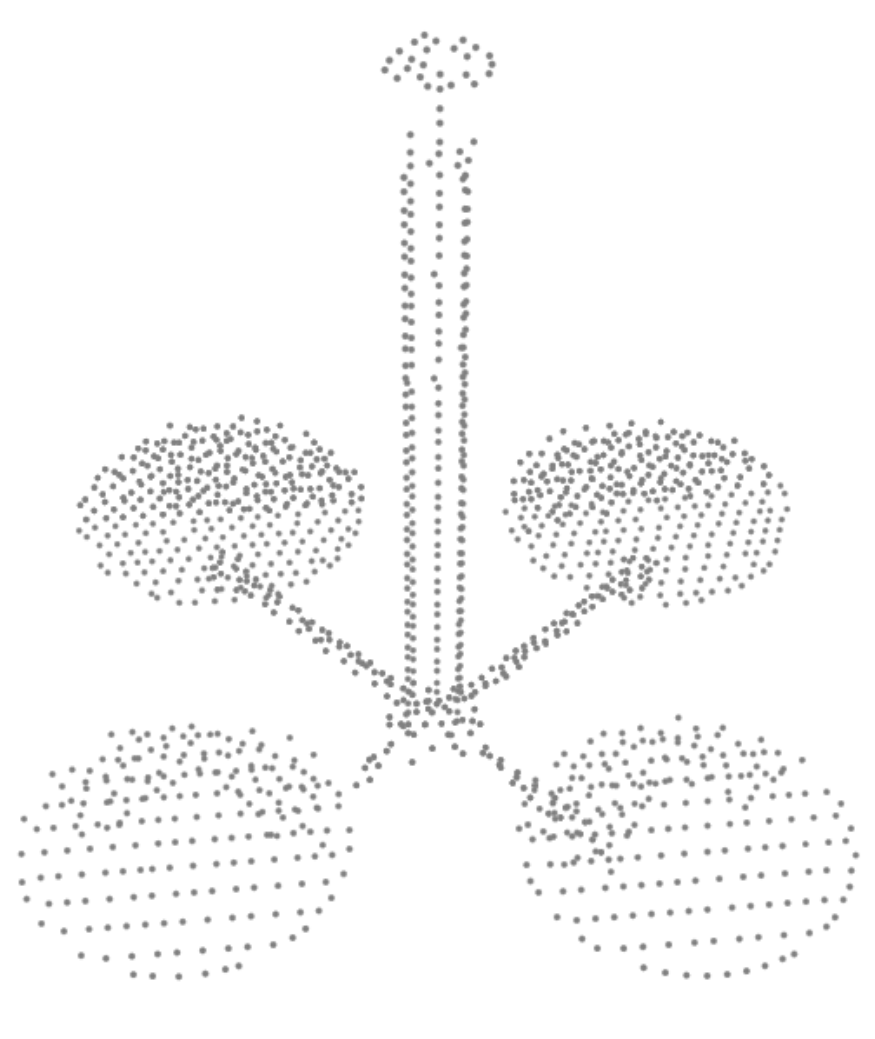}
	\end{subfigure}\hfill%
	\begin{subfigure}{\sunit}
		\centering
		\includegraphics[width=\sunitc]{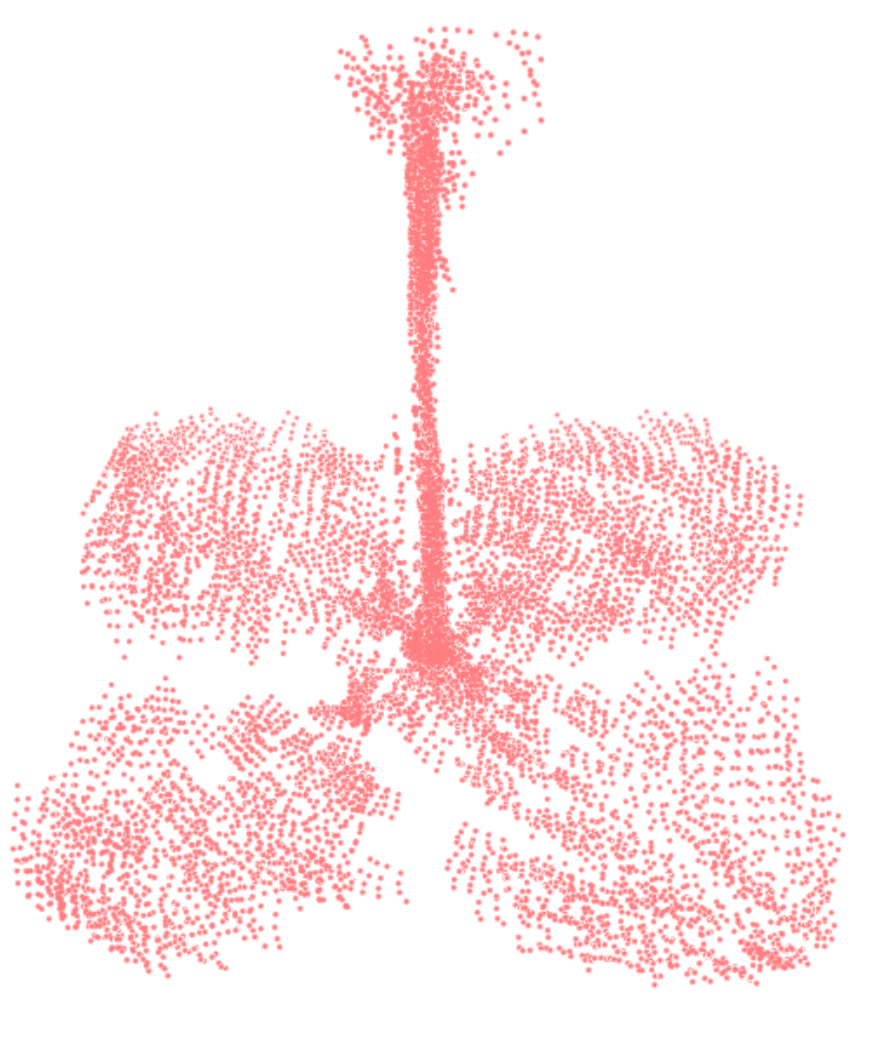}
	\end{subfigure}\hfill%
	\begin{subfigure}{\sunit}
		\centering
		\includegraphics[width=\sunitc]{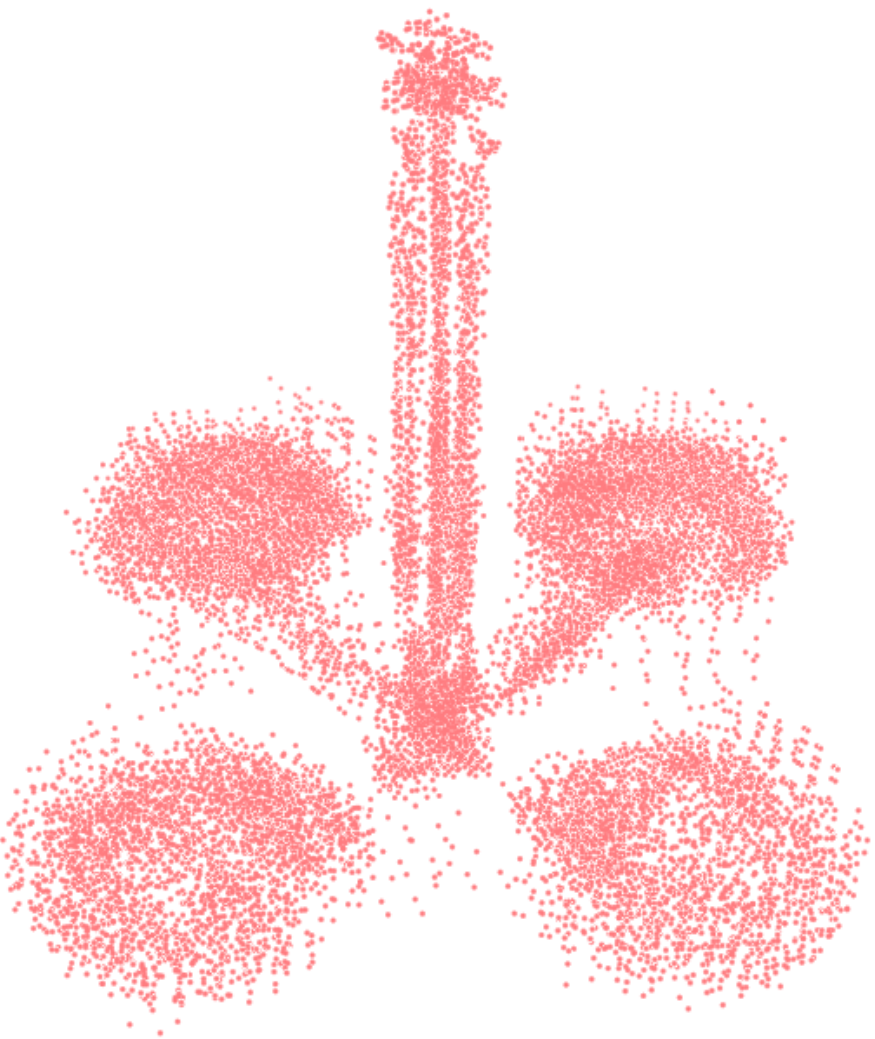}
	\end{subfigure}\hfill%
	\begin{subfigure}{\sunit}
		\centering
		\includegraphics[width=\sunitc]{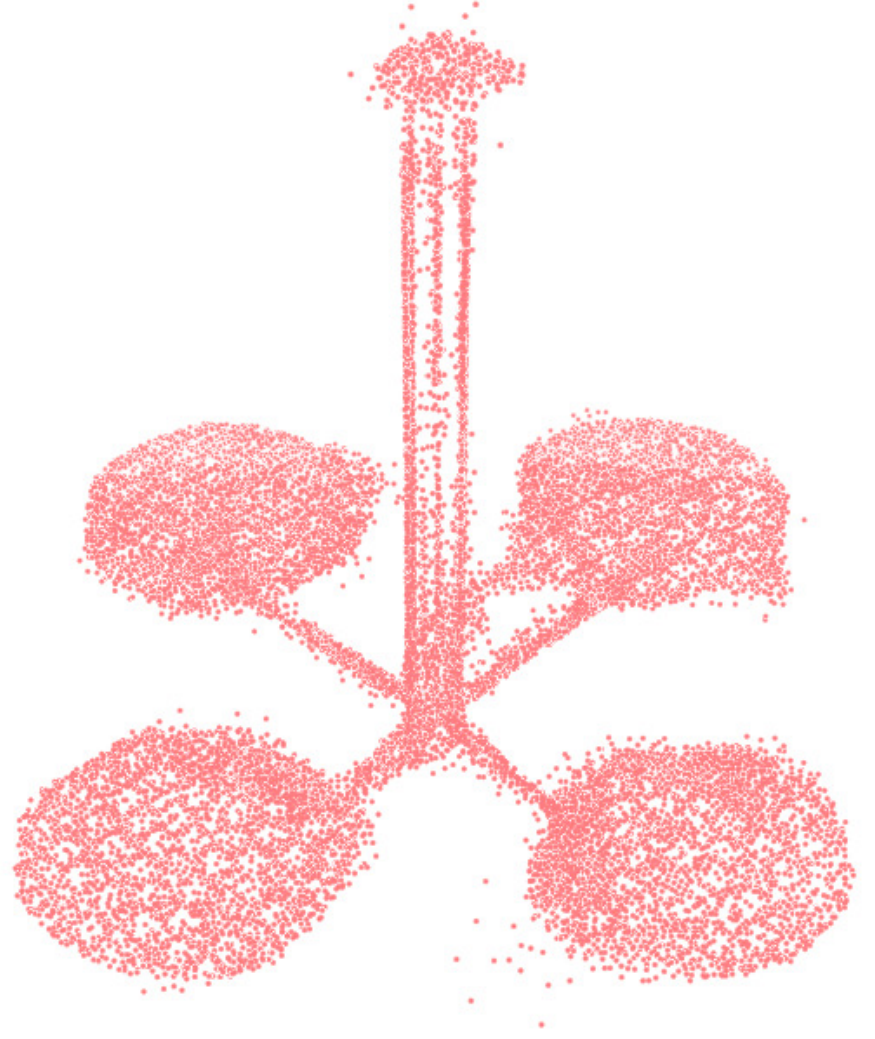}
	\end{subfigure}\hfill%
	\begin{subfigure}{\sunit}
		\centering
		\includegraphics[width=\sunitc]{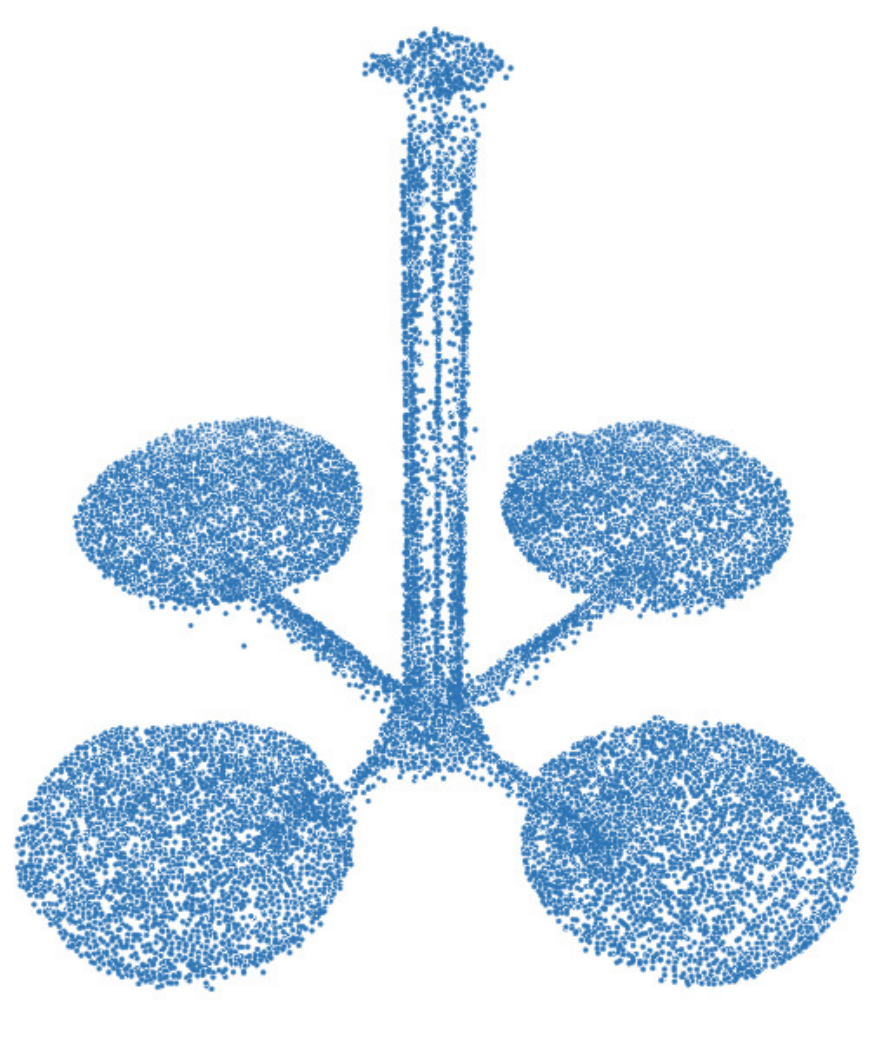}
	\end{subfigure}\hfill%
	\begin{subfigure}{\sunit}
		\centering
		\includegraphics[width=\sunitc]{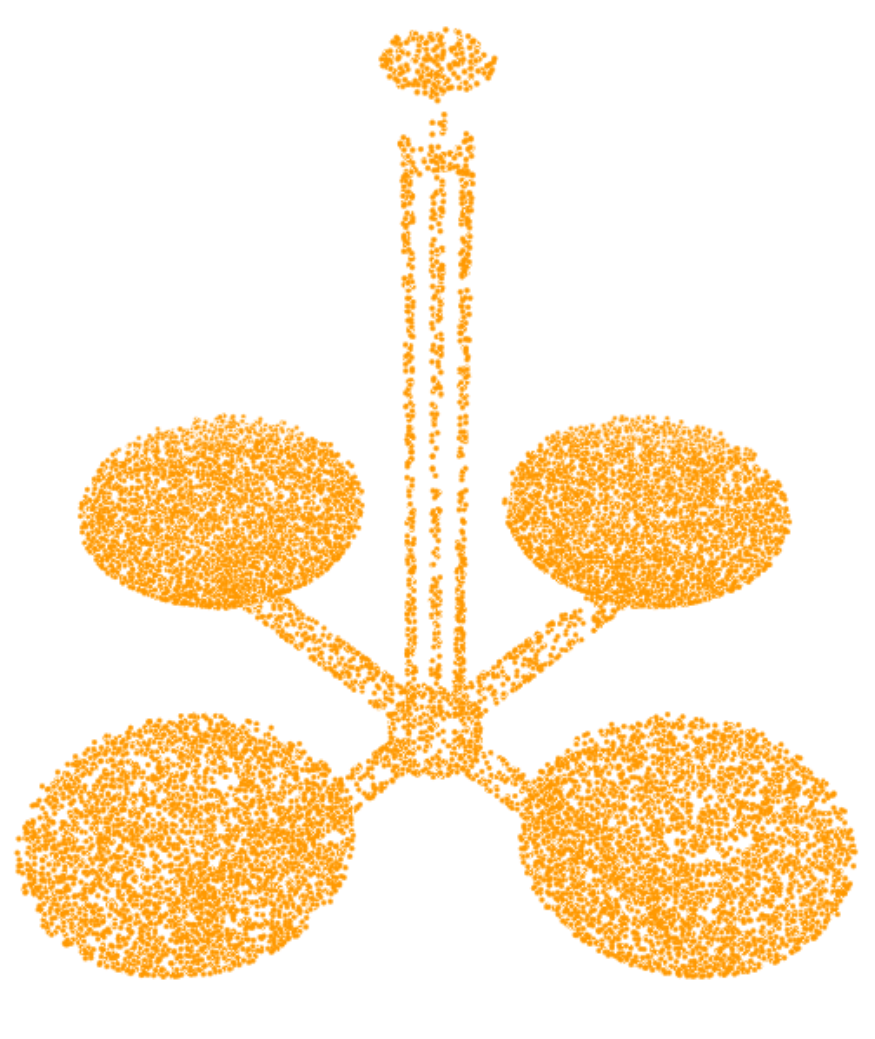}
	\end{subfigure}

	\begin{subfigure}{\sunit}
		\centering
		\includegraphics[width=\sunitc]{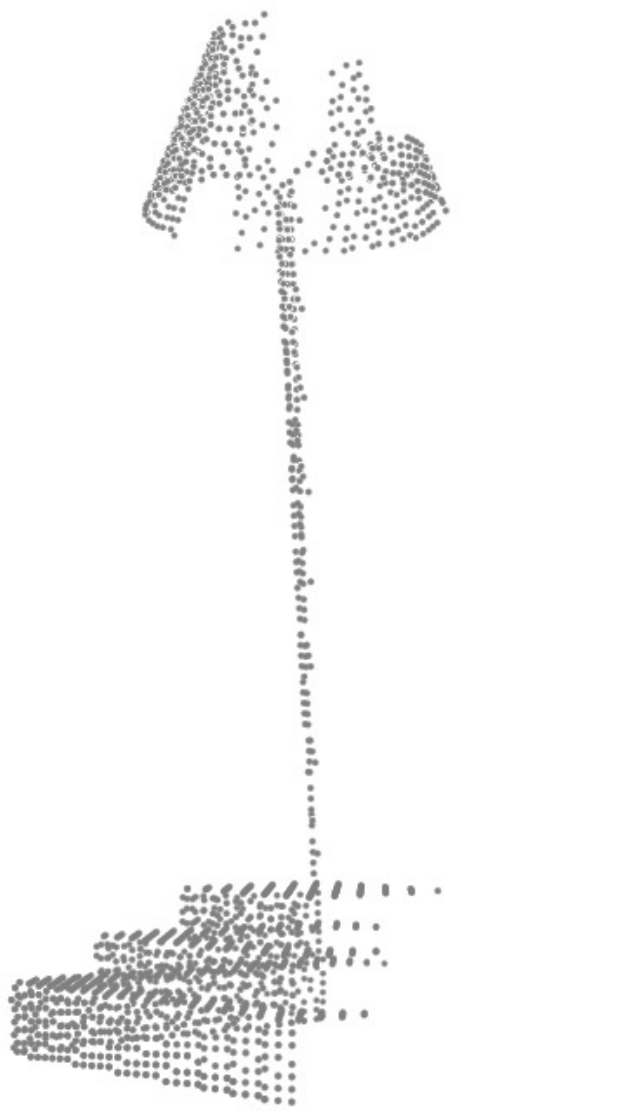}
	\end{subfigure}\hfill%
	\begin{subfigure}{\sunit}
		\centering
		\includegraphics[width=\sunitc]{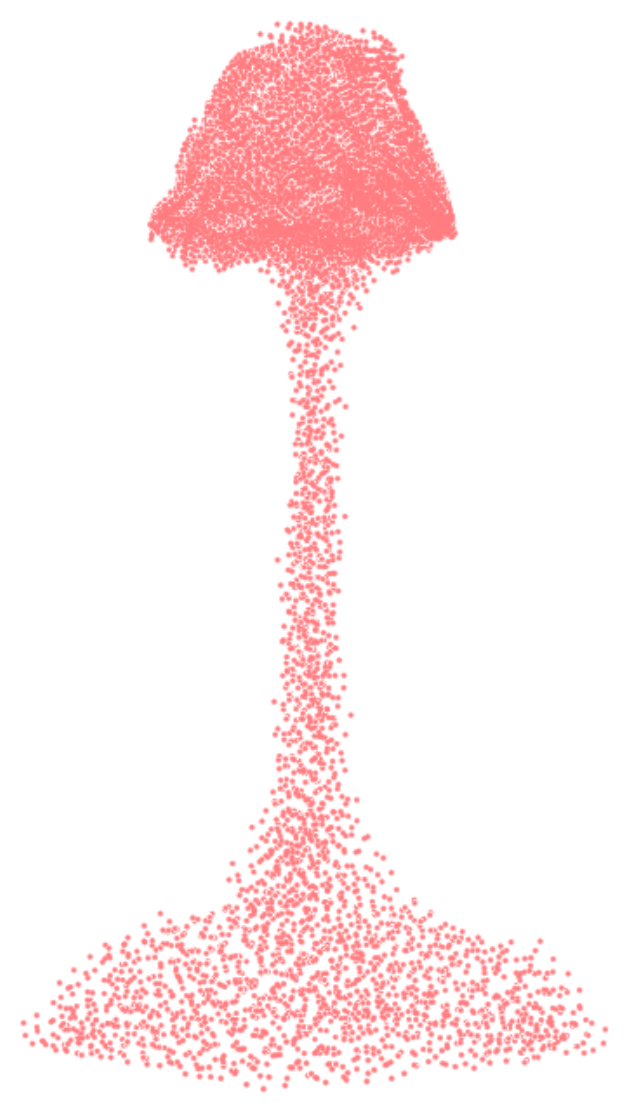}
	\end{subfigure}\hfill%
	\begin{subfigure}{\sunit}
		\centering
		\includegraphics[width=\sunitc]{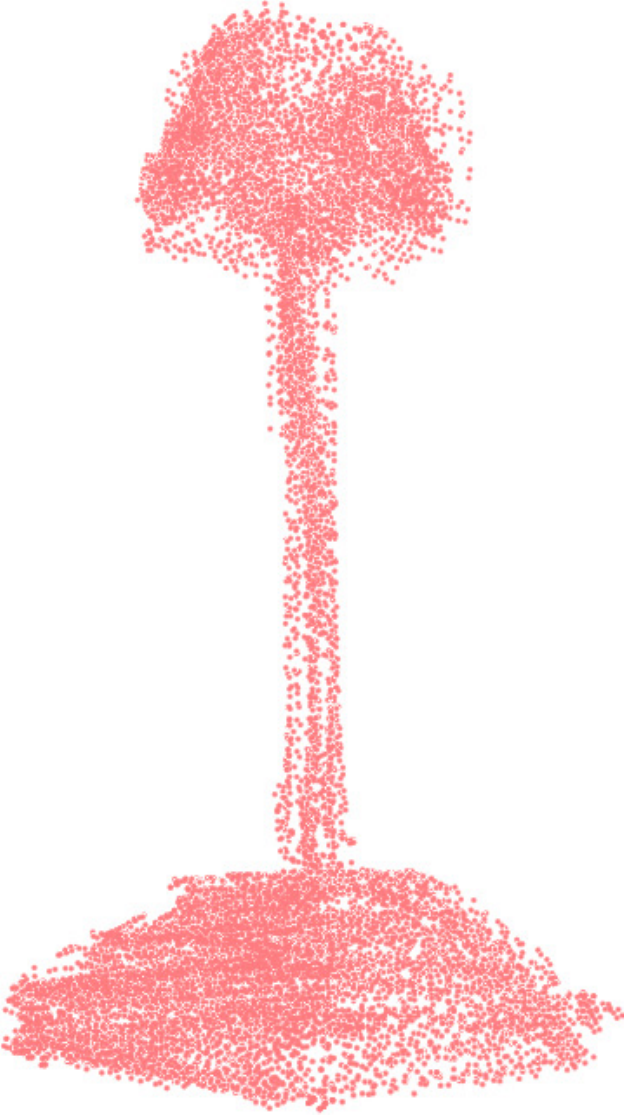}
	\end{subfigure}\hfill%
	\begin{subfigure}{\sunit}
		\centering
		\includegraphics[width=\sunitc]{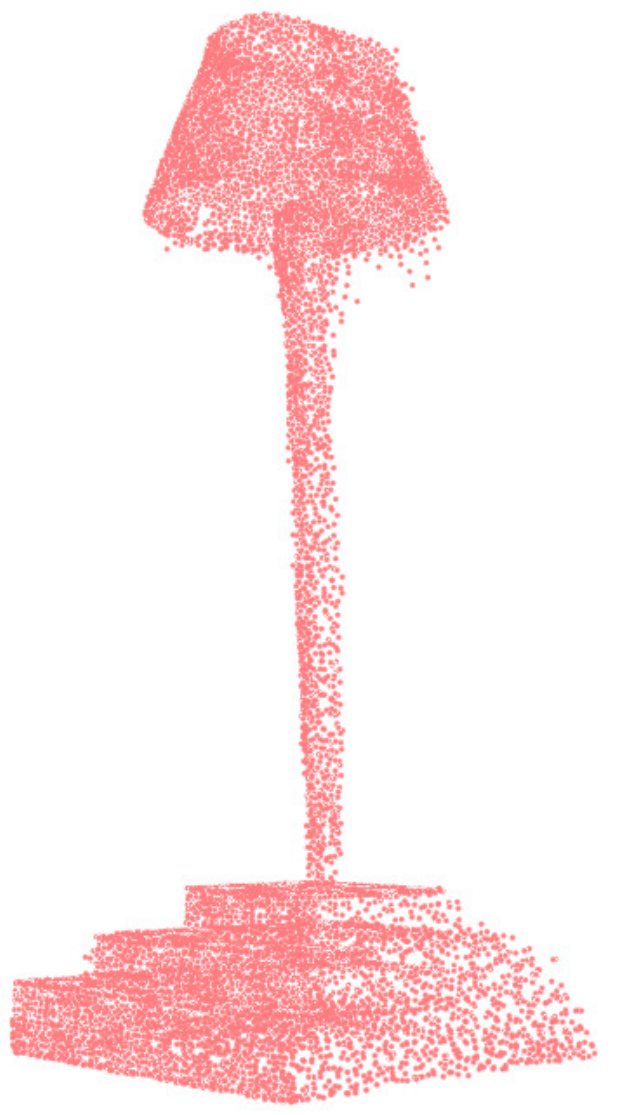}
	\end{subfigure}\hfill%
	\begin{subfigure}{\sunit}
		\centering
		\includegraphics[width=\sunitc]{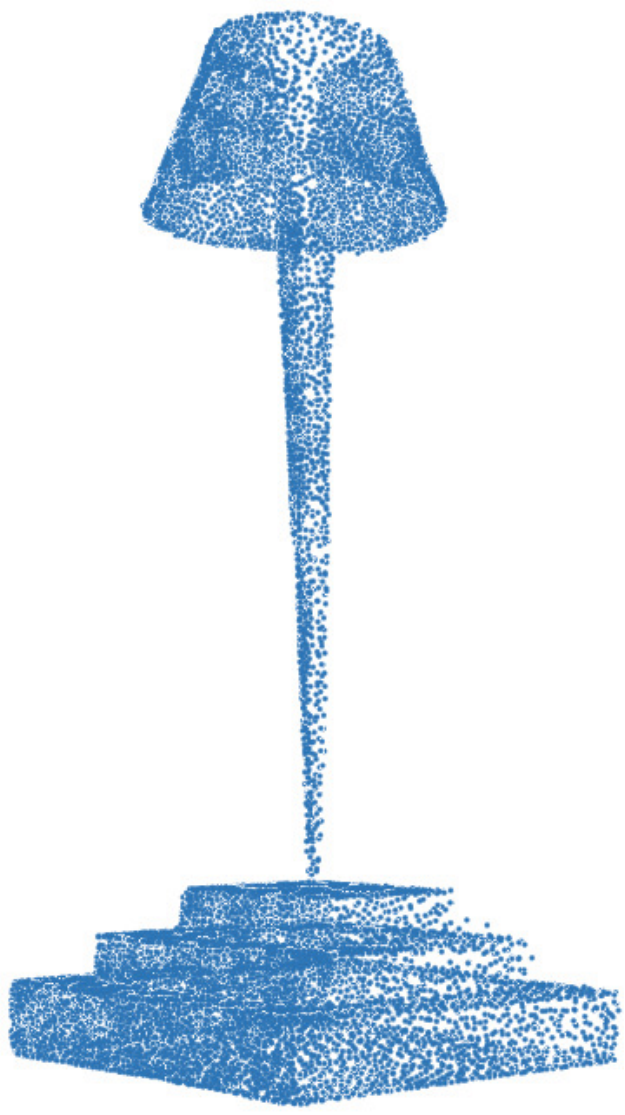}
	\end{subfigure}\hfill%
	\begin{subfigure}{\sunit}
		\centering
		\includegraphics[width=\sunitc]{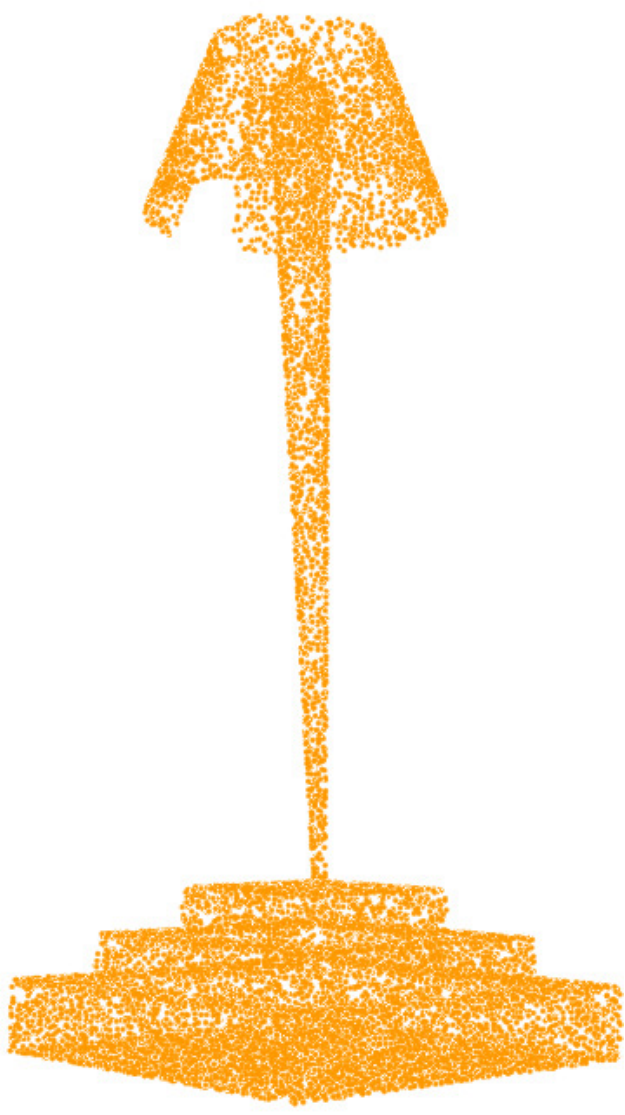}
	\end{subfigure}

	\vspace{5pt}
	
	\begin{subfigure}{\sunit}
		\centering
		\includegraphics[width=\sunitd]{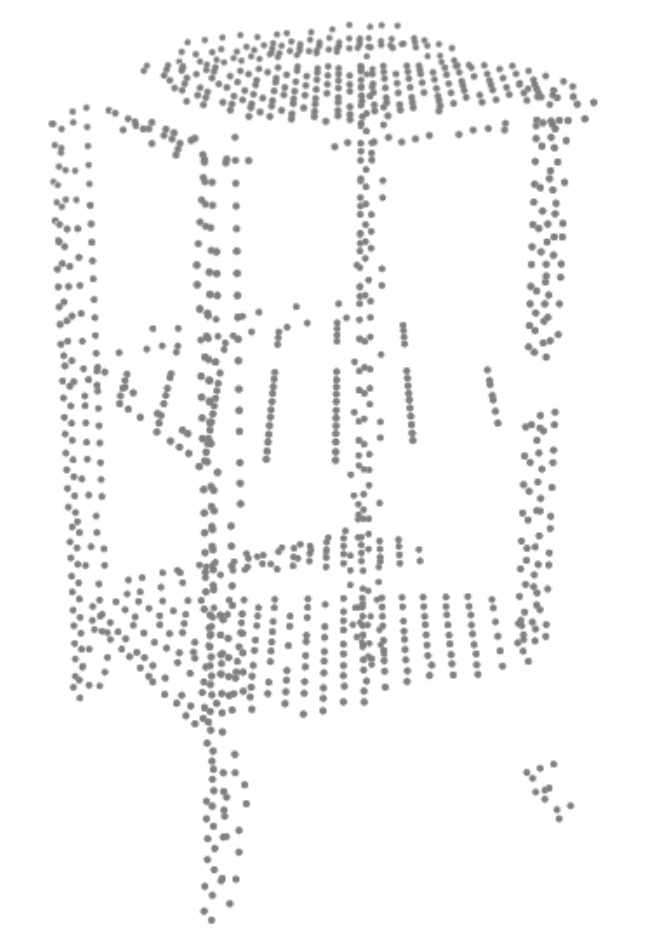}
	\end{subfigure}\hfill%
	\begin{subfigure}{\sunit}
		\centering
		\includegraphics[width=\sunitd]{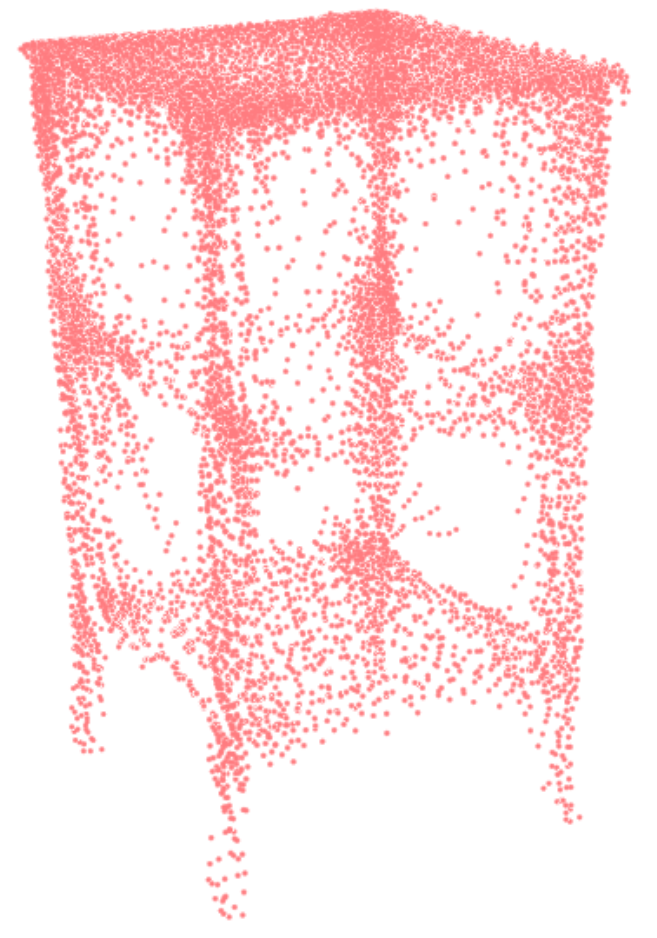}
	\end{subfigure}\hfill%
	\begin{subfigure}{\sunit}
		\centering
		\includegraphics[width=\sunitd]{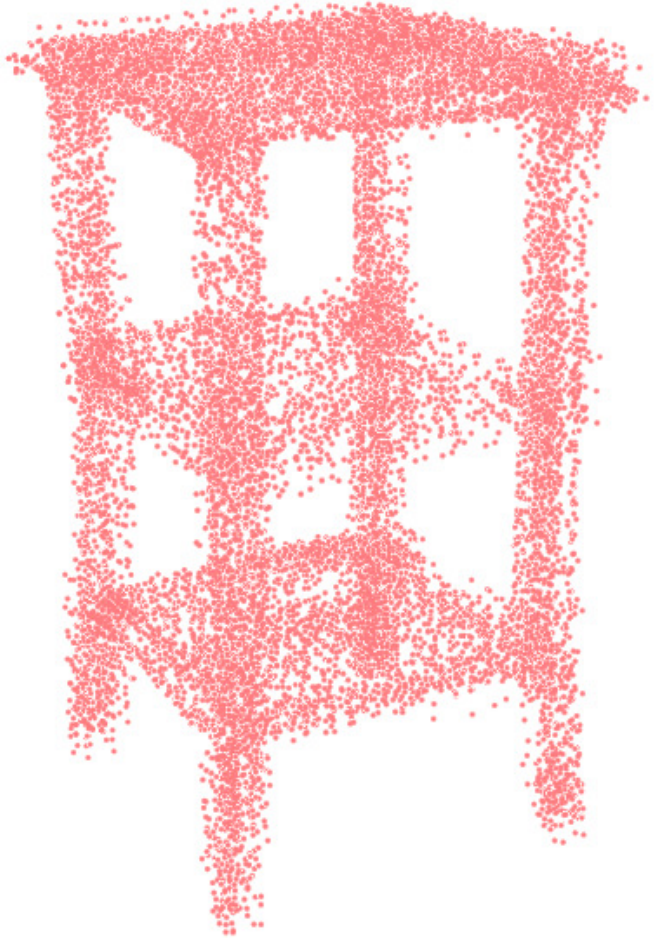}
	\end{subfigure}\hfill%
	\begin{subfigure}{\sunit}
		\centering
		\includegraphics[width=\sunitd]{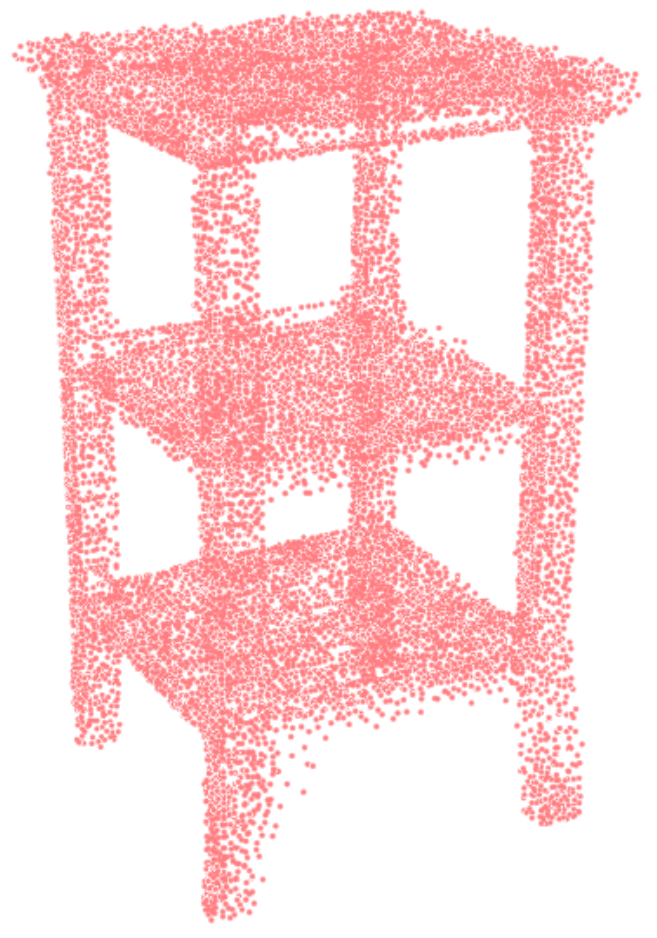}
	\end{subfigure}\hfill%
	\begin{subfigure}{\sunit}
		\centering
		\includegraphics[width=\sunitd]{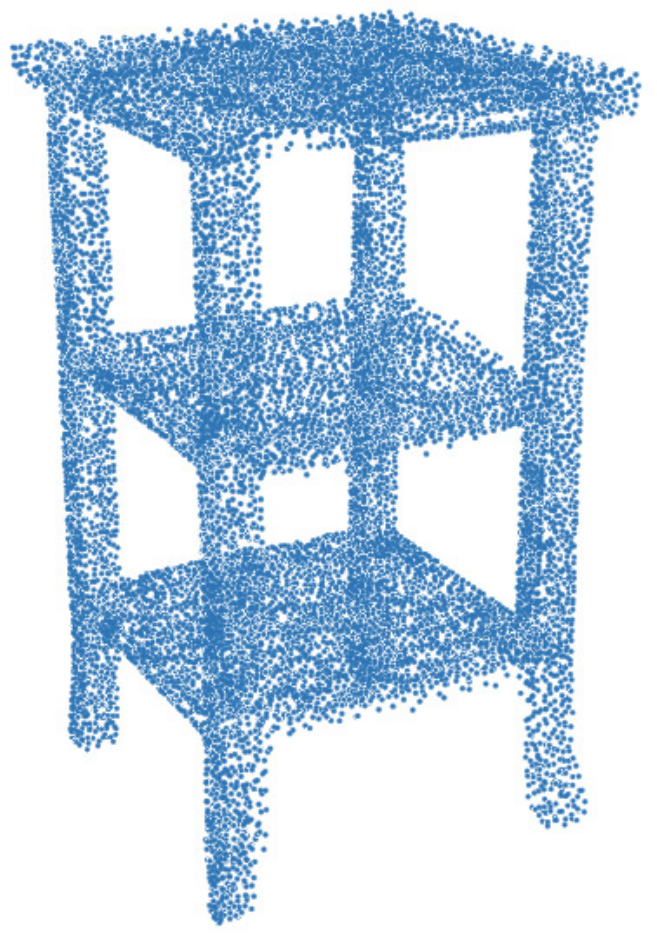}
	\end{subfigure}\hfill%
	\begin{subfigure}{\sunit}
		\centering
		\includegraphics[width=\sunitd]{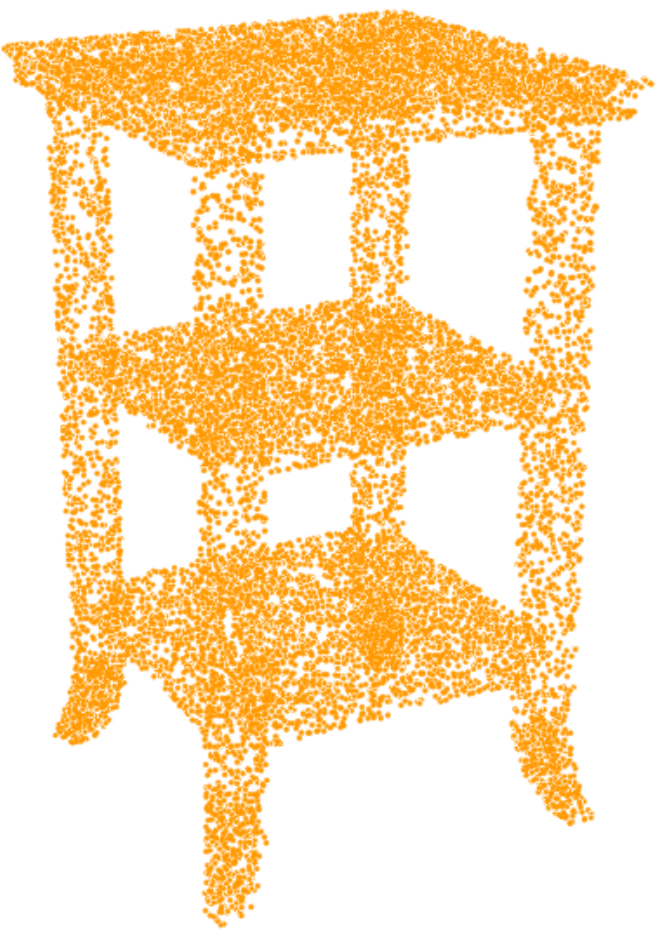}
	\end{subfigure}

	\vspace{5pt}
	
	\begin{subfigure}{\sunit}
		\centering
		\includegraphics[width=\linewidth]{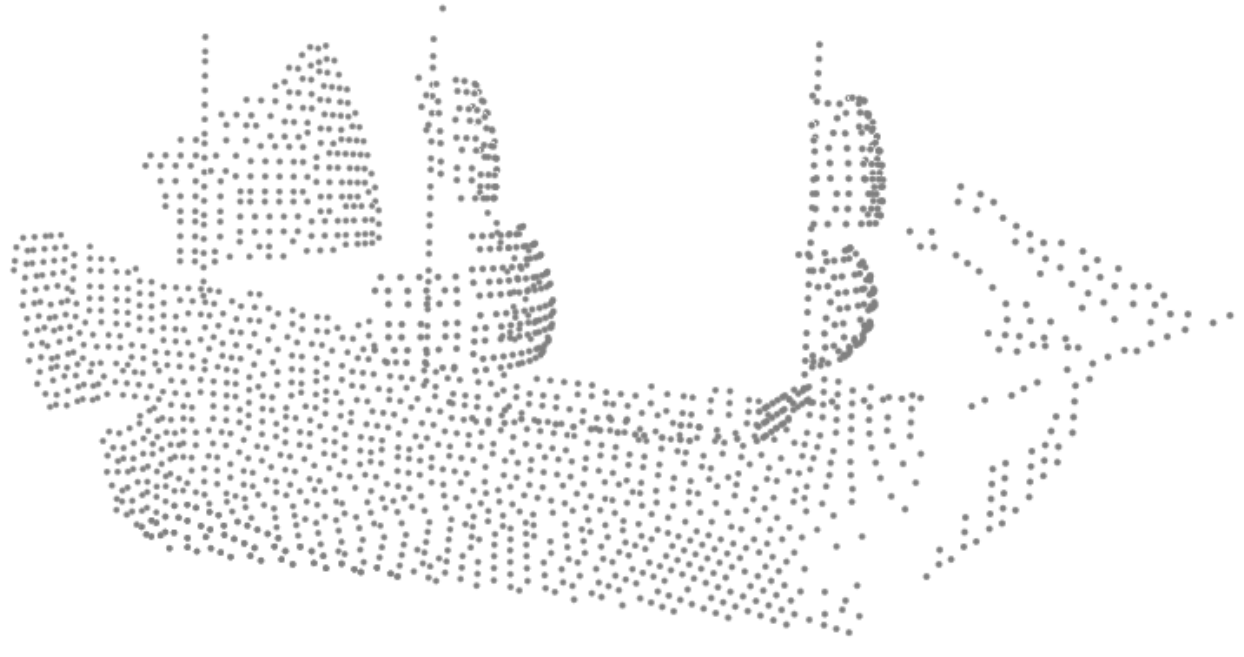}
	\end{subfigure}\hfill%
	\begin{subfigure}{\sunit}
		\centering
		\includegraphics[width=\linewidth]{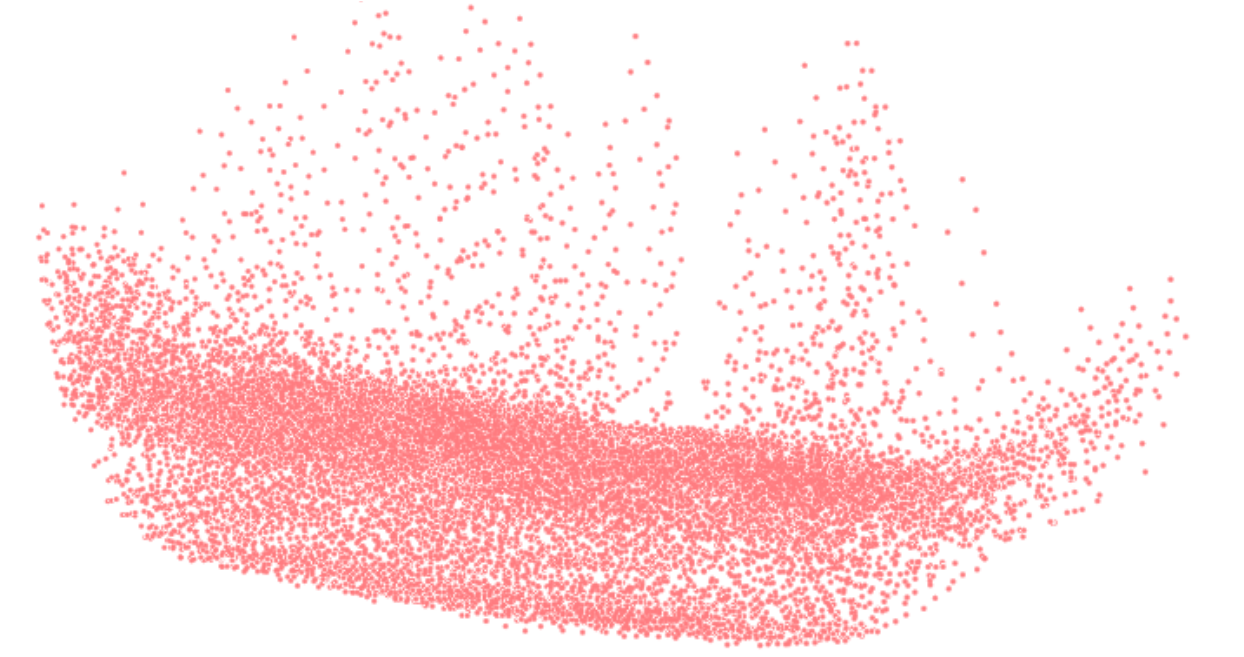}
	\end{subfigure}\hfill%
	\begin{subfigure}{\sunit}
		\centering
		\includegraphics[width=\linewidth]{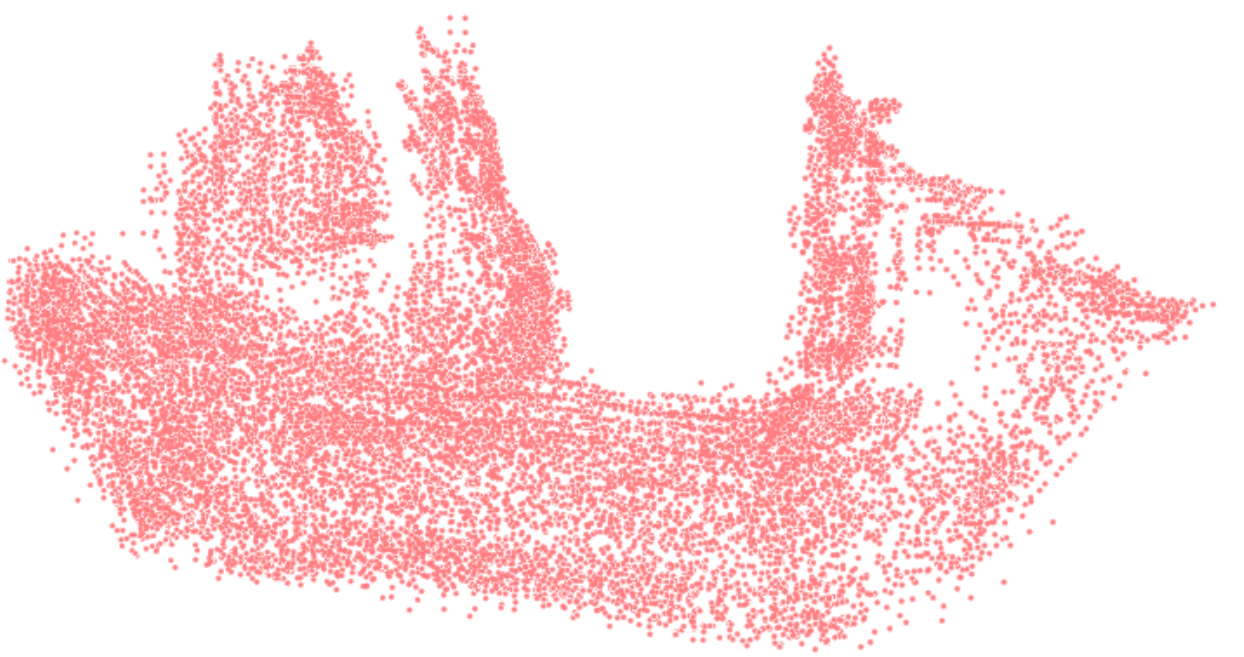}
	\end{subfigure}\hfill%
	\begin{subfigure}{\sunit}
		\centering
		\includegraphics[width=\linewidth]{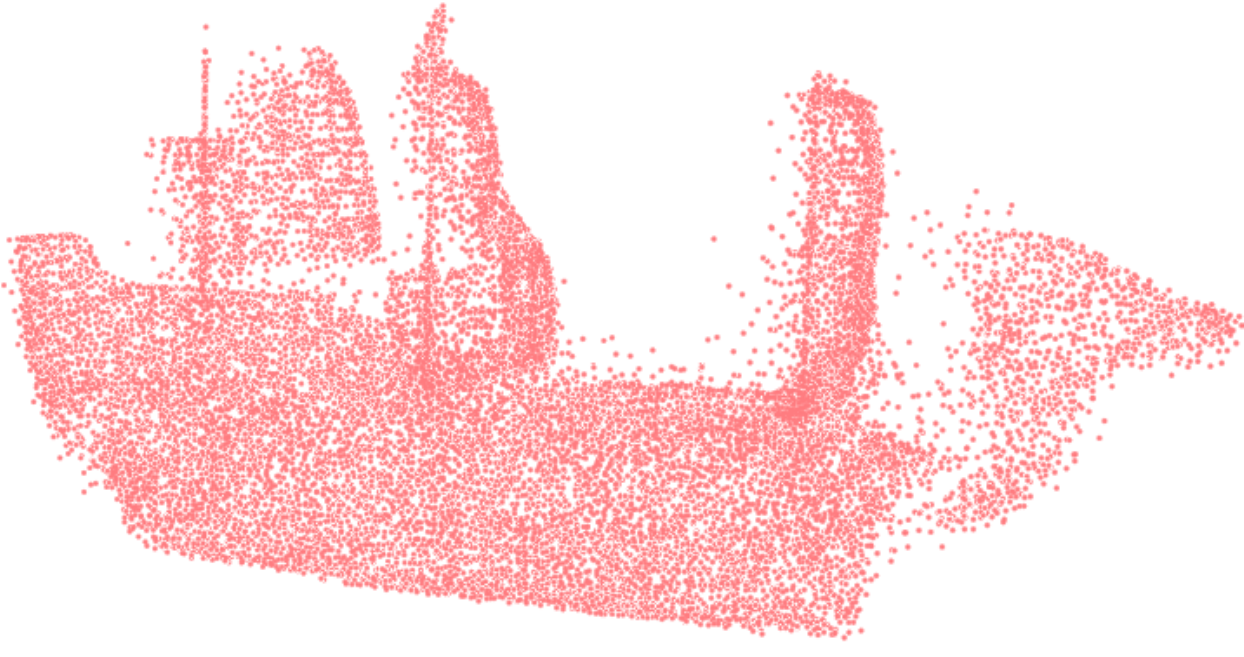}
	\end{subfigure}\hfill%
	\begin{subfigure}{\sunit}
		\centering
		\includegraphics[width=\linewidth]{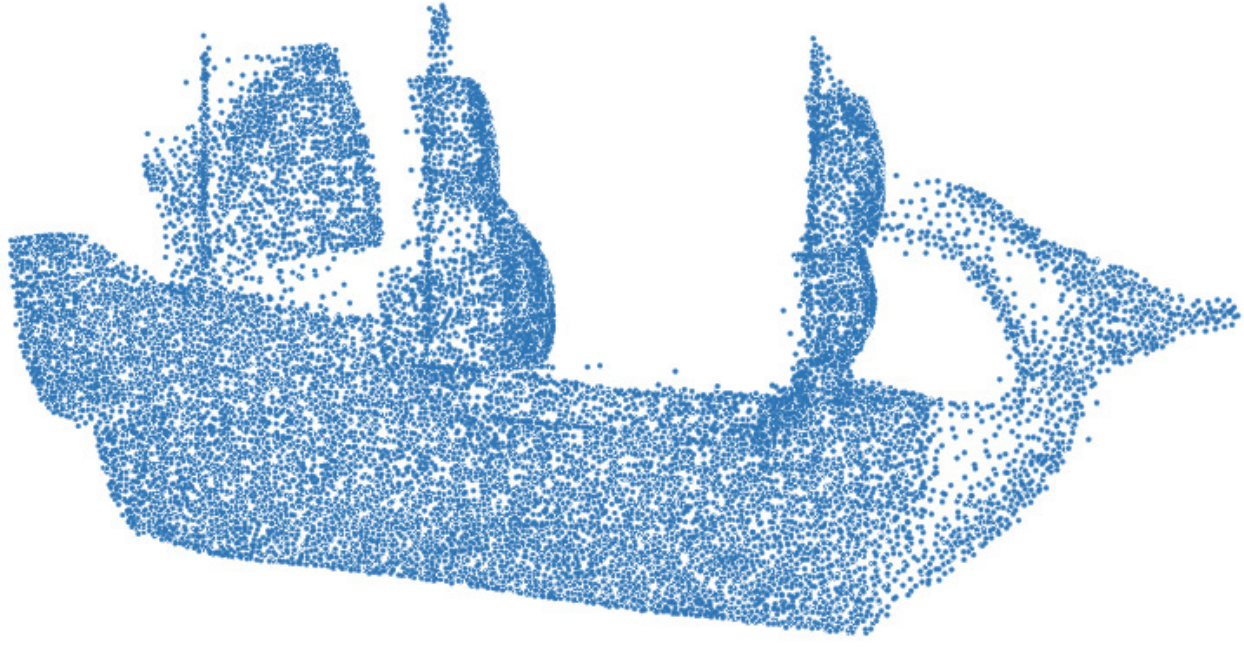}
	\end{subfigure}\hfill%
	\begin{subfigure}{\sunit}
		\centering
		\includegraphics[width=\linewidth]{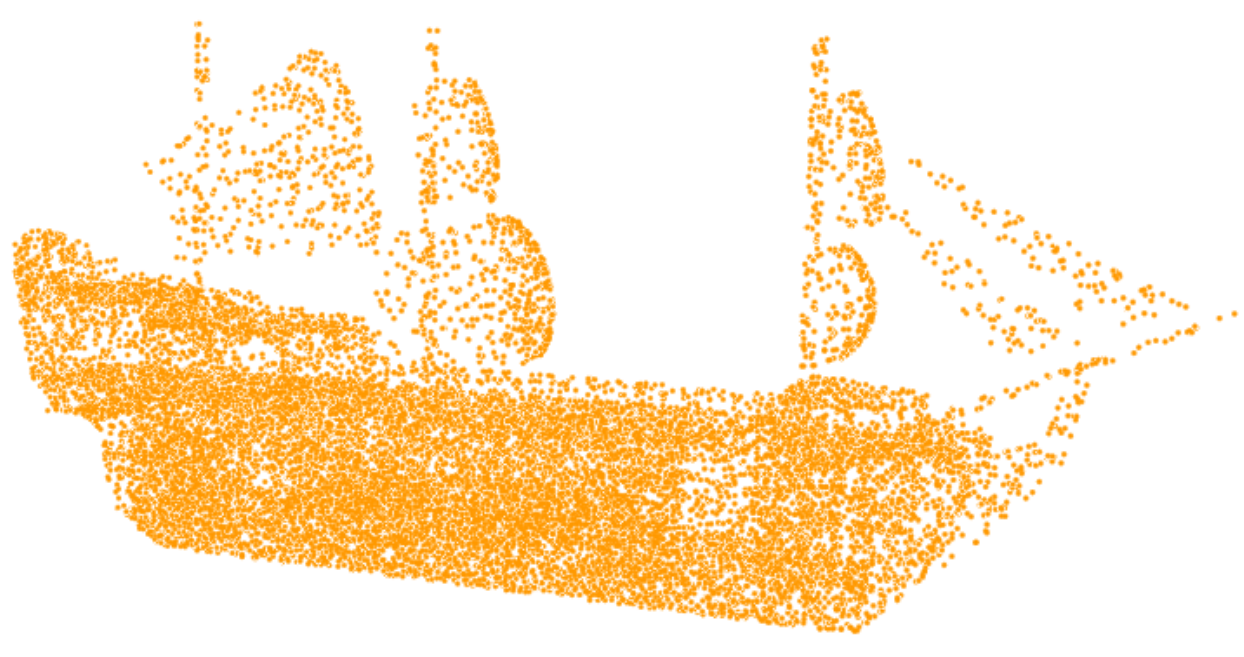}
	\end{subfigure}

	\vspace{5pt}

	\begin{subfigure}{\sunit}
		\centering
		\includegraphics[width=\linewidth]{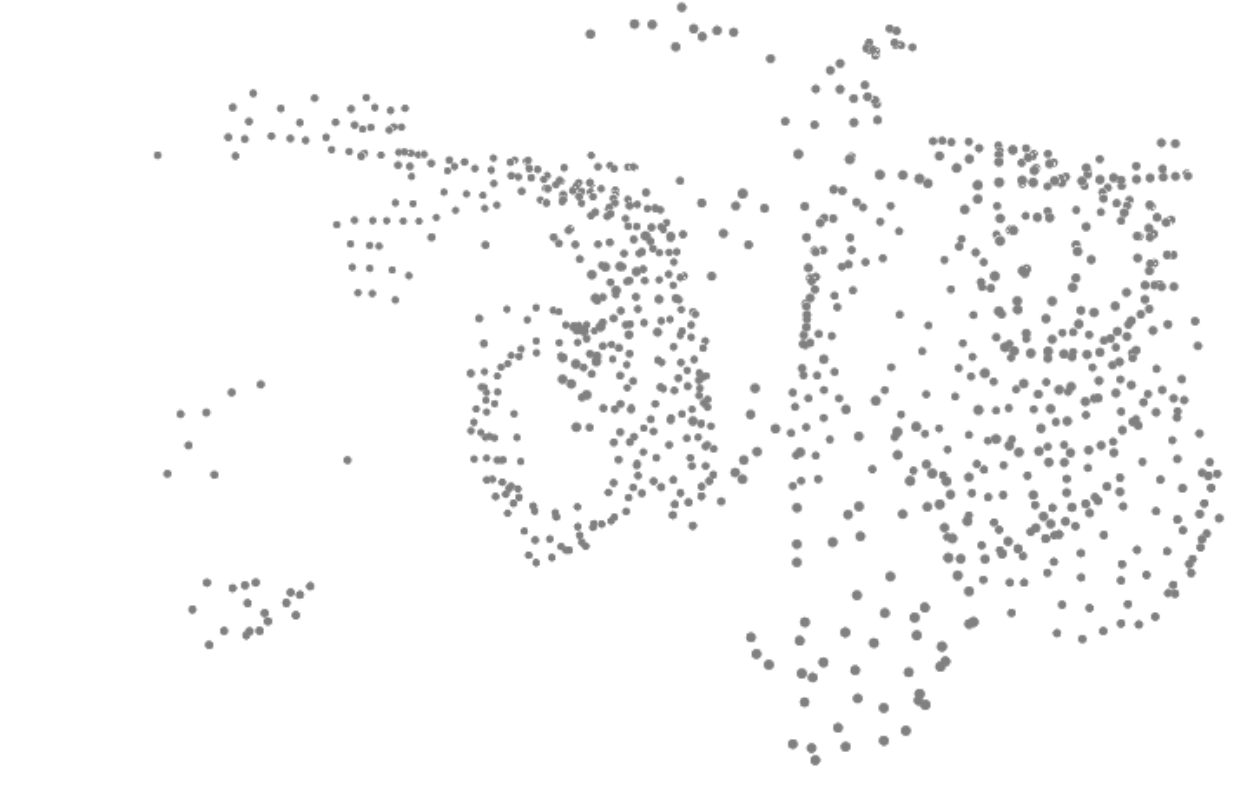}
	\end{subfigure}\hfill%
	\begin{subfigure}{\sunit}
		\centering
		\includegraphics[width=\linewidth]{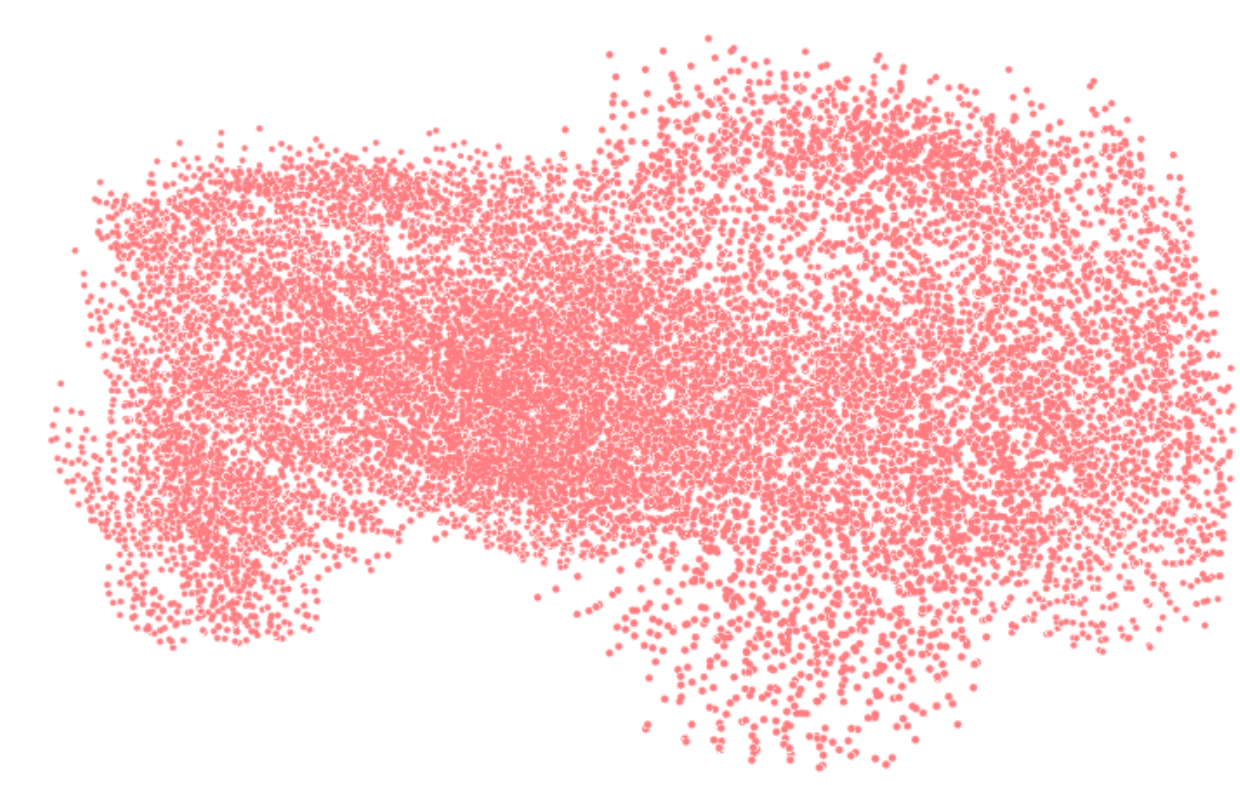}
	\end{subfigure}\hfill%
	\begin{subfigure}{\sunit}
		\centering
		\includegraphics[width=\linewidth]{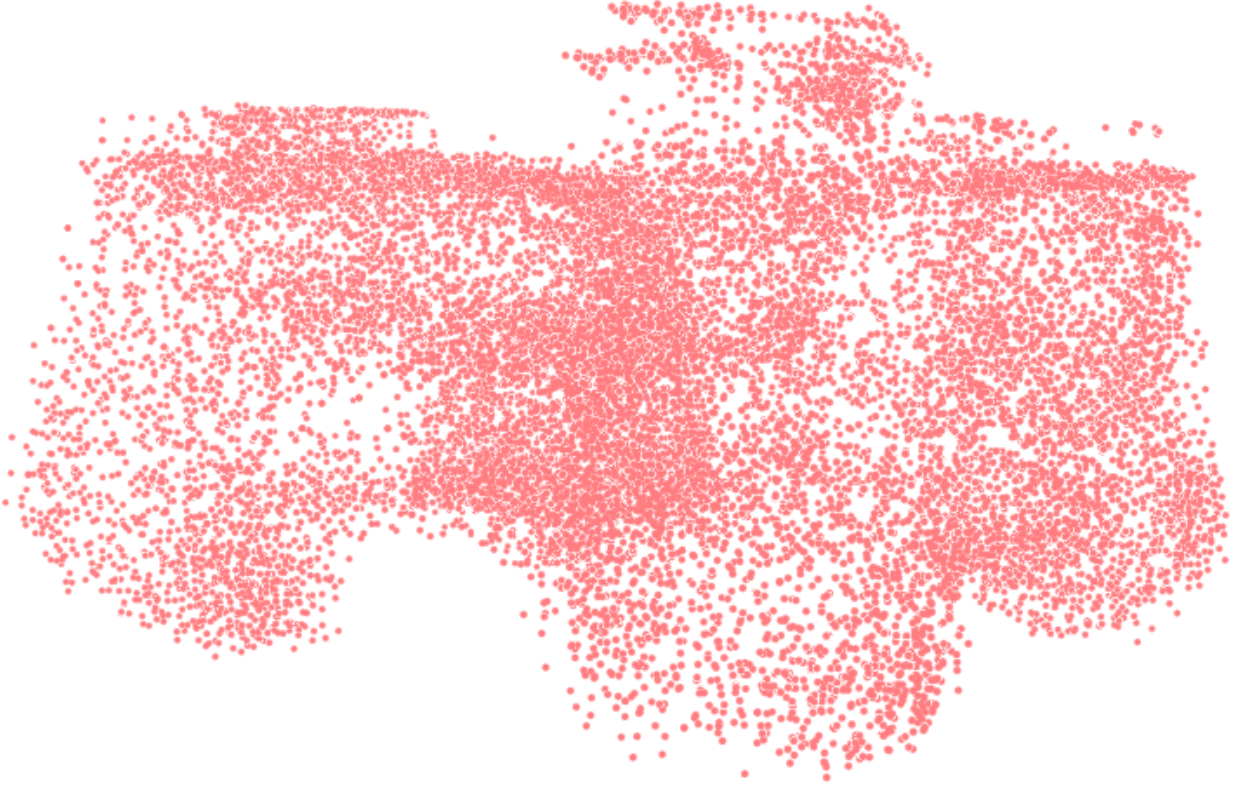}
	\end{subfigure}\hfill%
	\begin{subfigure}{\sunit}
		\centering
		\includegraphics[width=\linewidth]{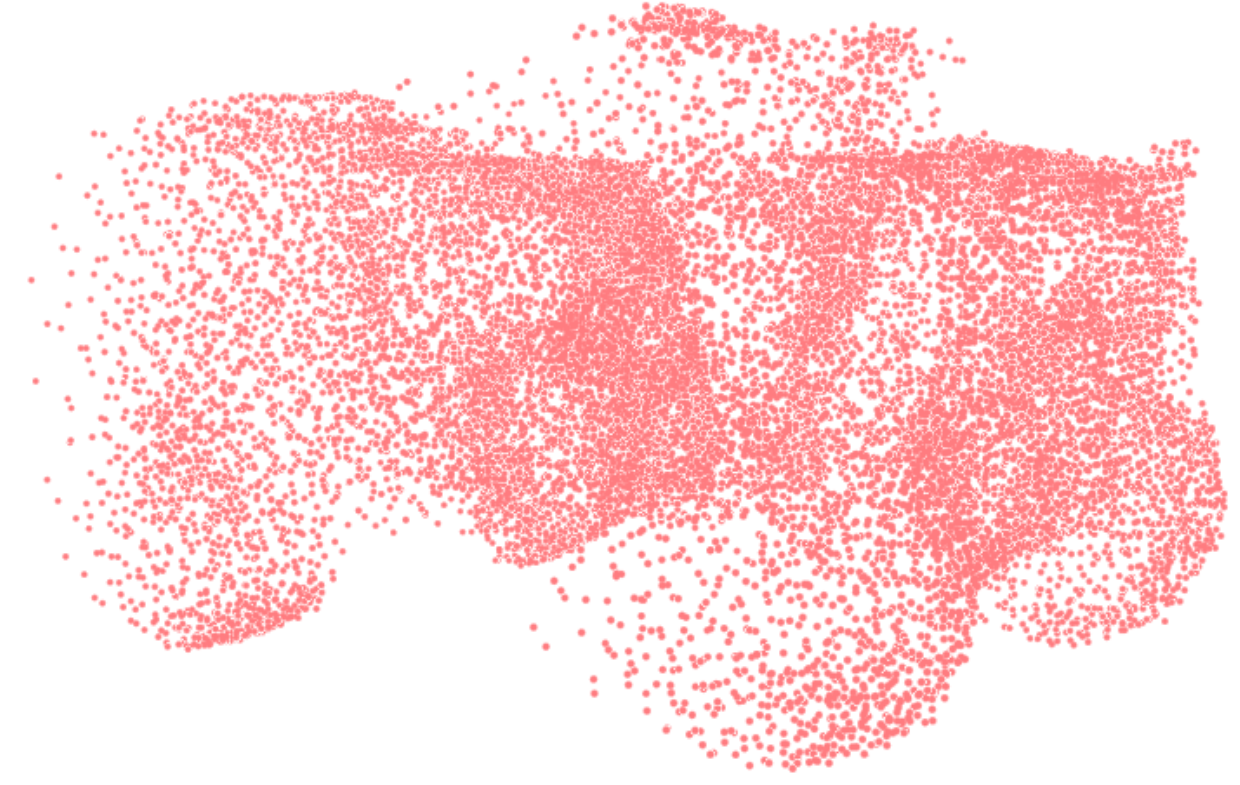}
	\end{subfigure}\hfill%
	\begin{subfigure}{\sunit}
		\centering
		\includegraphics[width=\linewidth]{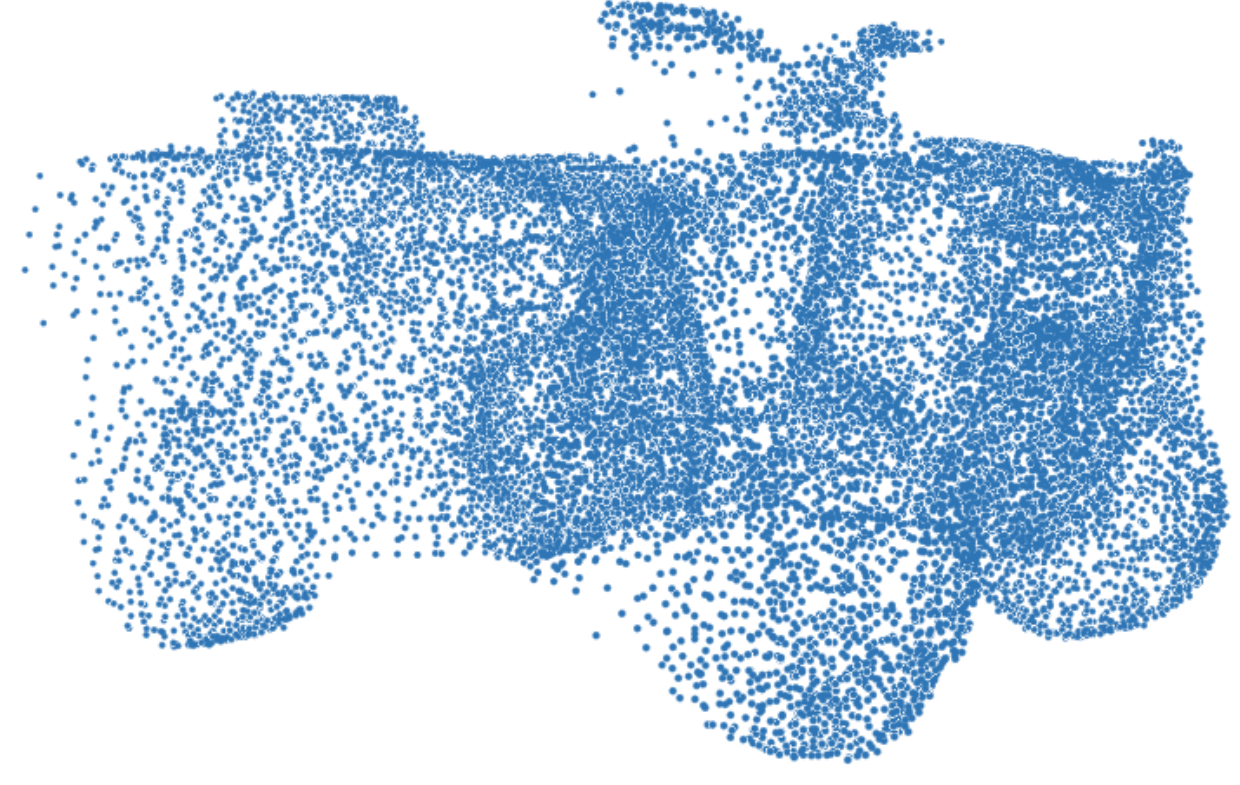}
	\end{subfigure}\hfill%
	\begin{subfigure}{\sunit}
		\centering
		\includegraphics[width=\linewidth]{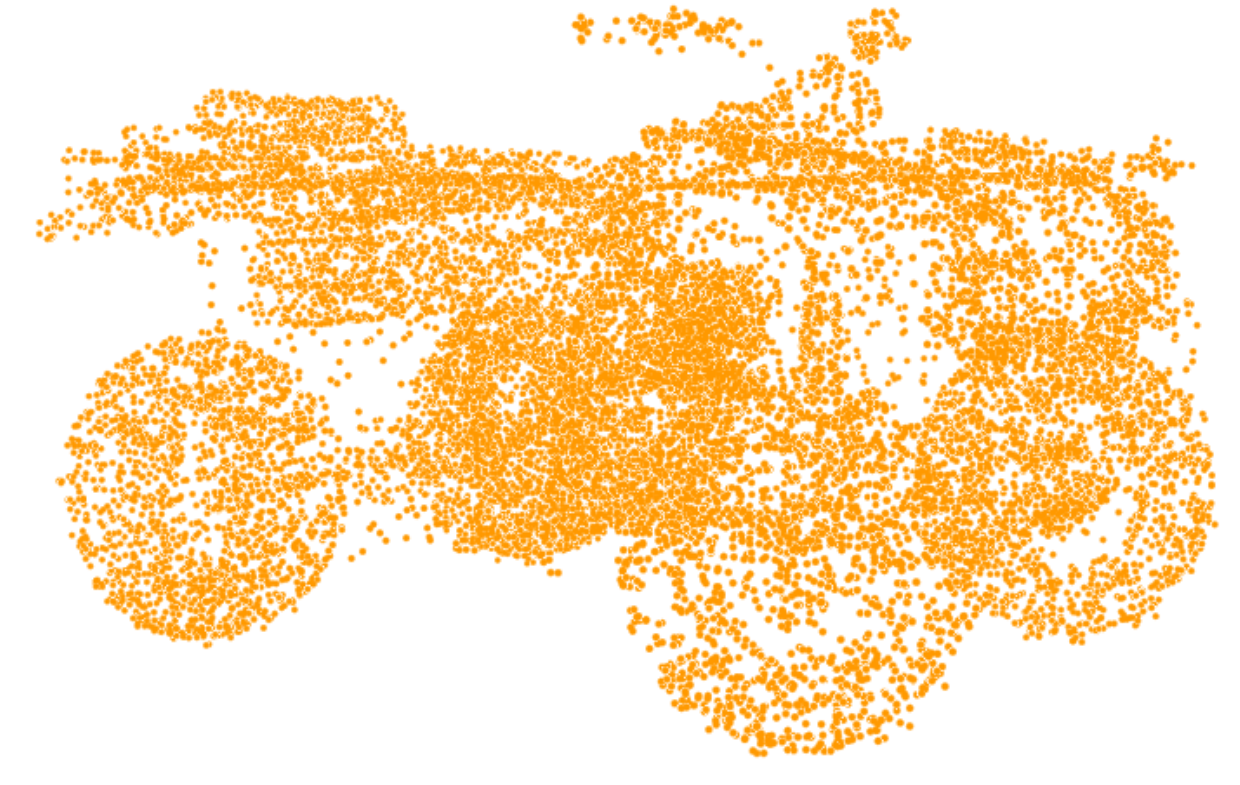}
	\end{subfigure}

	\vspace{5pt}
	
	\captionsetup[subfigure]{font=small,labelfont=small}
	\begin{subfigure}{\sunit}
		\centering
		\includegraphics[width=\sunita]{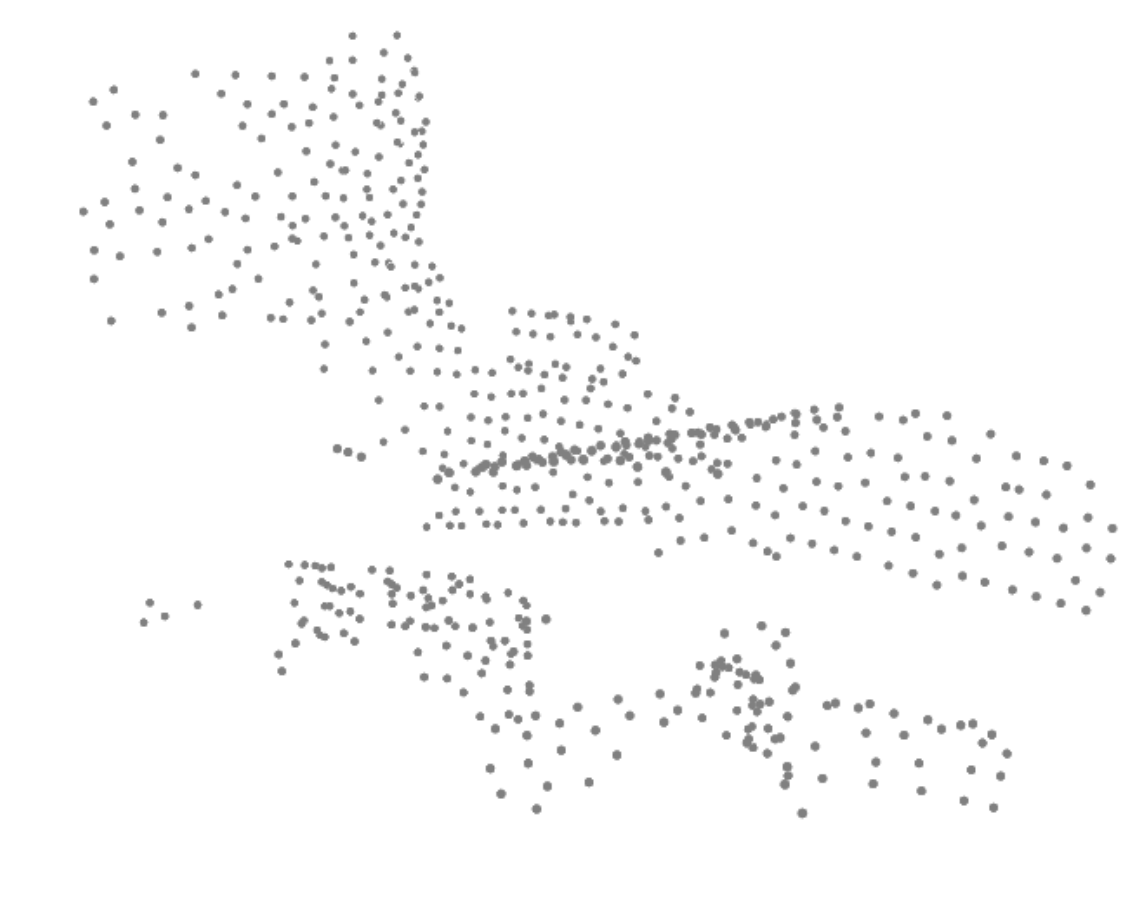}
		\caption{Input}
	\end{subfigure}\hfill%
	\begin{subfigure}{\sunit}
		\centering
		\includegraphics[width=\sunita]{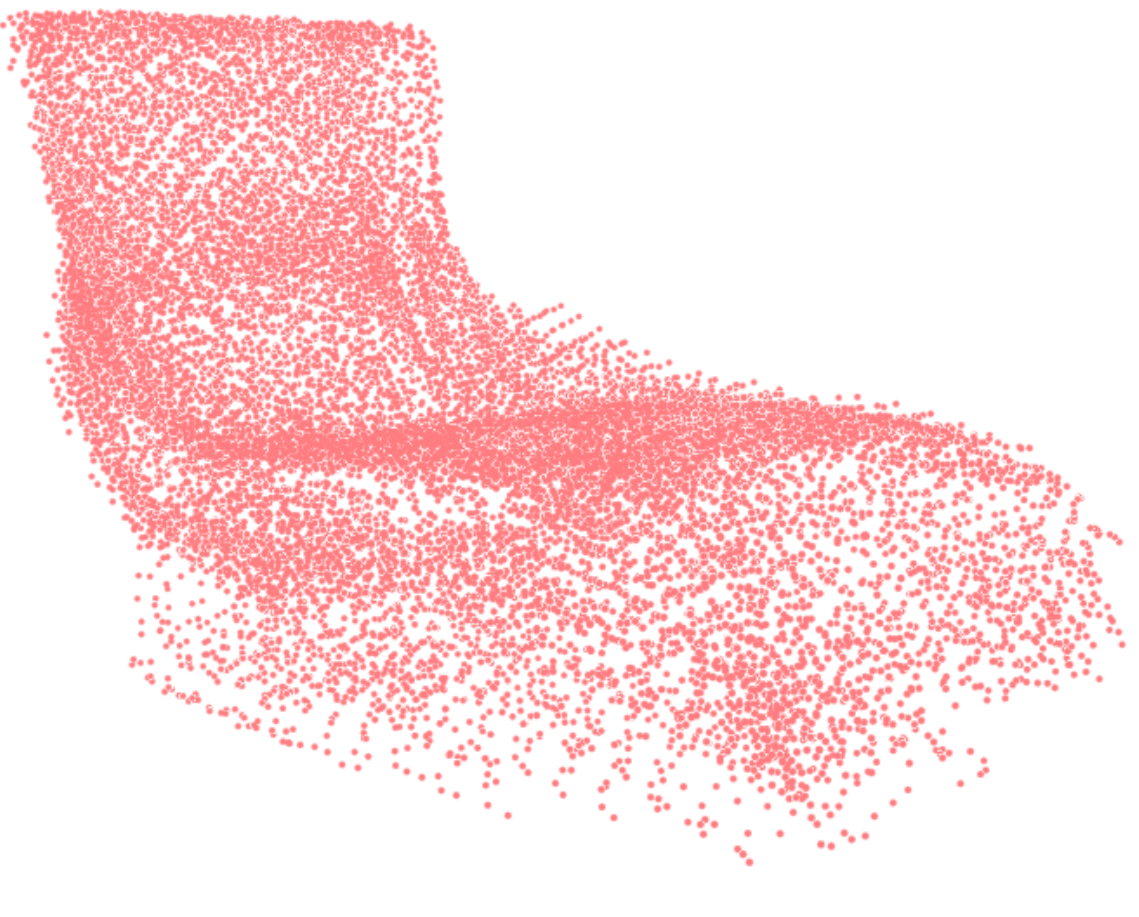}
		\caption{PCN}
	\end{subfigure}\hfill%
	\begin{subfigure}{\sunit}
		\centering
		\includegraphics[width=\sunita]{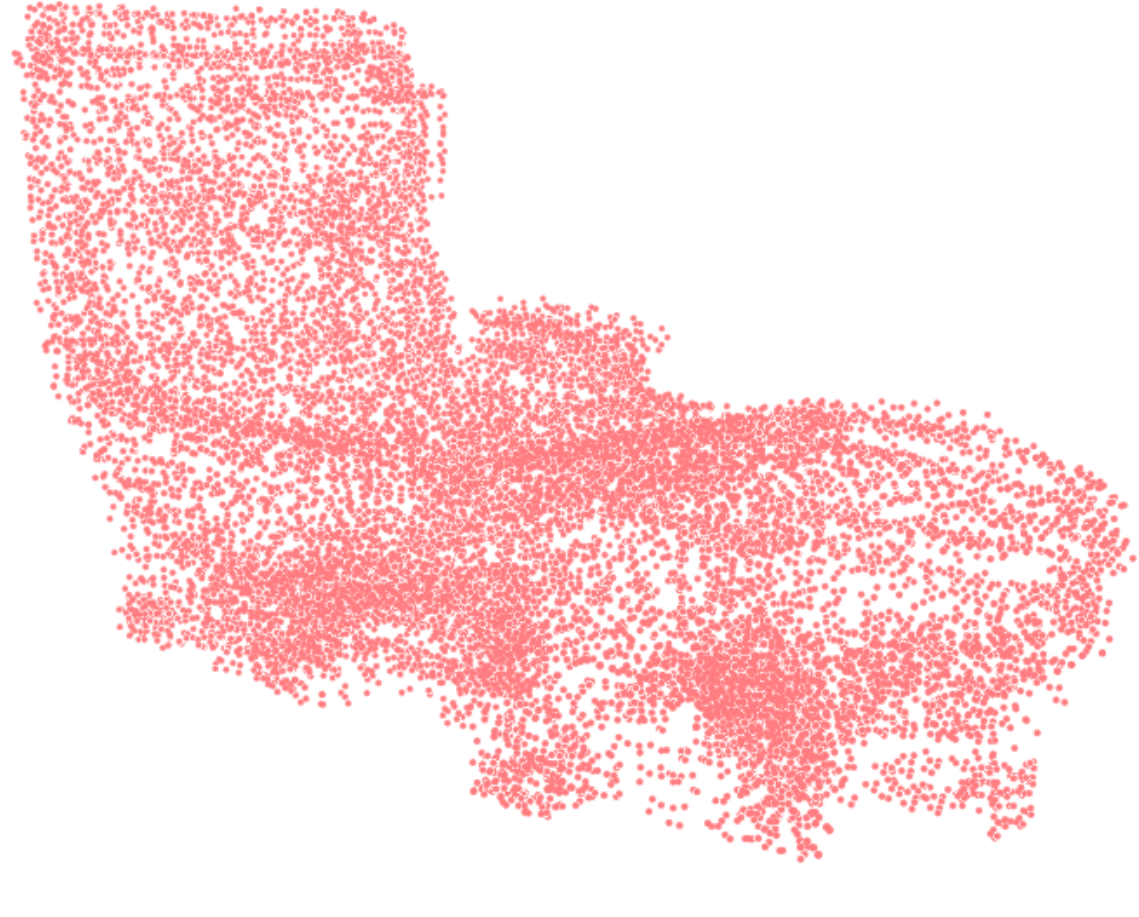}
		\caption{GRNet}
	\end{subfigure}\hfill%
	\begin{subfigure}{\sunit}
		\centering
		\includegraphics[width=\sunita]{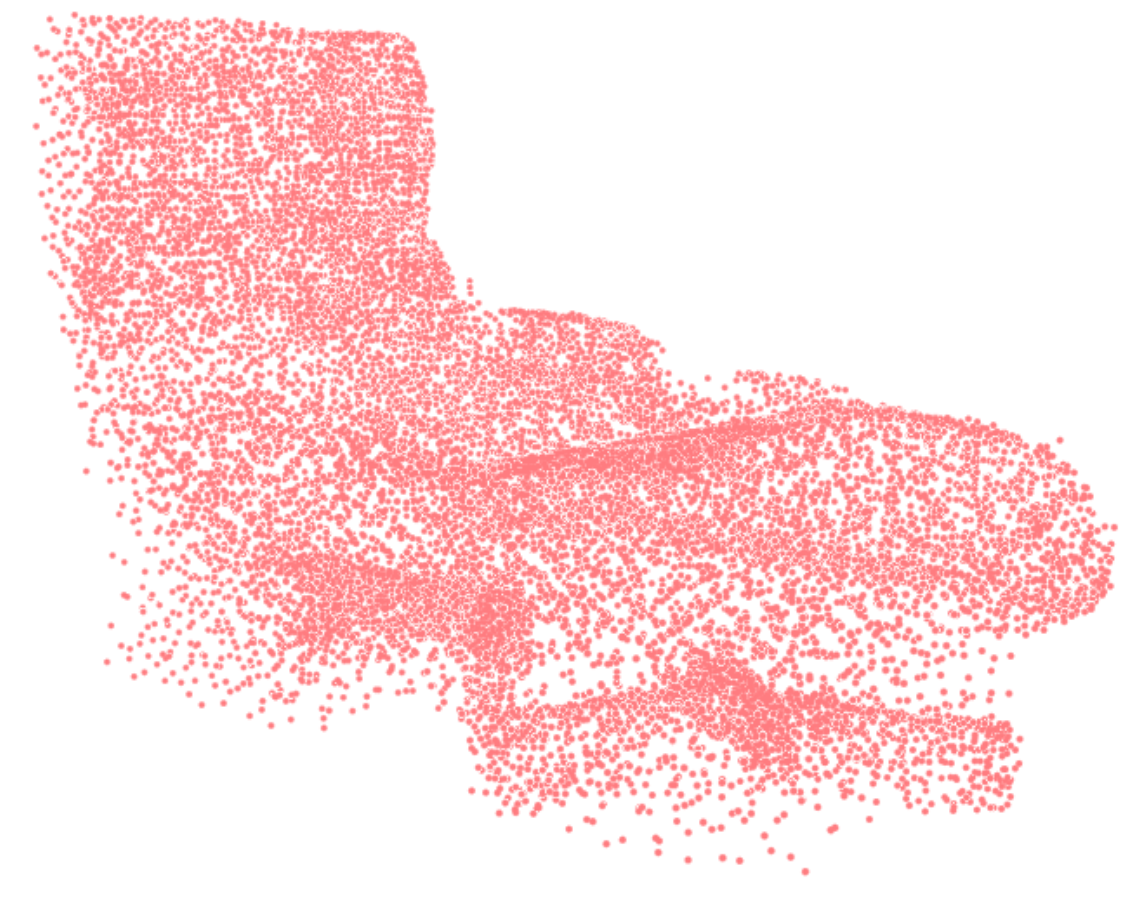}
		\caption{Snowflake}
	\end{subfigure}\hfill%
	\begin{subfigure}{\sunit}
		\centering
		\includegraphics[width=\sunita]{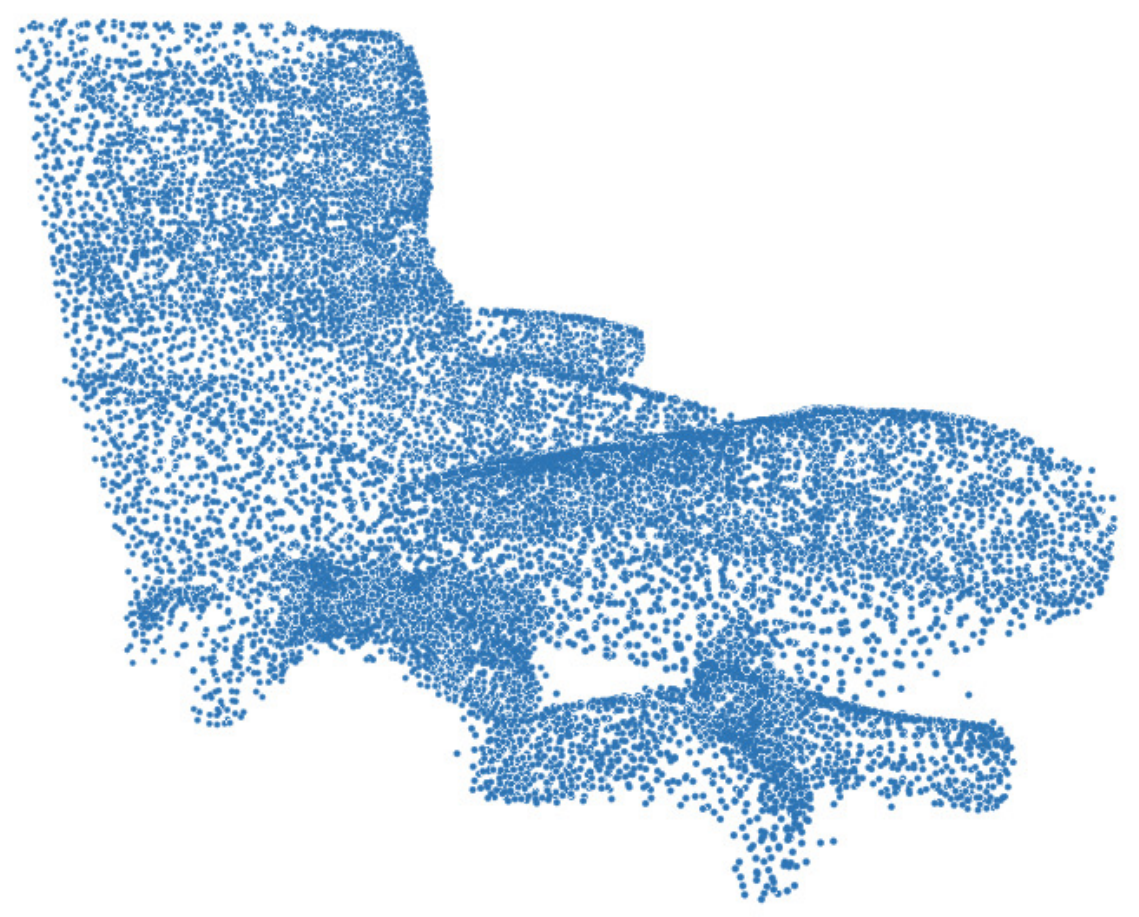}
		\caption{Ours}
	\end{subfigure}\hfill%
	\begin{subfigure}{\sunit}
		\centering
		\includegraphics[width=\sunita]{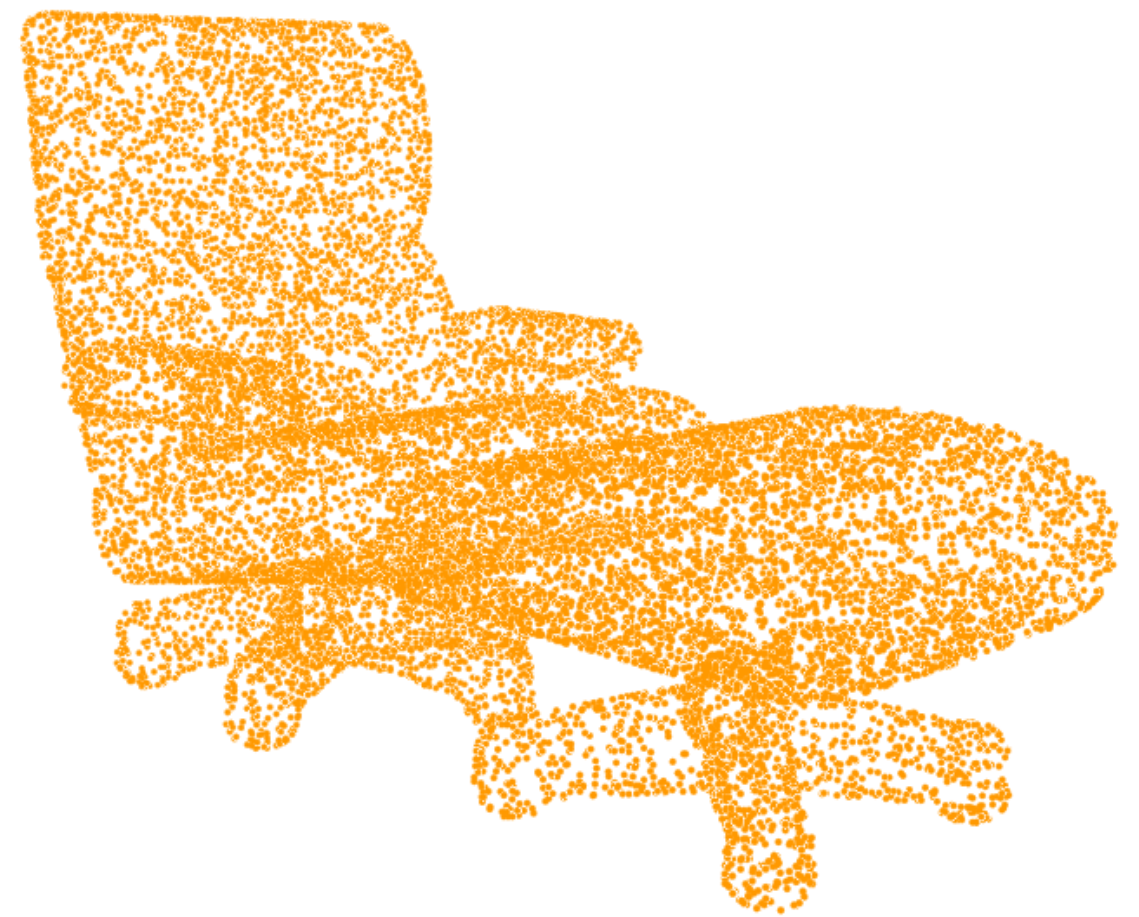}
		\caption{GT}
	\end{subfigure}
	
	\vspace{5pt}
	\caption{Visual comparisons on PCN dataset.}
	\label{fig:pcn_more}
\end{figure*}

\textbf{Qualitative results on PCN dataset.} In Fig.~\ref{fig:pcn_more}, we provide more visual results compared with PCN~\cite{yuan2018pcn}, GRNet~\cite{xie2020grnet} and SnowflakeNet~\cite{xiang2021snowflakenet}. All the results are obtained from their released pretrained models. We can see that SeedFormer performs clearly better than previous methods.

\vspace{5pt}
\noindent\textbf{More results on ShapeNet-55/34.} We report complete results of our method on ShapeNet-55 in Tab.~\ref{table:shapenet55_more} and results of novel categories on ShapeNet-34 in Tab.~\ref{table:shapenet21_more}. The models are tested under three difficulty levels: simple, moderate and hard. For novel objects of ShapeNet-34, we can see that SeedFormer achieves best scores on all categories.

\begin{table*}[t]
	\centering
	\scriptsize
	\setlength{\tabcolsep}{4.5pt}
	\caption{Detailed results for novel 21 categories on ShapeNet-34 dataset. $S.$, $M.$ and $H.$ stand for the simple, moderate and hard difficulty levels.}
	\vspace{5pt}
	\begin{tabular}{l|ccc|ccc|ccc|ccc}
		\toprule[1pt]
		& \multicolumn{3}{c|}{PCN~\cite{yuan2018pcn}} & \multicolumn{3}{c|}{GRNet~\cite{xie2020grnet}} & \multicolumn{3}{c|}{PoinTr~\cite{yu2021pointr}} & \multicolumn{3}{c}{SeedFormer} \\
		& S. & M. & H. & S. & M. & H. & S. & M. & H. & S. & M. & H.\\
		\midrule[0.3pt]
		bag			& 2.48 & 2.46 & 3.94 & 1.47 & 1.88 & 3.45 & 0.96 & 1.34 & 2.08 & \textbf{0.49} & \textbf{0.82} & \textbf{1.45} \\
		basket		& 2.79 & 2.51 & 4.78 & 1.78 & 1.94 & 4.18 & 1.04 & 1.40 & 2.90 & \textbf{0.60} & \textbf{0.85} & \textbf{1.98} \\
		birdhouse	& 3.53 & 3.47 & 5.31 & 1.89 & 2.34 & 5.16 & 1.22 & 1.79 & 3.45 & \textbf{0.72} & \textbf{1.19} & \textbf{2.31} \\
		bowl		& 2.66 & 2.35 & 3.97 & 1.77 & 1.97 & 3.90 & 1.05 & 1.32 & 2.40 & \textbf{0.60} & \textbf{0.77} & \textbf{1.50} \\
		camera		& 4.84 & 5.30 & 8.03 & 2.31 & 3.38 & 7.20 & 1.63 & 2.67 & 4.97 & \textbf{0.89} & \textbf{1.77} & \textbf{3.75} \\
		can			& 1.95 & 1.89 & 5.21 & 1.53 & 1.80 & 3.08 & 0.80 & 1.17 & 2.85 & \textbf{0.56} & \textbf{0.89} & \textbf{1.57} \\
		cap			& 7.21 & 7.14 & 10.94 & 3.29 & 4.87 & 13.02 & 1.40 & 2.74 & 8.35 & \textbf{0.50} & \textbf{1.34} & \textbf{5.19} \\
		keyboard	& 1.07 & 1.00 & 1.23 & 0.73 & 0.77 & 1.11 & 0.43 & 0.45 & 0.63 & \textbf{0.32} & \textbf{0.41} & \textbf{0.60} \\
		dishwasher	& 2.45 & 2.09 & 3.53 & 1.79 & 1.70 & 3.27 & 0.93 & 1.05 & 2.04 & \textbf{0.63} & \textbf{0.78} & \textbf{1.44} \\
		earphone	& 7.88 & 6.59 & 16.53 & 4.29 & 4.16 & 10.30 & 2.03 & 5.10 & 10.69 & \textbf{1.18} & \textbf{2.78} & \textbf{6.71} \\
		helmet		& 6.15 & 6.41 & 9.16 & 3.06 & 4.38 & 10.27 & 1.86 & 3.30 & 6.96 & \textbf{1.10} & \textbf{2.27} & \textbf{4.78} \\
		mailbox		& 2.74 & 2.68 & 4.31 & 1.52 & 1.90 & 4.33 & 1.03 & 1.47 & 3.34 & \textbf{0.56} & \textbf{0.99} & \textbf{2.06} \\
		microphone	& 4.36 & 4.65 & 8.46 & 2.29 & 3.23 & 8.41 & 1.25 & 2.27 & 5.47 & \textbf{0.80} & \textbf{1.61} & \textbf{4.21} \\
		microwaves	& 2.59 & 2.35 & 4.47 & 1.74 & 1.81 & 3.82 & 1.01 & 1.18 & 2.14 & \textbf{0.64} & \textbf{0.83} & \textbf{1.69} \\
		pillow		& 2.09 & 2.16 & 3.54 & 1.43 & 1.69 & 3.43 & 0.92 & 1.24 & 2.39 & \textbf{0.43} & \textbf{0.66} & \textbf{1.45} \\
		printer		& 3.28 & 3.60 & 5.56 & 1.82 & 2.41 & 5.09 & 1.18 & 1.76 & 3.10 & \textbf{0.69} & \textbf{1.25} & \textbf{2.33} \\
		remote		& 0.95 & 1.08 & 1.58 & 0.82 & 1.02 & 1.29 & 0.44 & 0.58 & 0.78 & \textbf{0.27} & \textbf{0.42} & \textbf{0.61} \\
		rocket		& 1.39 & 1.22 & 2.01 & 0.97 & 0.79 & 1.60 & 0.39 & 0.72 & 1.39 & \textbf{0.28} & \textbf{0.51} & \textbf{1.02} \\
		skateboard	& 1.97 & 1.78 & 2.45 & 0.93 & 1.07 & 1.83 & 0.52 & 0.80 & 1.31 & \textbf{0.35} & \textbf{0.56} & \textbf{0.92} \\
		tower		& 2.37 & 2.40 & 4.35 & 1.35 & 1.80 & 3.85 & 0.82 & 1.35 & 2.48 & \textbf{0.51} & \textbf{0.92} & \textbf{1.87} \\
		washer		& 2.77 & 2.52 & 4.64 & 1.83 & 1.97 & 5.28 & 1.04 & 1.39 & 2.73 & \textbf{0.61} & \textbf{0.87} & \textbf{1.94} \\
		\midrule[0.3pt]
		mean		& 3.22 & 3.13 & 5.43 & 1.84 & 2.23 & 4.95 & 1.05 & 1.67 & 3.45 & \textbf{0.61} & \textbf{1.07} & \textbf{2.35} \\
		\bottomrule[1pt]
	\end{tabular}
	
	\label{table:shapenet21_more}
\end{table*}

\begin{table*}[t]
	\centering
	\scriptsize
	\setlength{\tabcolsep}{4.5pt}
	\caption{Detailed results on ShapeNet-55 dataset. $S.$, $M.$ and $H.$ stand for the simple, moderate and hard difficulty levels.}
	\vspace{5pt}
	\begin{tabular}{l|ccc|ccc|ccc|ccc}
		\toprule[1pt]
		 & \multicolumn{3}{c|}{PCN~\cite{yuan2018pcn}} & \multicolumn{3}{c|}{GRNet~\cite{xie2020grnet}} & \multicolumn{3}{c|}{PoinTr~\cite{yu2021pointr}} & \multicolumn{3}{c}{SeedFormer} \\
				& S. & M. & H. & S. & M. & H. & S. & M. & H. & S. & M. & H.\\
		\midrule[0.3pt]
		airplane 	   & 0.90 & 0.89 & 1.32 & 0.87 & 0.87 & 1.27 & 0.27 & 0.38 & 0.69 & \textbf{0.23} & \textbf{0.35} & \textbf{0.61} \\
		trash-bin 	   & 2.16 & 2.18 & 5.15 & 1.69 & 2.01 & 3.48 & 0.80 & 1.15 & 2.15 & \textbf{0.73} & \textbf{1.08} & \textbf{1.94} \\
		bag 		   & 2.11 & 2.04 & 4.44 & 1.41 & 1.70 & 2.97 & 0.53 & 0.74 & 1.51 & \textbf{0.43} & \textbf{0.67} & \textbf{1.28} \\
		basket 		   & 2.21 & 2.10 & 4.55 & 1.65 & 1.84 & 3.15 & 0.73 & 0.88 & 1.82 & \textbf{0.65} & \textbf{0.83} & \textbf{1.54} \\
		bathtub 	   & 2.11 & 2.09 & 3.94 & 1.46 & 1.73 & 2.73 & 0.64 & 0.94 & 1.68 & \textbf{0.52} & \textbf{0.82} & \textbf{1.45} \\
		bed 		   & 2.86 & 3.07 & 5.54 & 1.64 & 2.03 & 3.70 & 0.76 & 1.10 & 2.26 & \textbf{0.63} & \textbf{0.91} & \textbf{1.89} \\
		bench 		   & 1.31 & 1.24 & 2.14 & 1.03 & 1.09 & 1.71 & 0.38 & 0.52 & 0.94 & \textbf{0.32} & \textbf{0.42} & \textbf{0.84} \\
		birdhouse 	   & 3.29 & 3.53 & 6.69 & 1.87 & 2.40 & 4.71 & 0.98 & 1.49 & 3.13 & \textbf{0.76} & \textbf{1.30} & \textbf{2.46} \\
		bookshelf 	   & 2.70 & 2.70 & 4.61 & 1.42 & 1.71 & 2.78 & 0.71 & 1.06 & 1.93 & \textbf{0.57} & \textbf{0.84} & \textbf{1.57} \\
		bottle 		   & 1.25 & 1.43 & 4.61 & 1.05 & 1.44 & 2.67 & 0.37 & 0.74 & 1.50 & \textbf{0.31} & \textbf{0.63} & \textbf{1.21} \\
		bowl 		   & 2.05 & 1.83 & 3.66 & 1.60 & 1.77 & 2.99 & 0.68 & 0.78 & 1.44 & \textbf{0.56} & \textbf{0.65} & \textbf{1.18} \\
		bus 		   & 1.20 & 1.14 & 2.08 & 1.06 & 1.16 & 1.48 & 0.42 & 0.55 & 0.79 & \textbf{0.42} & \textbf{0.55} & \textbf{0.73} \\
		cabinet 	   & 1.60 & 1.49 & 3.47 & 1.27 & 1.41 & 2.09 & \textbf{0.55} & \textbf{0.66} & 1.16 & 0.57 & 0.69 & \textbf{1.05} \\
		camera 		   & 4.05 & 4.54 & 8.27 & 2.14 & 3.15 & 6.09 & 1.10 & 2.03 & 4.34 & \textbf{0.83} & \textbf{1.68} & \textbf{3.45} \\
		can 		   & 2.02 & 2.28 & 6.48 & 1.58 & 2.11 & 3.81 & 0.68 & 1.19 & 2.14 & \textbf{0.58} & \textbf{1.03} & \textbf{1.79} \\
		cap 		   & 1.82 & 1.76 & 4.20 & 1.17 & 1.37 & 3.05 & 0.46 & 0.62 & 1.64 & \textbf{0.33} & \textbf{0.45} & \textbf{1.18} \\
		car 		   & 1.48 & 1.47 & 2.60 & 1.29 & 1.48 & 2.14 & \textbf{0.64} & 0.86 & 1.25 & 0.65 & \textbf{0.86} & \textbf{1.17} \\
		cellphone 	   & 0.80 & 0.79 & 1.71 & 0.82 & 0.91 & 1.18 & 0.32 & \textbf{0.39} & 0.60 & \textbf{0.31} & 0.40 & \textbf{0.54} \\
		chair 		   & 1.70 & 1.81 & 3.34 & 1.24 & 1.56 & 2.73 & 0.49 & 0.74 & 1.63 & \textbf{0.41} & \textbf{0.65} & \textbf{1.38} \\
		clock 		   & 2.10 & 2.01 & 3.98 & 1.46 & 1.66 & 2.67 & 0.62 & 0.84 & 1.65 & \textbf{0.53} & \textbf{0.74} & \textbf{1.35} \\
		keyboard 	   & 0.82 & 0.82 & 1.04 & 0.74 & 0.81 & 1.09 & 0.30 & 0.39 & \textbf{0.45} & \textbf{0.28} & \textbf{0.36} & \textbf{0.45} \\
		dishwasher 	   & 1.93 & 1.66 & 4.39 & 1.43 & 1.59 & 2.53 & \textbf{0.55} & 0.69 & 1.42 & 0.56 & \textbf{0.69} & \textbf{1.30} \\
		display 	   & 1.56 & 1.66 & 3.26 & 1.13 & 1.38 & 2.29 & 0.48 & 0.67 & 1.33 & \textbf{0.39} & \textbf{0.59} & \textbf{1.10} \\
		earphone 	   & 3.13 & 2.94 & 7.56 & 1.78 & 2.18 & 5.33 & 0.81 & 1.38 & 3.78 & \textbf{0.64} & \textbf{1.04} & \textbf{2.75} \\
		faucet 		   & 3.21 & 3.48 & 7.52 & 1.81 & 2.32 & 4.91 & 0.71 & 1.42 & 3.49 & \textbf{0.55} & \textbf{1.15} & \textbf{2.63} \\
		filecabinet    & 2.02 & 1.97 & 4.14 & 1.46 & 1.71 & 2.89 & \textbf{0.63} & \textbf{0.84} & 1.69 & \textbf{0.63} & \textbf{0.84} & \textbf{1.49} \\
		guitar 		   & 0.42 & 0.38 & 1.23 & 0.44 & 0.48 & 0.76 & 0.14 & 0.21 & 0.42 & \textbf{0.13} & \textbf{0.19} & \textbf{0.32} \\
		helmet 		   & 3.76 & 4.18 & 7.53 & 2.33 & 3.18 & 6.03 & 0.99 & 1.93 & 4.22 & \textbf{0.79} & \textbf{1.52} & \textbf{3.61} \\
		jar 		   & 2.57 & 2.82 & 6.00 & 1.72 & 2.37 & 4.37 & 0.77 & 1.33 & 2.87 & \textbf{0.63} & \textbf{1.13} & \textbf{2.36} \\
		knife 		   & 0.94 & 0.62 & 1.37 & 0.72 & 0.66 & 0.96 & 0.20 & 0.33 & 0.56 & \textbf{0.15} & \textbf{0.28} & \textbf{0.45} \\
		lamp 		   & 3.10 & 3.45 & 7.02 & 1.68 & 2.43 & 5.17 & 0.64 & 1.40 & 3.58 & \textbf{0.45} & \textbf{1.06} & \textbf{2.67} \\
		laptop 		   & 0.75 & 0.79 & 1.59 & 0.83 & 0.87 & 1.28 & \textbf{0.32} & \textbf{0.34} & 0.60 & \textbf{0.32} & 0.37 & \textbf{0.55} \\
		loudspeaker    & 2.50 & 2.45 & 5.08 & 1.75 & 2.08 & 3.45 & 0.78 & 1.16 & 2.17 & \textbf{0.67} & \textbf{1.01} & \textbf{1.80} \\
		mailbox 	   & 1.66 & 1.74 & 5.18 & 1.15 & 1.59 & 3.42 & 0.39 & 0.78 & 2.56 & \textbf{0.30} & \textbf{0.67} & \textbf{2.04} \\
		microphone 	   & 3.44 & 3.90 & 8.52 & 2.09 & 2.76 & 5.70 & 0.70 & 1.66 & 4.48 & \textbf{0.62} & \textbf{1.61} & \textbf{3.66} \\
		microwaves 	   & 2.20 & 2.01 & 4.65 & 1.51 & 1.72 & 2.76 & 0.67 & 0.83 & 1.82 & \textbf{0.63} & \textbf{0.79} & \textbf{1.47} \\
		motorbike 	   & 2.03 & 2.01 & 3.13 & 1.38 & 1.52 & 2.26 & 0.75 & 1.10 & 1.92 & \textbf{0.68} & \textbf{0.96} & \textbf{1.44} \\
		mug 		   & 2.45 & 2.48 & 5.17 & 1.75 & 2.16 & 3.79 & 0.91 & 1.17 & 2.35 & \textbf{0.79} & \textbf{1.03} & \textbf{2.06} \\
		piano 		   & 2.64 & 2.74 & 4.83 & 1.53 & 1.82 & 3.21 & 0.76 & 1.06 & 2.23 & \textbf{0.62} & \textbf{0.87} & \textbf{1.79} \\
		pillow 		   & 1.85 & 1.81 & 3.68 & 1.42 & 1.67 & 3.04 & 0.61 & 0.82 & 1.56 & \textbf{0.48} & \textbf{0.75} & \textbf{1.41} \\
		pistol 		   & 1.25 & 1.17 & 2.65 & 1.11 & 1.06 & 1.76 & 0.43 & 0.66 & 1.30 & \textbf{0.37} & \textbf{0.56} & \textbf{0.96} \\
		flowerpot 	   & 3.32 & 3.39 & 6.04 & 2.02 & 2.48 & 4.19 & 1.01 & 1.51 & 2.77 & \textbf{0.93} & \textbf{1.30} & \textbf{2.32} \\
		printer 	   & 2.90 & 3.19 & 5.84 & 1.56 & 2.38 & 4.24 & 0.73 & 1.21 & 2.47 & \textbf{0.58} & \textbf{1.11} & \textbf{2.13} \\
		remote 		   & 0.99 & 0.97 & 2.04 & 0.89 & 1.05 & 1.29 & 0.36 & 0.53 & 0.71 & \textbf{0.29} & \textbf{0.46} & \textbf{0.62} \\
		rifle 		   & 0.98 & 0.80 & 1.31 & 0.83 & 0.77 & 1.16 & 0.30 & 0.45 & 0.79 & \textbf{0.27} & \textbf{0.41} & \textbf{0.66} \\
		rocket 		   & 1.05 & 1.04 & 1.87 & 0.78 & 0.92 & 1.44 & 0.23 & 0.48 & 0.99 & \textbf{0.21} & \textbf{0.46} & \textbf{0.83} \\
		skateboard 	   & 1.04 & 0.94 & 1.68 & 0.82 & 0.87 & 1.24 & 0.28 & 0.38 & 0.62 & \textbf{0.23} & \textbf{0.32} & \textbf{0.62} \\
		sofa 		   & 1.65 & 1.61 & 2.92 & 1.35 & 1.45 & 2.32 & 0.56 & 0.67 & 1.14 & \textbf{0.50} & \textbf{0.62} & \textbf{1.02} \\
		stove 		   & 2.07 & 2.02 & 4.72 & 1.46 & 1.72 & 3.22 & 0.63 & 0.92 & 1.73 & \textbf{0.59} & \textbf{0.87} & \textbf{1.49} \\
		table 		   & 1.56 & 1.50 & 3.36 & 1.15 & 1.33 & 2.33 & 0.46 & 0.64 & 1.31 & \textbf{0.41} & \textbf{0.58} & \textbf{1.18} \\
		telephone 	   & 0.80 & 0.80 & 1.67 & 0.81 & 0.89 & 1.18 & \textbf{0.31} & \textbf{0.38} & 0.59 & \textbf{0.31} & 0.39 & \textbf{0.55} \\
		tower 		   & 1.91 & 1.97 & 4.47 & 1.26 & 1.69 & 3.06 & 0.55 & 0.90 & 1.95 & \textbf{0.47} & \textbf{0.84} & \textbf{1.65} \\
		train 		   & 1.50 & 1.41 & 2.37 & 1.09 & 1.14 & 1.61 & \textbf{0.50} & 0.70 & 1.12 & 0.51 & \textbf{0.66} & \textbf{1.01} \\
		watercraft 	   & 1.46 & 1.39 & 2.40 & 1.09 & 1.12 & 1.65 & 0.41 & 0.62 & 1.07 & \textbf{0.35} & \textbf{0.56} & \textbf{0.92} \\
		washer 		   & 2.42 & 2.31 & 6.08 & 1.72 & 2.05 & 4.19 & 0.75 & 1.06 & 2.44 & \textbf{0.64} & \textbf{0.91} & \textbf{2.04} \\
		\midrule[0.3pt]
		mean 		   & 1.96 & 1.98 & 4.09 & 1.35 & 1.63 & 2.86 & 0.58 & 0.88 & 1.80 & \textbf{0.50} & \textbf{0.77} & \textbf{1.49} \\
		\bottomrule[1pt]
	\end{tabular}
	
	\label{table:shapenet55_more}
\end{table*}

%
%

\label{sec:supp}

\end{document}